\documentclass[preprint,12pt]{elsarticle}

\usepackage{amsmath,amsfonts}
\usepackage{array}
\usepackage{url}
\usepackage{mathtools}
\usepackage{graphicx, wrapfig, subcaption, setspace, booktabs}
\usepackage{algorithm, algpseudocode}
\usepackage{microtype}
\usepackage{caption}
\usepackage{hyperref}

\usepackage{graphicx} 

\journal{Neural Networks}

\begin{document}

\begin{frontmatter}

\title{Rethinking Deep Clustering Paradigms: Self-Supervision Is All You Need.}





\author[1]{\texorpdfstring{Amal Shaheen\corref{equal}}{Amal Shaheen}}
\author[2]{\texorpdfstring{Nairouz Mrabah\corref{equal}}{Nairouz Mrabah}}

\author[1]{Riadh Ksantini}
\author[1]{Abdulla Alqaddoumi}
\cortext[equal]{These authors contributed equally to this work.}

\address[1]{Computer Science, College of IT, UOB, Kingdom of Bahrain}
\address[2]{Computer Science, Université du Québec à Montréal, Montréal, QC, Canada}

\begin{abstract}
The recent advances in deep clustering have been made possible by significant progress in self-supervised and pseudo-supervised learning. However, the trade-off between self-supervision and pseudo-supervision can give rise to three primary issues. The joint training causes Feature Randomness and Feature Drift, whereas the independent training causes Feature Randomness and Feature Twist. In essence, using pseudo-labels generates random and unreliable features. The combination of pseudo-supervision and self-supervision drifts the reliable clustering-oriented features. Moreover, moving from self-supervision to pseudo-supervision can twist the curved latent manifolds. This paper addresses the limitations of existing deep clustering paradigms concerning Feature Randomness, Feature Drift, and Feature Twist. We propose a new paradigm with a new strategy that replaces pseudo-supervision with a second round of self-supervision training. The new strategy makes the transition between instance-level self-supervision and neighborhood-level self-supervision smoother and less abrupt. Moreover, it prevents the drifting effect that is caused by the strong competition between instance-level self-supervision and clustering-level pseudo-supervision. Moreover, the absence of the pseudo-supervision prevents the risk of generating random features. With this novel approach, our paper introduces a Rethinking of the Deep Clustering Paradigms, denoted by R-DC. Our model is specifically designed to address three primary challenges encountered in Deep Clustering: Feature Randomness, Feature Drift, and Feature Twist. Experimental results conducted on six datasets have shown that the two-level self-supervision training yields substantial improvements, as evidenced by the results of the clustering and ablation study. Furthermore, experimental comparisons with nine state-of-the-art clustering models have clearly shown that our strategy leads to a significant enhancement in performance.
\end{abstract}

\begin{highlights}

\item Analyses the drawbacks of pseudo-supervision from the perspective of Feature Randomness, Feature Drift, and Feature Twist and its limited capacity to improve the clustering results.

\item Introduces a novel deep clustering paradigm that eliminates pseudo-supervision in favor of a second round of self-supervision based on proximity-level information.

\item Analyses the advantages of the new paradigm from the perspective of Feature Randomness, Feature Drift, and Feature Twist. Eliminating pseudo-supervision prevents Feature Randomness and Feature Drift from taking place. Furthermore, the new paradigm alleviates Feature Twist.

\item Introduces a novel deep clustering approach that follows the proposed paradigm. Our method selects the core points and the most reliable neighbors of these points to perform proximity-level self-supervision. This ensures a smooth transition from instance-level to neighborhood-level self-supervision, which preserves the global structure of data.

\item Shows significant performance improvements on six datasets over state-of-the-art deep clustering methods. The obtained results confirm the effectiveness of the proposed approach in improving clustering accuracy and their capacity to address the geometric distortions under the transition regime from pretraining to finetuning.

\end{highlights}


\end{frontmatter}
 


\section{Introduction}\label{sec1}

\subsection{Background}
Clustering serves the purpose of grouping similar samples together, enabling a deeper understanding of the data and aiding in decision-making processes. Moreover, clustering plays a significant role in exploratory data analysis by revealing connections and inter-dependencies among variables. Its importance spans across various domains such as biology \cite{karim2021deep}, education 
, \cite{ezugwu2022comprehensive}, financial analysis \cite{kou2014evaluation}, social sciences \cite{chang2011classification}, and medical domain like 
brain tumor detection \cite{arunkumar2019k}, and predicting autism \cite{stevens2019identification}.

Traditional clustering approaches can be a good choice for low-dimensional, low-semantic, and small-scale datasets. For instance, K-means 
presumes that the clusters are spherical and equal in size, which may not be true for real-world datasets with curved clustering patterns. Another traditional method is hierarchical clustering \cite{ran2023comprehensive}, which builds a hierarchy of clusters from the bottom up or from the top down. However, it has limited capacity to deal with large-scale datasets. Clustering based on density approaches may also struggle to handle clusters of varied density. Partitioning Around Medoids (PAM) \cite{xu2005survey} and Fuzzy C-means (FCM) \cite{bezdek1984fcm} have found use in a variety of applications \cite{berget2008new}. On the one hand, PAM chooses representative objects known as medoids as cluster centers and seeks to minimize the sum of dissimilarities between data points and medoids. Nevertheless, this method has high computational complexity and is not well-suited for large-scale datasets. On the other hand, FCM provides fuzzy partitioning by assigning data points to numerous clusters with varying degrees of membership. However, similar to K-means, FCM assumes that the clusters are spherical and may struggle with curved clustering structures.


To address the limitation of traditional clustering methods, Deep Clustering (DC) has emerged as a cutting-edge strategy in the field of machine learning. It combines the power of deep learning with the principles of clustering to uncover hidden structures and patterns within complex data \cite{xu2021graph}. In particular, the DC strategy can be used to cluster high-dimensional, 
high-semantic, and large-scale datasets \cite{mrabah2024contrastive}, which are typically difficult to handle by the traditional clustering approaches. 


The notable advancements in deep clustering can be primarily credited to significant progress in self-supervision \cite{liu2022graph} and pseudo-supervision \cite{mrabah2020adversarial}. On one side, self-supervision involves solving a pretext task that necessitates a high-level comprehension of the data semantics to be solved. For instance, existing self-supervision techniques include predicting image colors \cite{zhang2016colorful}, predicting unpainted patches of images \cite{pathak2016context}, and predicting sequences of a ``jigsaw puzzle" \cite{noroozi2016unsupervised}. On the flip side, pseudo-supervision can address the primary task by leveraging a clustering algorithm to learn pseudo-labels. These pseudo-labels are then utilized as a supervisory signal to train the model. Based on pseudo-supervision, the trained model can acquire clustering-oriented features. 

The existing DC paradigms conventionally operate through the integration of pseudo-supervision and self-supervision, either \textit{sequentially} or \textit{concurrently}. Most deep clustering paradigms require two training phases: pretraining and finetuning. The most common deep clustering paradigm, which we describe as ``typical'', entails an initial pretraining phase driven by self-supervision, followed by a finetuning phase that leverages a linear combination of pseudo-supervision and self-supervision. This approach is deemed \emph{typical} due to its prevalent adoption and extensive use in the literature. However, the dual application of pseudo-supervision and self-supervision, either \textit{sequentially} or \textit{concurrently}, introduces various challenges. These challenges add to the error-prone nature of the pseudo-supervision task.


\subsection{Problem Statement}

\subsubsection{Feature Twist}


We examine the interplay between self-supervision and pseudo-supervision when they are applied sequentially, from a geometric perspective. Investigating the geometric characteristics within deep representations is an emerging field of research. It has a promising potential to provide valuable insights for downstream tasks. In particular, understanding the latent space geometry of deep clustering approaches can enhance clustering performance and facilitate the development of new algorithms. It is interesting to assess two metrics: Intrinsic-Dimension (ID) and Linear-Intrinsic-Dimension (LID). The ID measures the number of dimensions that are necessary to represent the data points. As for the LID, it measures the dimension of the minimal subspace capable of encompassing the latent manifolds. The value of LID - ID quantifies how many coordinates must be added to a space to transform it into a linear space. The higher the difference, the more curved the manifold. Due to the non-uniform distribution of points in a curved manifold, it is difficult to estimate its ID. Facco et al. \cite{facco2017estimating} have proposed a method called TwoNN that can assess the ID by taking into account the two nearest neighbors of each point. In another work, Ansuini et al.\cite{ansuini2019intrinsic} have adopted TwoNN for estimating the ID and have proposed an estimator for the LID based on Principal Component Analysis (PCA). As a result, they have found that training neural networks in a supervised way leads to the development of curved and low-dimensional latent manifolds. 

To understand the relation between self-supervision and pseudo-supervision under an independent training mode, Mrabah et al. \cite{mrabahescaping} have studied the evolution of the ID and LID of the latent manifolds for graph-structured data using Graph Auto-Encoders (GAEs). They have found that the coarse flattening of the latent manifolds under the shift regime from self-supervision to pseudo-supervision twists the curved structures, which in turn decreases the clustering effectiveness. This problem is called Feature Twist (FT) \cite{mrabahescaping}. The FT problem highlights the challenge of exploiting the curved manifolds generated by the pretraining phase to construct relevant Euclidean-based latent representations (i.e., latent representations that can be used to identify clusters using Euclidean distance). However, this problem remains unexplored outside the context of Graph Neural Networks (GNNs) and graphs.

\begin{figure*}
  \centering
  
 \begin{subfigure}[b]{0.33\textwidth}
    \includegraphics[width=\linewidth]{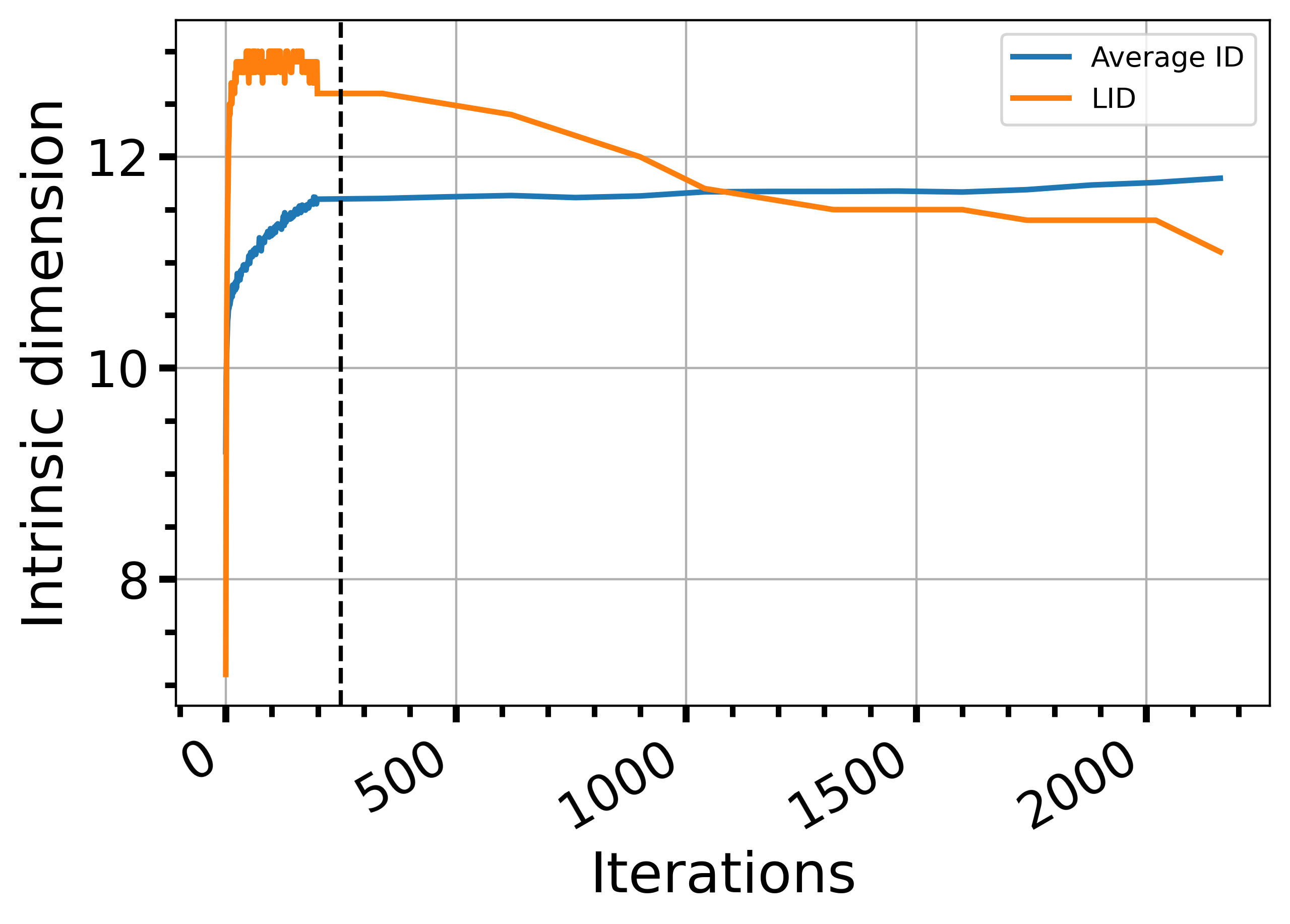}
    \caption{Phase 1: vanilla reconstruction; Phase 2: DEC loss; dataset: MNIST}
  \end{subfigure} \hfil
  \begin{subfigure}[b]{0.33\textwidth}
    \includegraphics[width=\linewidth]{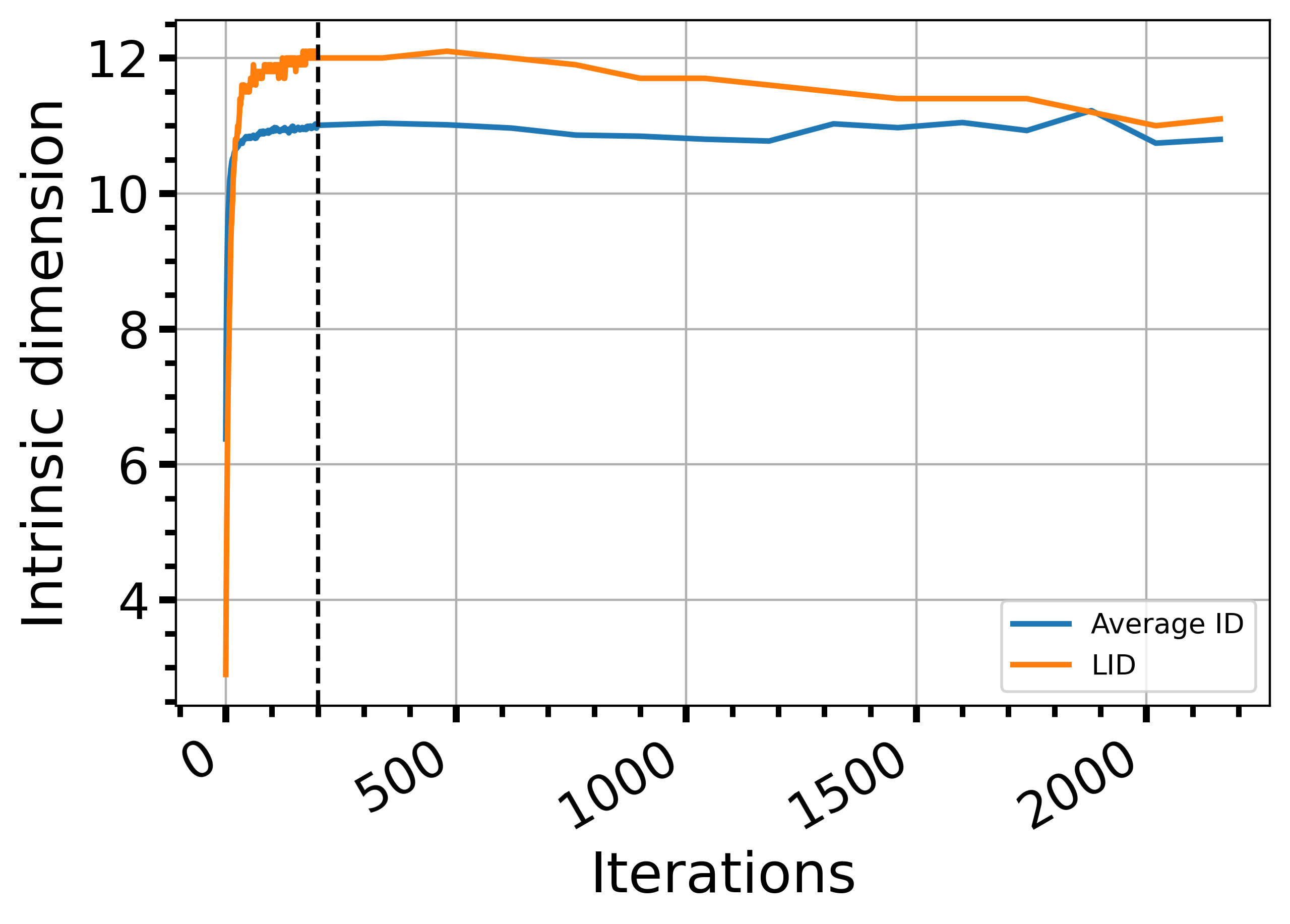}
    \caption{Phase 1: vanilla reconstruction; Phase 2: DEC loss; Dataset: FMNIST}
  \end{subfigure} \hfil

\begin{subfigure}[b]{0.33\textwidth}
    \includegraphics[width=\linewidth]{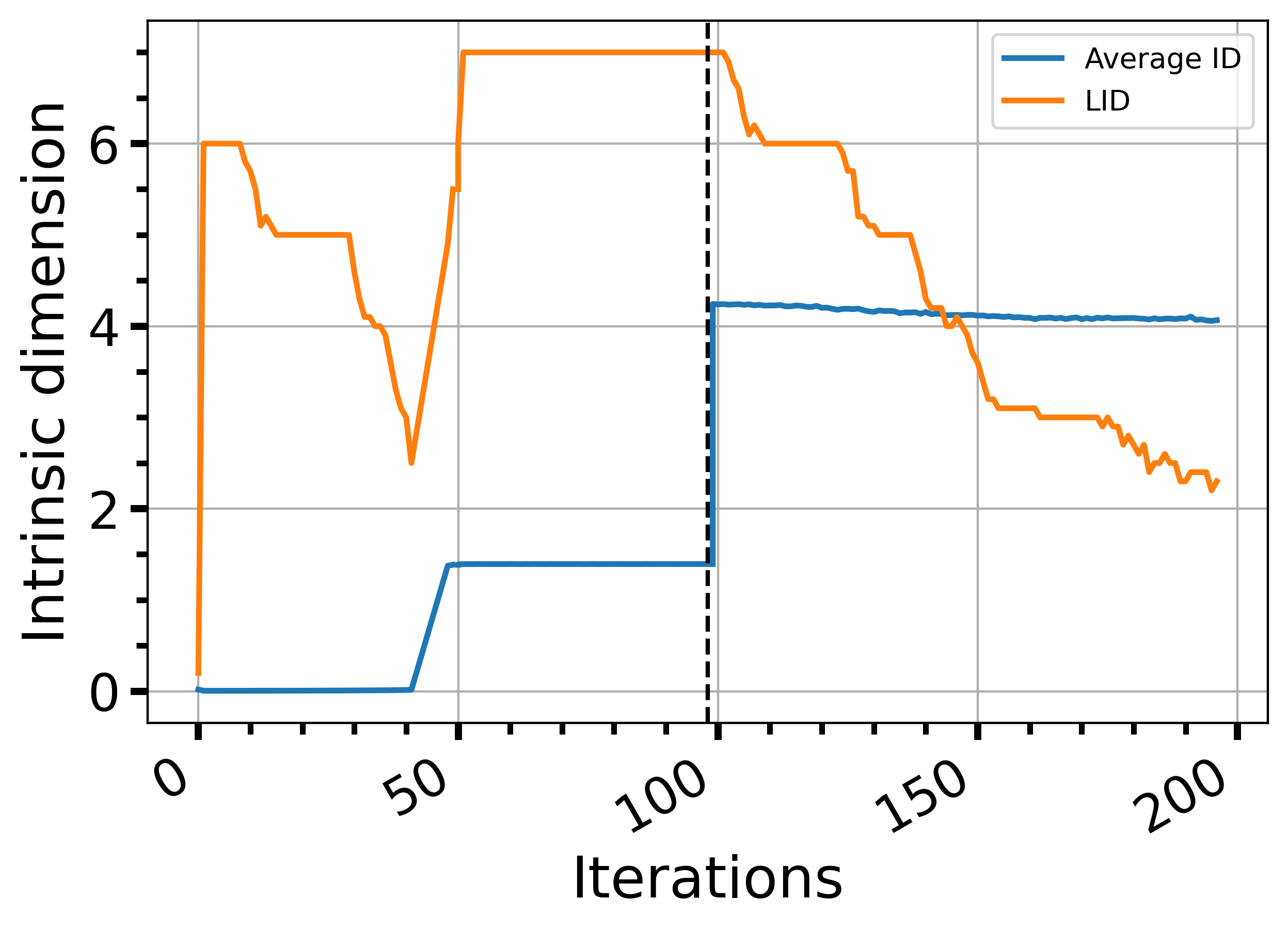}
    
    \caption{Phase 1: contrastive learning; Phase 2: DEC loss; dataset: MNIST}
  \end{subfigure} \hfil
  \begin{subfigure}[b]{0.33\textwidth}
    \includegraphics[width=\linewidth]{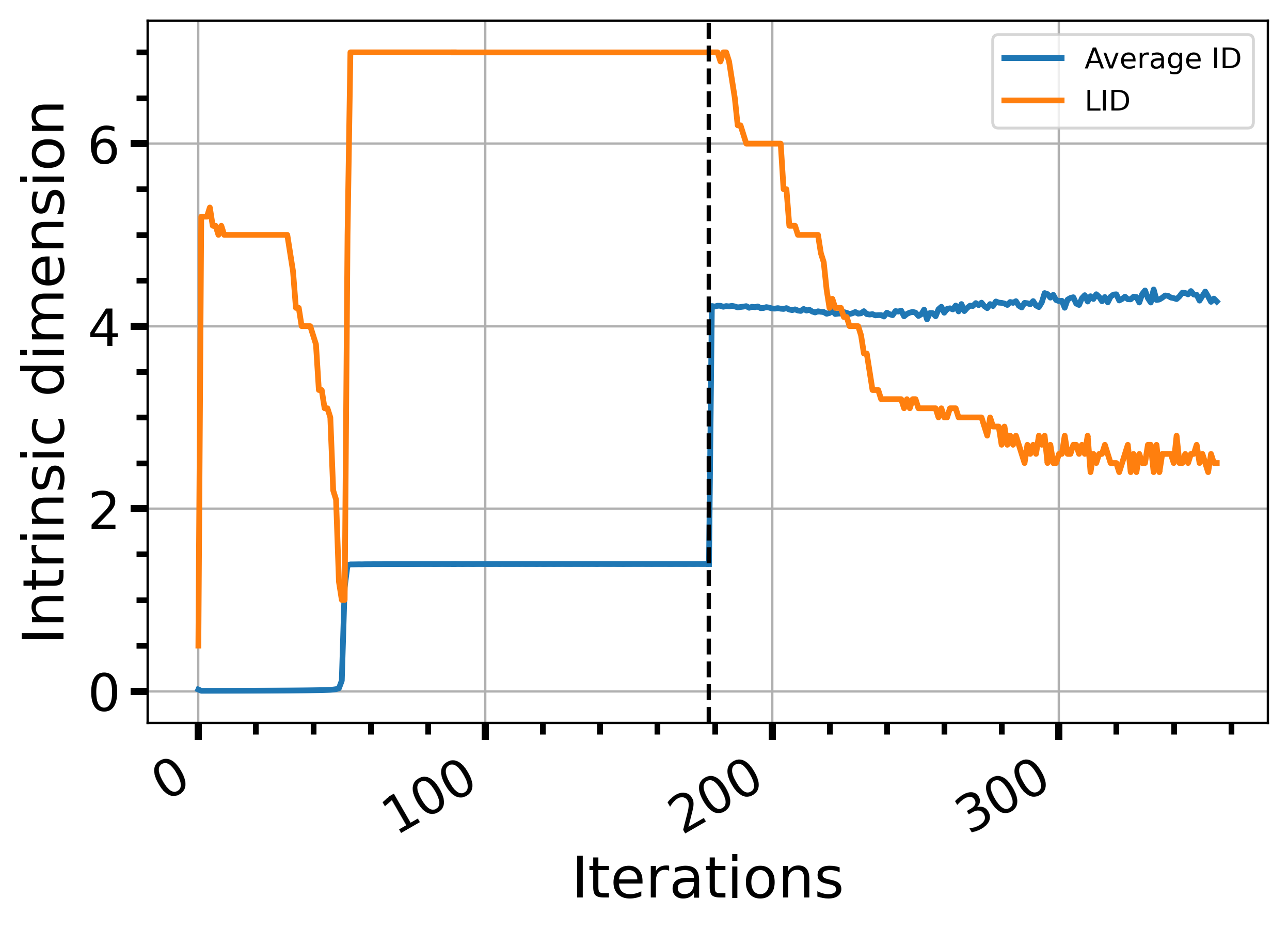}
    \caption{Phase 1: contrastive learning; Phase 2: DEC loss; Dataset: FMNIST}
  \end{subfigure} \hfil
  
  \caption{\textbf{First evidence of Feature Twist}. Average ID and LID of DEC on MNIST and FMNIST based on two pretraining strategies: vanilla reconstruction and instance-level contrastive learning. Average ID: average ID of the clustering manifolds. LID: number of dimensions that can capture $90\%$ of the covariance matrix (linear correlations) estimated based on PCA (Principal Component Analysis).}
  \label{fig:FT_1_FT}
\end{figure*}

In this work, we establish the existence of the FT problem for vanilla auto-encoders and Euclidean representations. Instance-level self-supervision techniques constitute one of the main pretraining tasks for most DC methods. Several variants of reconstruction functions (e.g., vanilla reconstruction, denoising reconstruction, interpolation-based reconstruction) \cite{Vincent2008extracting, Berthelot2019understanding} and instance-level contrastive learning \cite{chen2020simple, he2020momentum} are among the principal self-supervision tasks for pretraining the DC approaches \cite{mrabah2020adversarial, guo2017improved, li2021contrastive, dang2021nearest}. We conducted the first set of experiments by pretraining the deep Embedding Clustering model (DEC) \cite{xie2016unsupervised} for $400$ iterations based on two pretraining loss functions: vanilla reconstruction and instance-level contrastive learning. For contrastive learning, we opt for the Normalized Temperature-scaled Cross Entropy Loss \cite{chen2020simple}. We explore the evolution of the ID and the LID of the latent manifolds on two datasets (MNIST \cite{lecun2010mnist} and FMNIST \cite{xiao2017fashion}). In Figure \ref{fig:FT_1_FT}, we provide the results of the first geometric investigation. As we can see, the average ID and LID evolve almost identically for the first few iterations of the pretraining phase. After that, a clear gap between the lines gradually takes place. This result indicates that the pretraining strategy starts by learning linear correlations between the latent features and then transforms the initial flat manifolds into \textit{curved} ones. Therefore, we conclude that the Euclidean geometry is inappropriate to assess the latent similarities at the end of the pretraining phase. Without prior knowledge, it is not possible to systematically identify a non-euclidean metric that can capture the latent similarities for any dataset. In the clustering phase, we observe that the LID decreases considerably after the pretraining stage for the four cases studied in Figure \ref{fig:FT_1_FT}. This result implies that the embedded manifolds undergo substantial transformation: from curved to flattened manifolds. This transformation can bring geometric deterioration caused by twisting the curved structures while flattening the latent manifolds as we will show on a 2D synthetic dataset.



\begin{figure*}[!htbp]
\centering
\begin{subfigure}[b]{0.15\textwidth}
    \includegraphics[width=\linewidth]{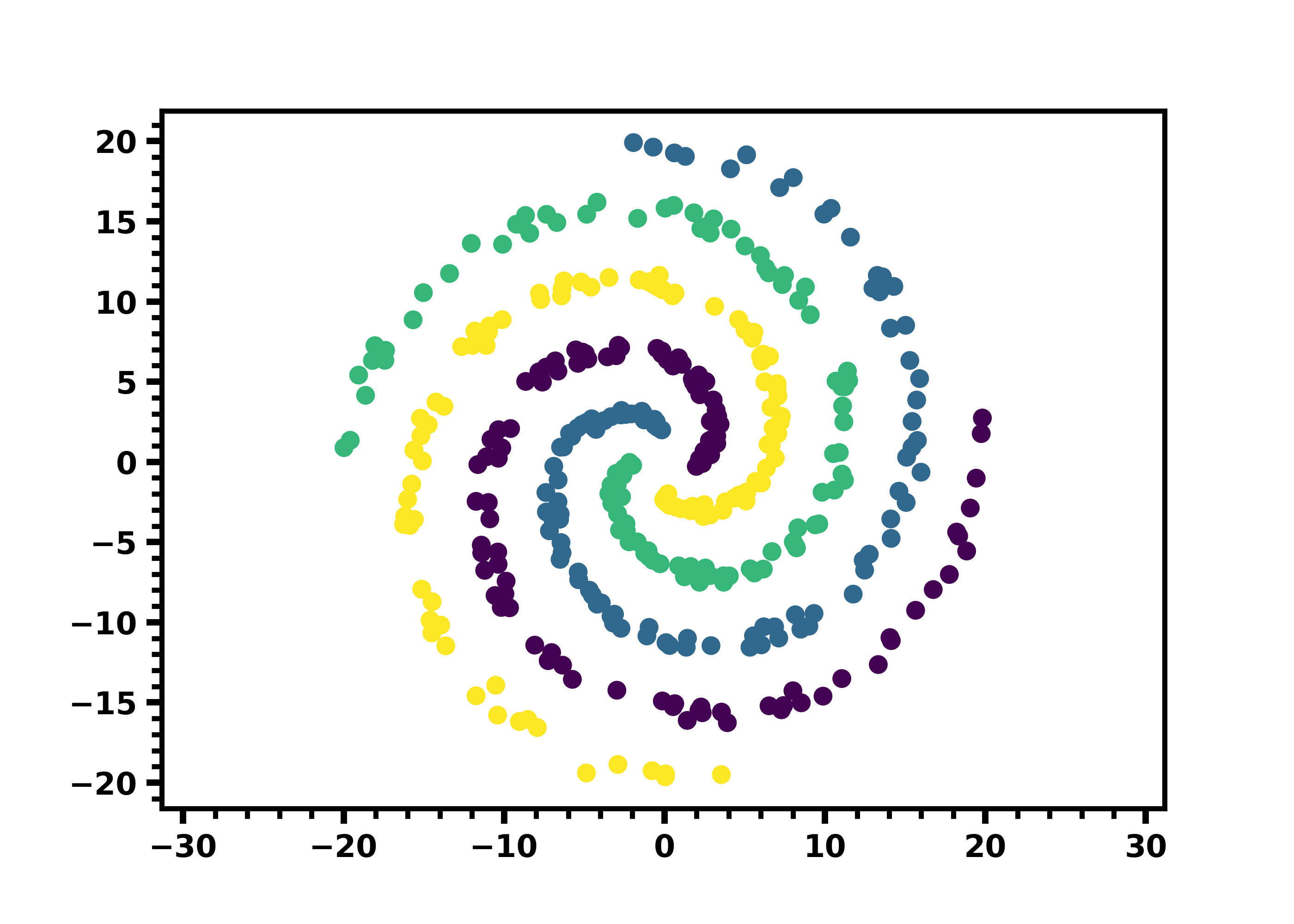}
    \caption{Input data}
\end{subfigure}
\begin{subfigure}[b]{0.15\textwidth}
    \includegraphics[width=\linewidth]{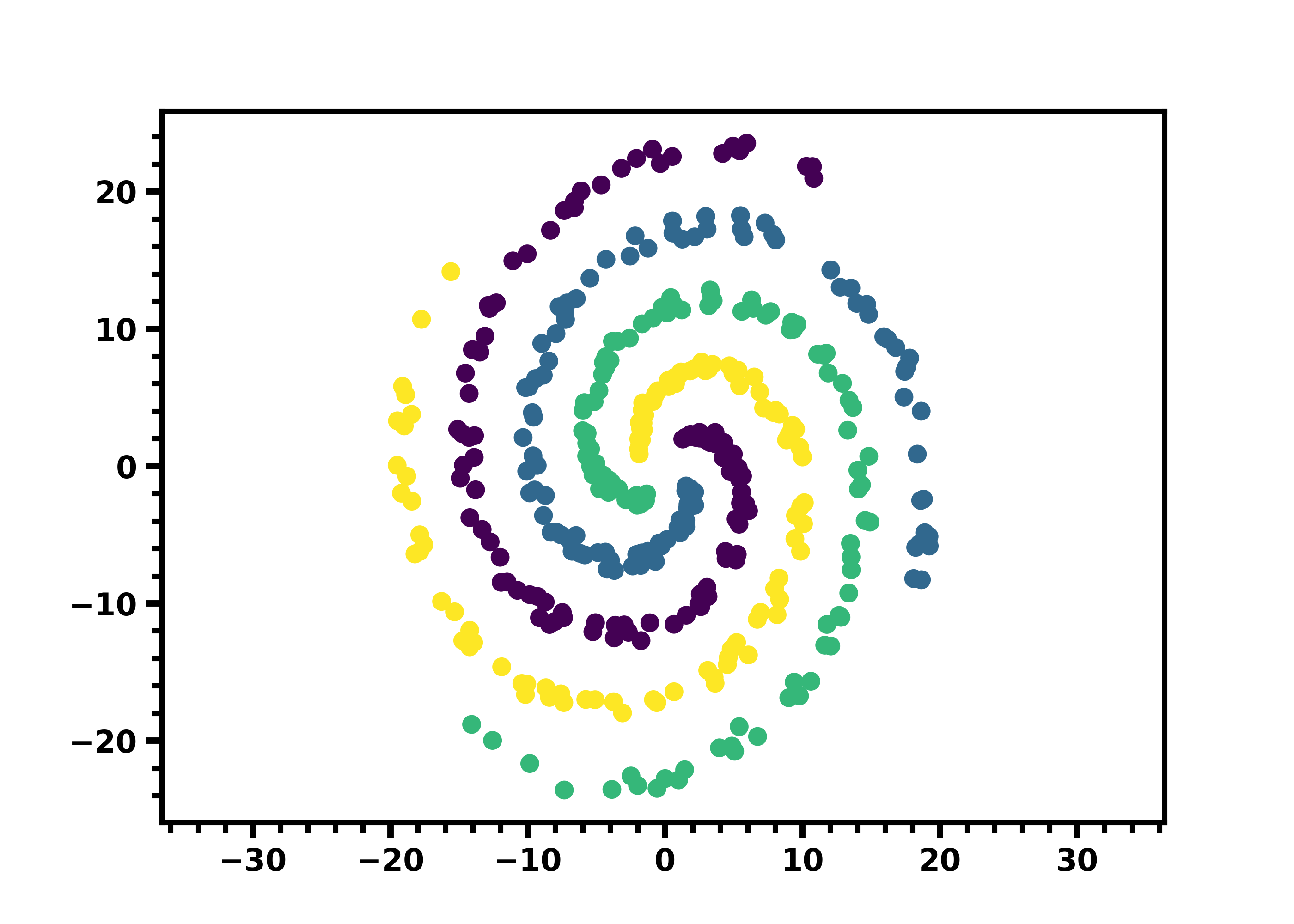}
    \caption{Epoch $200$}
\end{subfigure}
\begin{subfigure}[b]{0.15\textwidth}
    \includegraphics[width=\linewidth]{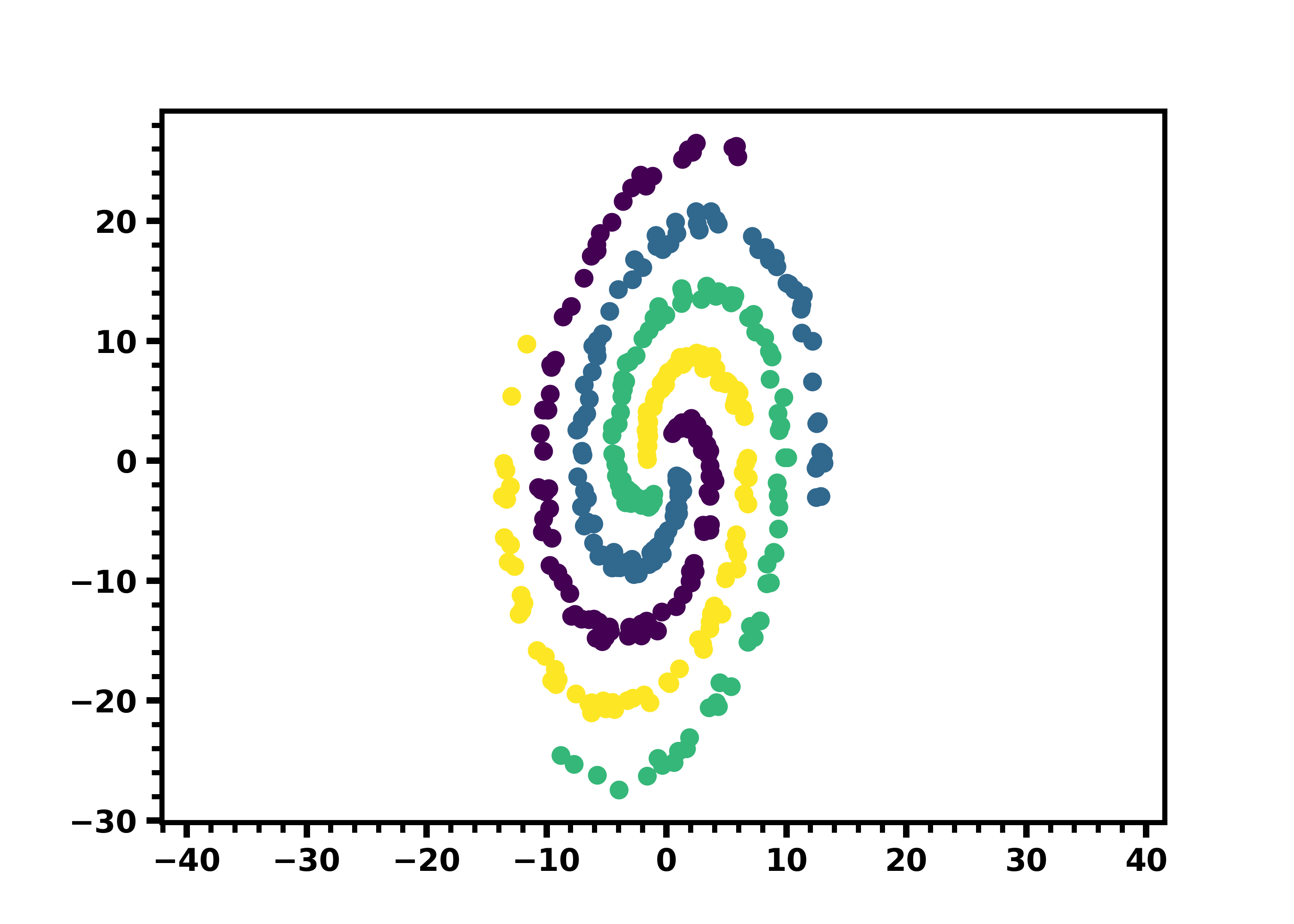}
    \caption{Epoch $225$}
\end{subfigure}
\begin{subfigure}[b]{0.15\textwidth}
    \includegraphics[width=\linewidth]{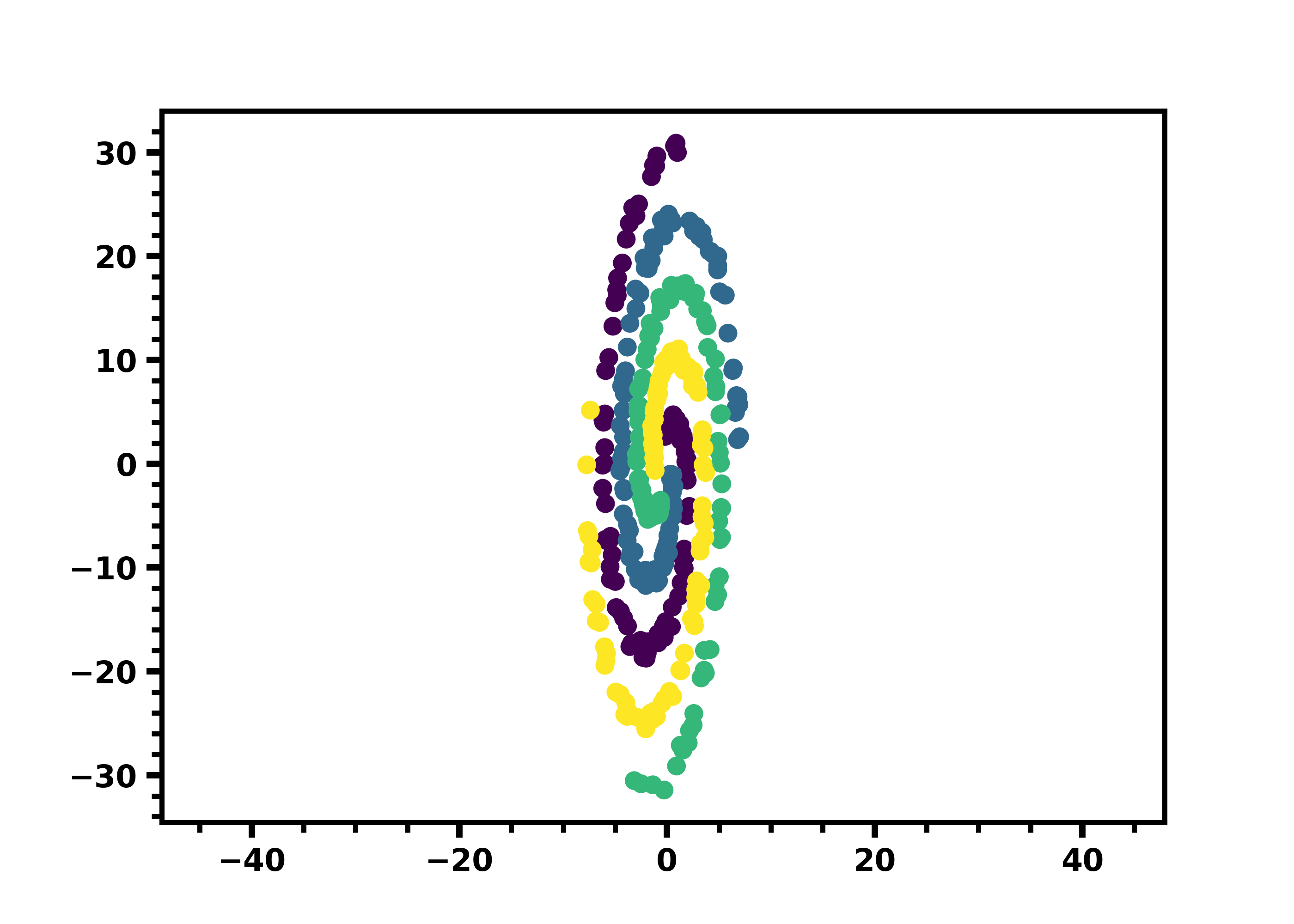}
    \caption{Epoch $250$}
\end{subfigure}
\begin{subfigure}[b]{0.15\textwidth}
    \includegraphics[width=\linewidth]{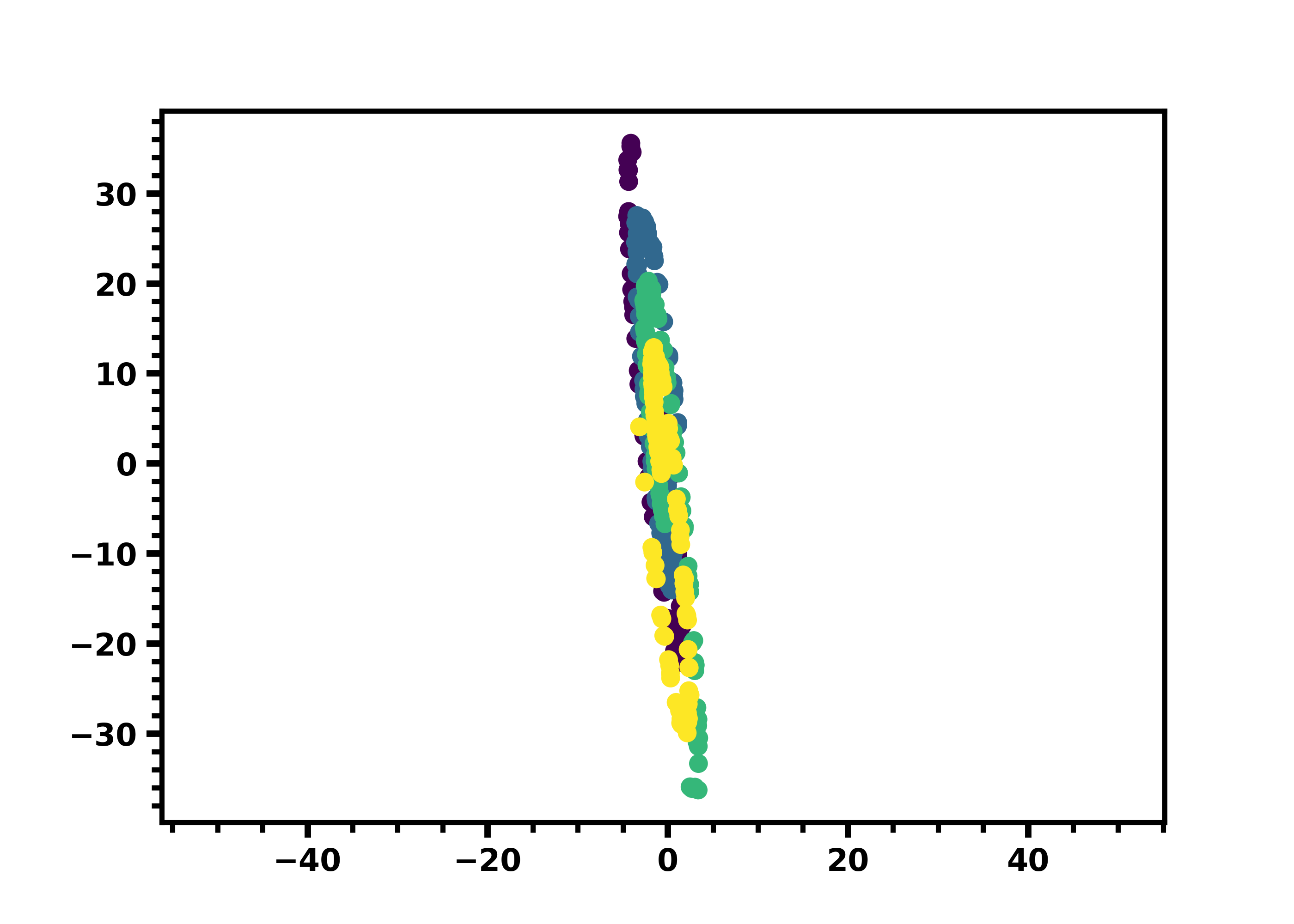}
    \caption{Epoch $275$}
\end{subfigure}
\begin{subfigure}[b]{0.15\textwidth}
    \includegraphics[width=\linewidth]{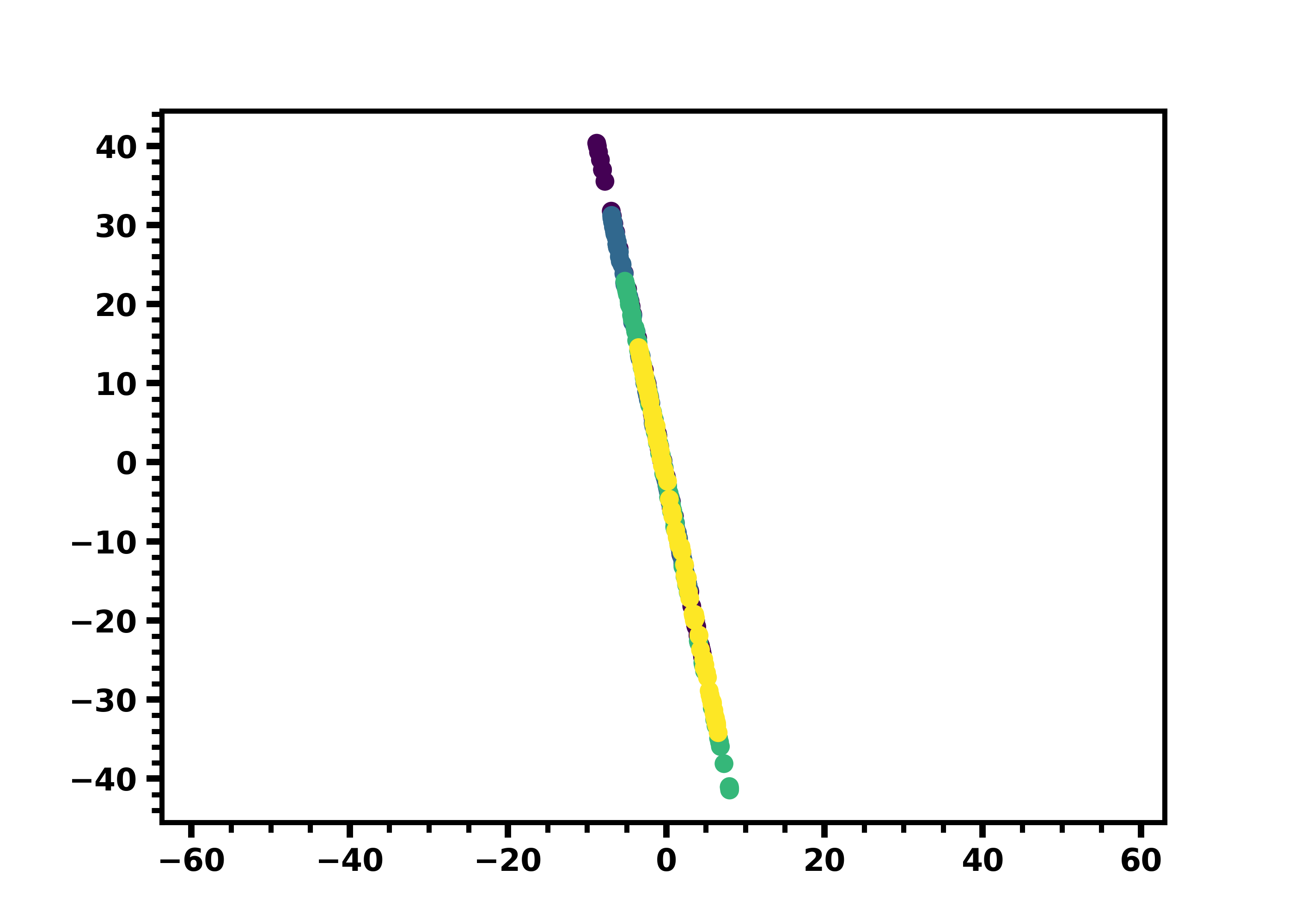}
    \caption{Epoch $300$}
\end{subfigure}
\caption{\textbf{Second evidence of Feature Twist}. Collapse of the latent structures in the clustering phase.}
\label{fig:FT_2}
\end{figure*}

Due to the limitations of visualizing the curved latent structures in high-dimensional space for the initial experiments, we conduct a second experiment using a 2D synthetic dataset with four curved clusters. In this experiment, we leverage a linear two-layer auto-encoder to map the data onto a 2D latent space. This model undergoes a pretraining phase with vanilla reconstruction for 200 epochs, followed by a fine-tuning phase with the DEC clustering objective. The resulting visualizations of the embedded space, as shown in Figure \ref{fig:FT_2}, reveal that the clustering process causes the curved manifolds to become flattened. This flattening process introduces geometric distortions, twisting the curved structures and altering the latent manifolds. This phenomenon, known as the Feature Twist problem, forces points from originally distinct clusters to merge, thereby compromising the clustering structures.

To address FT, Mrabah et al \cite{mrabahescaping} have proposed FT-VGAE, which is a Variational Graph auto-encoder model that can reduce the effect of the abrupt transition from self-supervision to pseudo-supervision. It initiates the training process by minimizing the adjacency reconstruction cost. In the second training phase, the model optimizes a different self-supervision task that smooths out the local latent curvatures while preserving the global curved structure. Lastly, the third training phase puts into action a pseudo-supervision task that minimizes the clustering loss of Deep Embedding Clustering (DEC) \cite{xie2016unsupervised}. Although FT-VGAE has shown effectiveness in alleviating the effect of FT, it is specifically tailored for graph datasets. It is not clear how to adopt the same strategy for Euclidean-based representations such as image datasets. Furthermore, this model has three training phases with a quadratic computational complexity for each one. This aspect constrains the applicability of this approach to small databases. Moreover, FT-VGAE follows the typical deep clustering paradigm, and thus it requires a pseudo-supervision task. This characteristic poses a challenge in terms of balancing the trade-off between self-supervision and pseudo-supervision when performed concurrently \cite{mrabah2021rethinking}.


\subsubsection{Feature Randomness and Feature Drift}

We examine the interplay between self-supervision and pseudo-supervision when they are applied concurrently. Previous works \cite{mrabah2020adversarial, mrabah2020deep} have shown that the typical DC paradigm is governed by the trade-off between Feature Randomness (FR) and Feature Drift (FD). The FR problem is related to the pseudo-supervision task. More precisely, the use of pseudo-labels to learn clustering-oriented features can mislead the trained network. While some of the pseudo-labels align with the ground-truth labels, some of them do not correspond to the true categories. Thus, the network can learn random features that capture irrelevant similarities. For instance, the DEC model performs pseudo-supervision without introducing an auxiliary self-supervision task during the finetuning stage. Thus, this model has no mechanism to prevent random projections caused by pseudo-supervision. To mitigate the effect of random features, several deep clustering models \cite{mrabah2020adversarial, guo2017improved, dizaji2017deep} leverage self-supervision strategies not only as a pretraining task but also as an auxiliary task during the second phase. The reconstruction is the de facto self-supervision technique for existing deep clustering models \cite{guo2017improved, dizaji2017deep, jiang2017variational}. However, the discriminative features learned by the clustering objective can be easily drifted by the self-supervision task. In other words, the FD problem is caused by the strong competition between self-supervision and pseudo-supervision. For instance, the Improved Deep Embedding Clustering model (IDEC) \cite{guo2017improved} improves DEC by performing joint clustering and vanilla reconstruction during the second phase. Yet, it suffers from FD. The clustering loss reduces the within-cluster variance and strengthens the between-cluster variance. As opposed to that, the vanilla reconstruction loss aims to preserve the within- and between-cluster variances.

To address the trade-off between FR and FD, Deep Clustering with a Dynamic Auto-Encoder (DynAE) \cite{mrabah2020deep} adjusts the self-supervision task in a progressive way. The training process of DynAE consists of two phases. During the pretraining phase, the network performs a pretext task (i.e., reconstruction with adversarially constrained interpolation \cite{Berthelot2019understanding}). In the subsequent clustering phase, the network is fine-tuned using a dynamic combination of self-supervision and pseudo-supervision. In essence, the dynamic learning mechanism consists of gradually and smoothly transforming the reconstruction loss into a clustering-oriented self-supervision task. In particular, this model constructs the decoded centroid images of the high-confidence samples (i.e., samples with low-entropy clustering assignment scores) and performs vanilla reconstruction for the low-confidence points (i.e., samples with high-entropy clustering assignment scores). Consequently, the loss function changes progressively during the clustering phase to reduce the influence of FD without causing excessive FR. While DynAE has shown promising results, it still suffers from the trade-off between FR and FD to some extent. Furthermore, this approach does not consider the abrupt transition from pretraining to clustering, which means it suffers from FT. 

\begin{figure}
    \includegraphics[width=\linewidth]{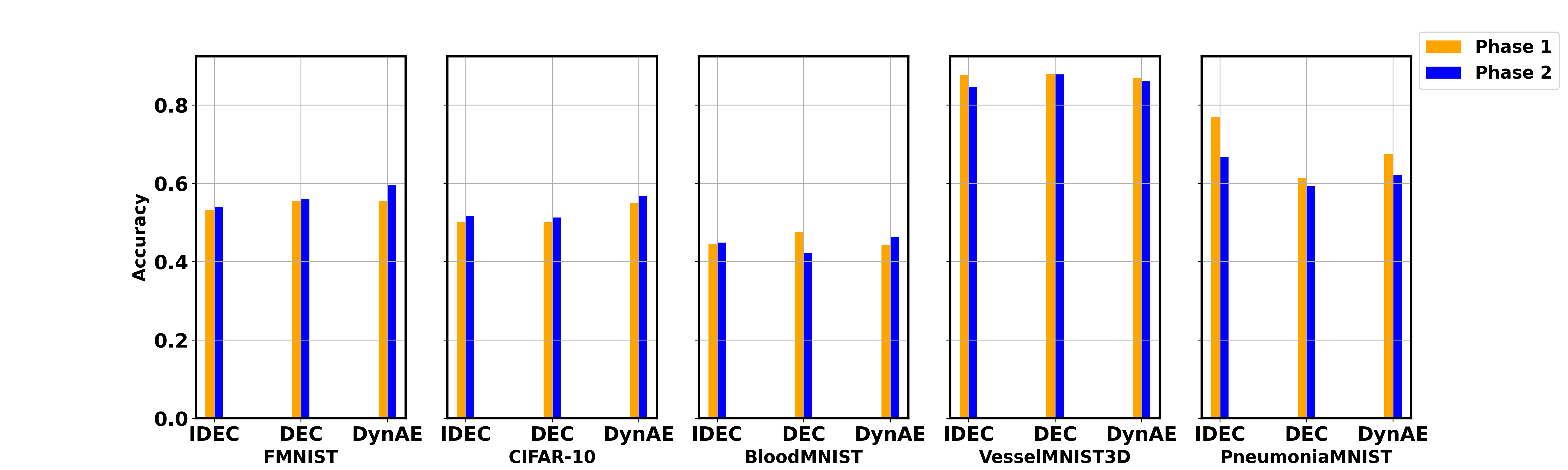}
     \caption{The Comparison in terms of Accuracy between Phase 1 (i.e., pretraining) and Phase 2 (i.e., finetuning) for DynAE, IDEC, and DEC using five Datasets.}
  \label{fig:resulta}
\end{figure}

As discussed before, the existing DC paradigms are associated with three core limitations: FR, FD, and FT. These problems control the relationship between self-supervision and pseudo-supervision whether for joint training (FR and FD) or independent training (FR and FT). First, training the model with pseudo-labels during the second phase leads to the generation of random and unreliable features. Second, combining pseudo-supervision and self-supervision during the second phase drifts a portion of the clustering-oriented features. Third, the shift from self-supervision to pseudo-supervision between the first and second phases twists the curved latent manifolds. These problems deteriorate the clustering performance. In Figure \ref{fig:resulta}, we illustrate the advancement achieved in the second phase by various state-of-the-art deep clustering methods on five datasets. All the considered models follow the typical DC paradigm. Principally, we shed light on the progress achieved by the clustering phase compared with the pretraining phase. As we can see, most of the improvement comes from the self-supervision pre-training phase. Notably, the enhancement attributed to the second phase does not surpass a $5\%$ increase in clustering performance. These results put into question the significance of the pseudo-supervision task. By replacing pseudo-supervision with another relevant task, it might be possible to prevent FR, FD, and FT.

\subsection{Rethinking DC Paradigms}

To address the limitations of existing DC paradigms concerning FR, FD, and FT, we introduce a new paradigm. We argue that pseudo-supervision is not required in deep clustering to achieve high clustering performance. Principally, the new paradigm consists of dispensing with the clustering-level pseudo-supervision task, which is inherently prone to high error rates and can not easily align with instance-level self-supervision tasks during both joint and independent training. In particular, we replace pseudo-supervision with a second round of self-supervision training. The second self-supervision task should ensure a smooth transition from the first to the second phase and should be more relevant to the clustering task than the first self-supervision task. More precisely, 
we initially pretrain a neural network based on instance-level self-supervision. After that, we finetune the model based on neighborhood-level self-supervision. Finally, we apply a traditional clustering method (e.g., k-means) on the latent codes obtained after the second phase.


The proposed DC paradigm has three advantages over the up-to-date framework. First, we discuss the impact of our paradigm with respect to FT. Our empirical findings offer compelling evidence that neighborhood-level self-supervised training preserves the global structure of the latent manifolds constructed based on instance-level pretraining. The transition between instance-level and neighborhood-level self-supervision is smoother and less abrupt than the transition between instance-level self-supervision and clustering-level pseudo-supervision. Second, we discuss the impact of our paradigm with respect to FR. Eliminating the clustering-level pseudo-supervision task prevents the risk of generating random features, which are caused by the error-prone nature of the pseudo-labels. Last but not least, the proposed paradigm circumvents the FD challenge. Getting rid of pseudo-supervision relinquishes the need to perform joint clustering and self-supervised training. Particularly, the proposed paradigm prevents the drifting effect caused by the strong competition between two tasks: instance-level self-supervision and clustering-level pseudo-supervision. 


According to the new paradigm, we propose a novel approach R-DC that follows a two-step strategy. First, we pretrain an auto-encoder model based on adversarially constrained interpolation \cite{Berthelot2019understanding}, which is an instance-level self-supervision loss. After the pretraining phase, we finetune the model using a proximity-level technique as a substitute for pseudo-supervision. We leverage a dual filtering mechanism to select the core points (i.e., points located in dense regions of the latent space) and the most reliable neighbors of these core points. Then, we compute the latent space centroids of these reliable nearest neighbors for each core point. Our objective function consists of two distinct components. For the first term, we train the auto-encoder to map the core samples to the decoded centroids of their most reliable nearest neighbors. As for the border points (i.e., points that are not core points), they undergo a vanilla reconstruction process. For the second term, we minimize a latent space loss function that pushes the embedding points toward the centroids of their most reliable nearest neighbors.

\textbf{\large{Contributions.}} \textbf{\large{(i)}} We propose a novel deep clustering paradigm that substitutes the clustering-level pseudo-supervision with proximity-level self-supervision. Eliminating pseudo-supervision prevents Feature Randomness and Feature Drift from taking place. Furthermore, the new paradigm alleviates the Feature Twist problem. In particular, the transition between instance-level and neighborhood-level self-supervision is smoother than the transition between instance-level self-supervision and clustering-level pseudo-supervision. \textbf{\large{(ii)}} We introduce a deep clustering approach that follows the new deep clustering paradigm. Our approach leverages dual-step filtering to select the core points and the most reliable neighbors of these core points. Our filtering mechanism ensures a smooth transition from vanilla reconstruction representing instance-level self-supervision to a nearest-neighbor centroid construction task representing proximity-level self-supervision. \textbf{\large{(iii)}} We conduct extensive experiments to show the merits of the proposed approach and paradigm. The obtained results provide compelling evidence that our model can significantly improve clustering performance and ensure a smooth geometric transition under the transition regime from instance-level to proximity-level self-supervision.

\section{Related Work}


Deep clustering methods can be classified into three main paradigms based on the interplay between self-supervision and pseudo-supervision. Each of these paradigms has its limitations when it comes to FR, FD, and FT. In this section, we critically examine the current DC approaches and shed light on the challenges they face regarding the randomness of learned features, potential manifold distortions caused by the transition from self-supervision to pseudo-supervision, and the drifting effect caused by the competition between self-supervision and pseudo-supervision.



\subsection{First Deep Clustering Paradigm}

\begin{figure}
    \includegraphics[width=\linewidth]{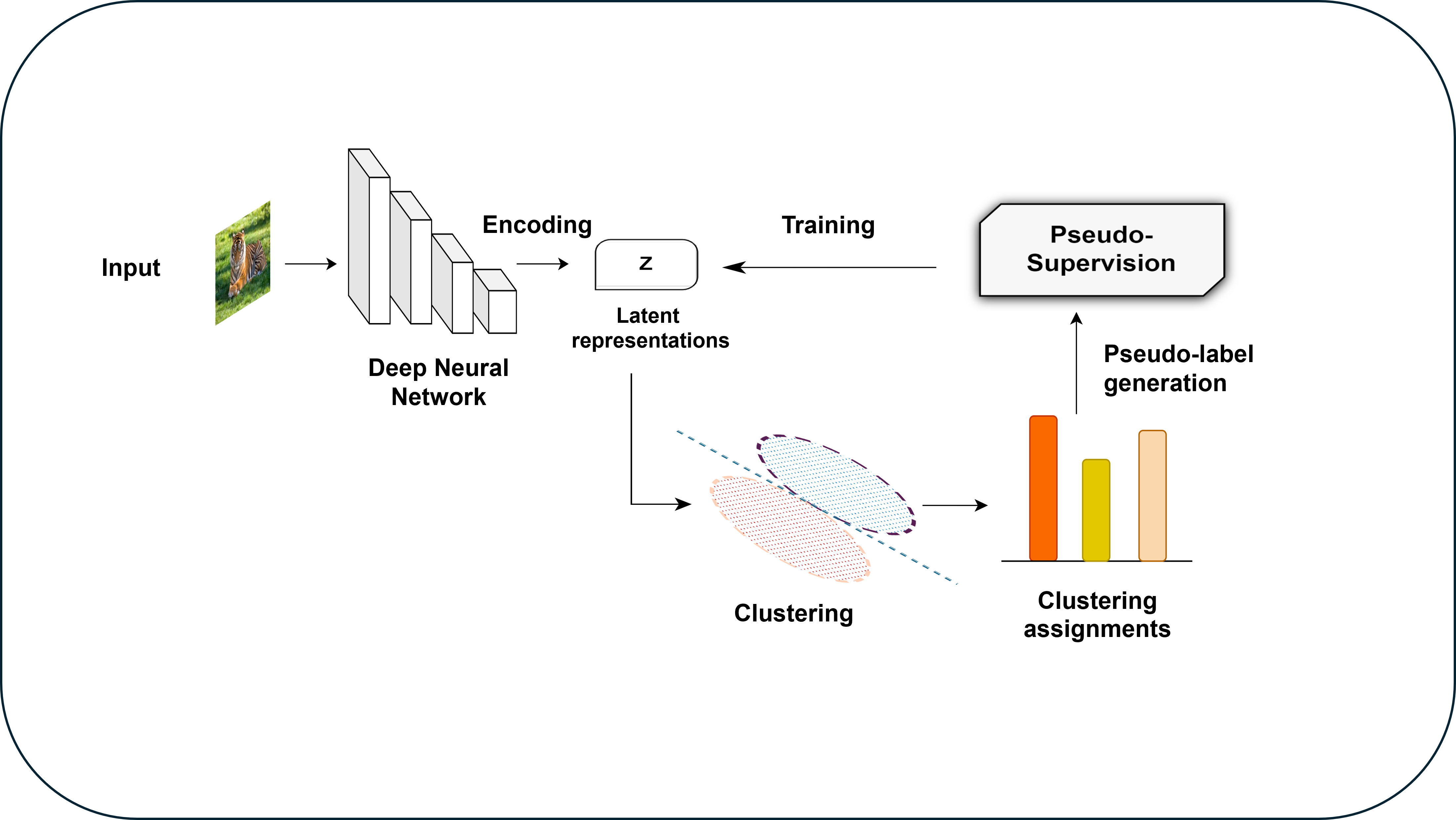}
     \caption{\textbf{The first deep clustering paradigm.} This paradigm does not require self-supervised training. It only uses pseudo-supervision to train a deep neural network.}
  \label{fig:pi2}
\end{figure}

The first deep clustering paradigm does not require self-supervised training. Instead, this paradigm uses pseudo-supervision to train a deep neural network and leverages the pseudo-labels to cluster the latent representations. Figure \ref{fig:pi2} summarizes the process of the first deep clustering paradigm. The process involves feeding images into the network, generating embeddings, and applying a clustering algorithm to form latent clusters. The pseudo-labels are assigned based on the clustering assignments. However, the use of pseudo-supervision without any protection or correction mechanism leads to 
the generation of excessive random features caused by the uncertainty inherent to the pseudo-labels. Thus, the first paradigm presents significant challenges in dealing with FR.


The work of Yang et al.\cite{yang2016joint} presents a Joint Unsupervised Learning (JULE) framework for Image Clustering. JULE follows the first DC paradigm. This approach does not require any self-supervised pertaining. Instead, it operates by directly generating pseudo-labels from the input images. More precisely, JULE harnesses a recurrent framework, where agglomerative clustering operations are integrated into a recurrent process with deep representations produced by a Convolutional Neural Network (CNN). Although this approach outperforms some state-of-the-art methods on various image clustering tasks, it has some limitations from the perspective of FR. The latent features used to generate the pseudo-labels based on agglomerative clustering do not necessarily represent the intrinsic clustering structures, which can lead to inaccurate pseudo-labels and the generation of random features.

Maximizing the mutual information between representations of data samples is a widely used clustering measure for several recent approaches. Invariant Information Clustering (IIC) \cite{ji2019invariant} is also a clustering method that follows the first DC paradigm. This approach works through an iterative process to calculate mutual information between the clustering assignments of paired data samples. The absence of a self-supervised pretraining phase eliminates the FT problem. Furthermore, the absence of joint pseudo-supervision and self-supervision eliminates the FD problem. However, the direct application of pseudo-supervision in IIC could probably hinder its ability to learn latent features with enough discriminative power to accurately unveil the true clustering structures. This limitation might lead to excessive FR, affecting the overall effectiveness of the clustering process.



PartItion Confidence mAximisation (PICA) \cite{huang2020deep} is also a direct pseudo-supervision clustering approach that follows the first DC paradigm. Unlike the previously discussed methods (i.e., JULE and IIC), PICA has a protection mechanism to alleviate the effect of FR. In particular, this method relies on the most confident clustering assignments for the pseudo-supervision task. More precisely, PICA focuses on learning robust latent representations of the data by minimizing a 
differentiable partition uncertainty index (PUI). This model has the advantage of not requiring two training phases, ensuring the avoidance of FT. Moreover, the absence of joint self-supervision and pseudo-supervision circumvents the occurrence of FD. However, PICA remains prone to FR due to the absence of self-supervised training that can counteract the generation of random features.

\subsection{Second Deep Clustering Paradigm}
\begin{figure}[!h]
    \includegraphics[width=\linewidth]{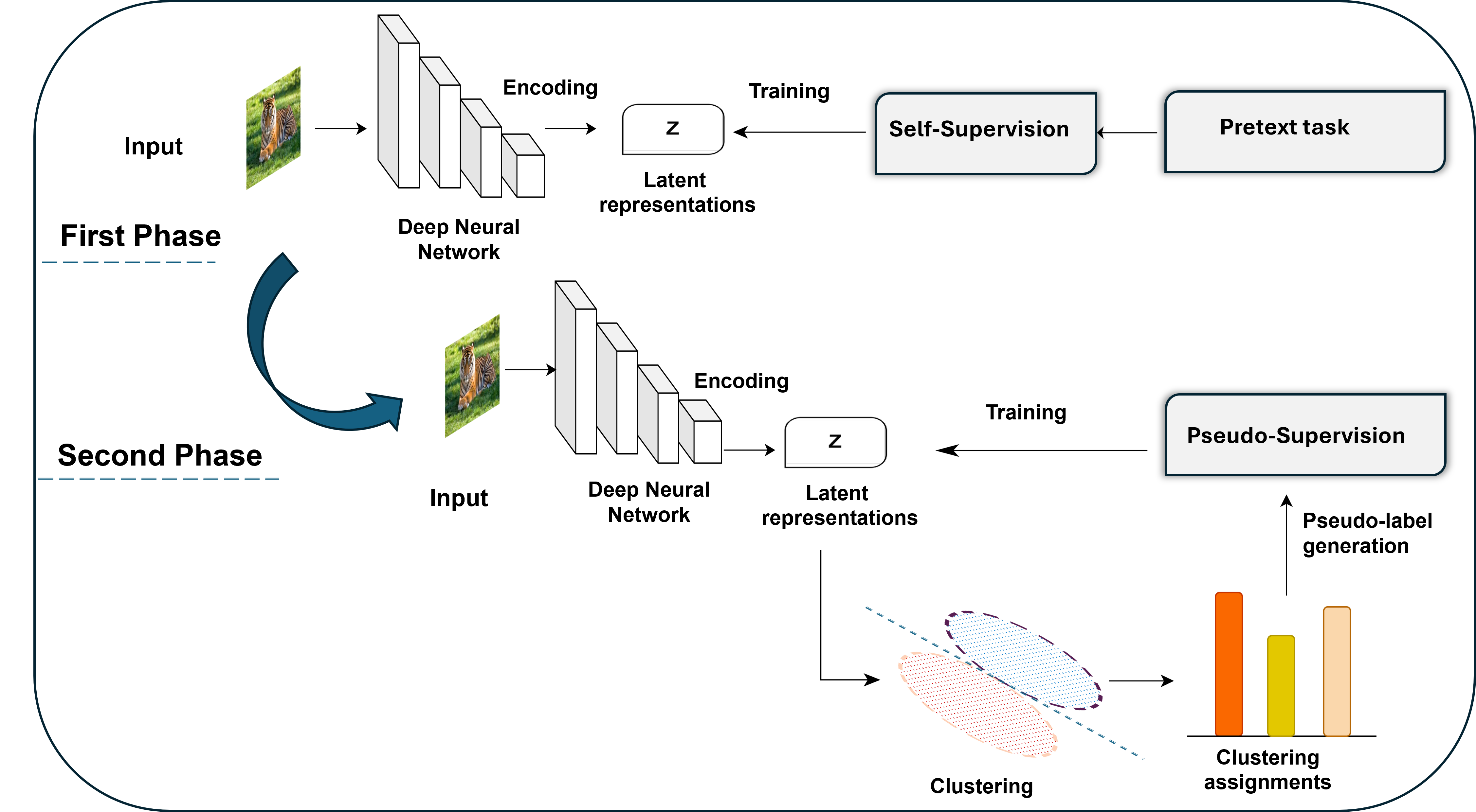}
     \caption{\textbf{The second deep clustering paradigm.} This paradigm involves two phases: pretraining and fine-tuning. Initially, a deep neural network is trained based on self-supervised learning. Then, the network is finetuned based on pseudo-supervision.}
  \label{fig:Picture2-deep}
\end{figure}

The second DC paradigm involves two phases, pretraining and fine-tuning. Figure \ref{fig:Picture2-deep} presents the phases of the second paradigm. Initially, a deep neural network is trained based on a self-supervision task to learn general-purpose and high-semantic features from the input data. Then, the network is further optimized using a clustering algorithm and pseudo-labels generated from the learned representations. This paradigm offers advantages over direct clustering methods as the pretraining phase leads to less FR. However, there are some potential drawbacks to this paradigm. First, the self-supervised task may not be fully aligned with the clustering task. This can introduce potential FR. Second, the transition between the pretraining and fine-tuning phases can cause FT. Both problems can affect the effectiveness of clustering.

The task of learning self-supervised representations is tackled in several DC approaches. Xie et al.\cite{xie2016unsupervised} have proposed a Deep Embedded Clustering (DEC) method. This approach is designed to learn feature representations and cluster assignments simultaneously. DEC performs a pretraining phase based on vanilla reconstruction. The reconstruction task is performed using a stacked auto-encoding process. Then, this approach finetunes the encoder based on pseudo-supervision. This is achieved by iteratively minimizing a Kullback-Leibler (KL) divergence between a clustering distribution and a refined target clustering distribution obtained by emphasizing the high-confidence assignments. As an advantage over the methods that follow the first DC paradigm, DEC is pretrained before performing embedding clustering. This reduces the problem of FR. From another angle, relying solely on pseudo-supervision in the finetuning phase still presents a potential risk for generating random pseudo-labels. Moreover, FT can take place during the transition between pretraining and fine-tuning. Another approach is Leveraging Tensor Kernels to Reduce Objective Function Mismatch in Deep Clustering (DDC-UCO) \cite{trosten2024leveraging}, which optimizes companion objectives to reduce the mismatch between pseudo-supervision and the auxiliary objective while maintaining the structure-preserving property of the latter. DDC-UCO establishes a better trade-off between FR and FD compared with several previous works, as shown by the empirical results of \cite{trosten2024leveraging}. However, because of the transition between the pretraining and finetuning phases, FT can take place. 

Several traditional clustering methods use direct distances between data points to form groups \cite{sarfraz2019efficient}. However, in a high-dimensional space, these distances may become less informative and fail to accurately capture the relationships between data points. Furthermore, several traditional clustering methods need thresholds or hyperparameters. FINCH \cite{sarfraz2019efficient} is a clustering algorithm that calculates the first neighbor relations between all data points and uses them to construct a hierarchical clustering. Moreover, it does not require any thresholds or hyperparameters. In the context of DC, Deep FINCH \cite{sarfraz2019efficient} adheres to the second DC paradigm. It initiates with a pretraining phase grounded in the basic reconstruction process. In this context, the pretraining phase is instrumental in reducing the risk of FR. Then, this approach finetunes the encoder parameters by performing joint clustering and feature learning. In particular, deep FINCH minimizes their proposed hierarchical clustering loss on the latent codes as a pseudo-supervision task. However, it is important to note that relying solely on pseudo-supervision in the finetuning phase may not be the best choice due to the potential FR risk. In addition, the transition between the self-supervised and pseudo-supervised phases can cause FT.


\subsection{Third Deep Clustering Paradigm}
\begin{figure}
    \includegraphics[width=\linewidth]{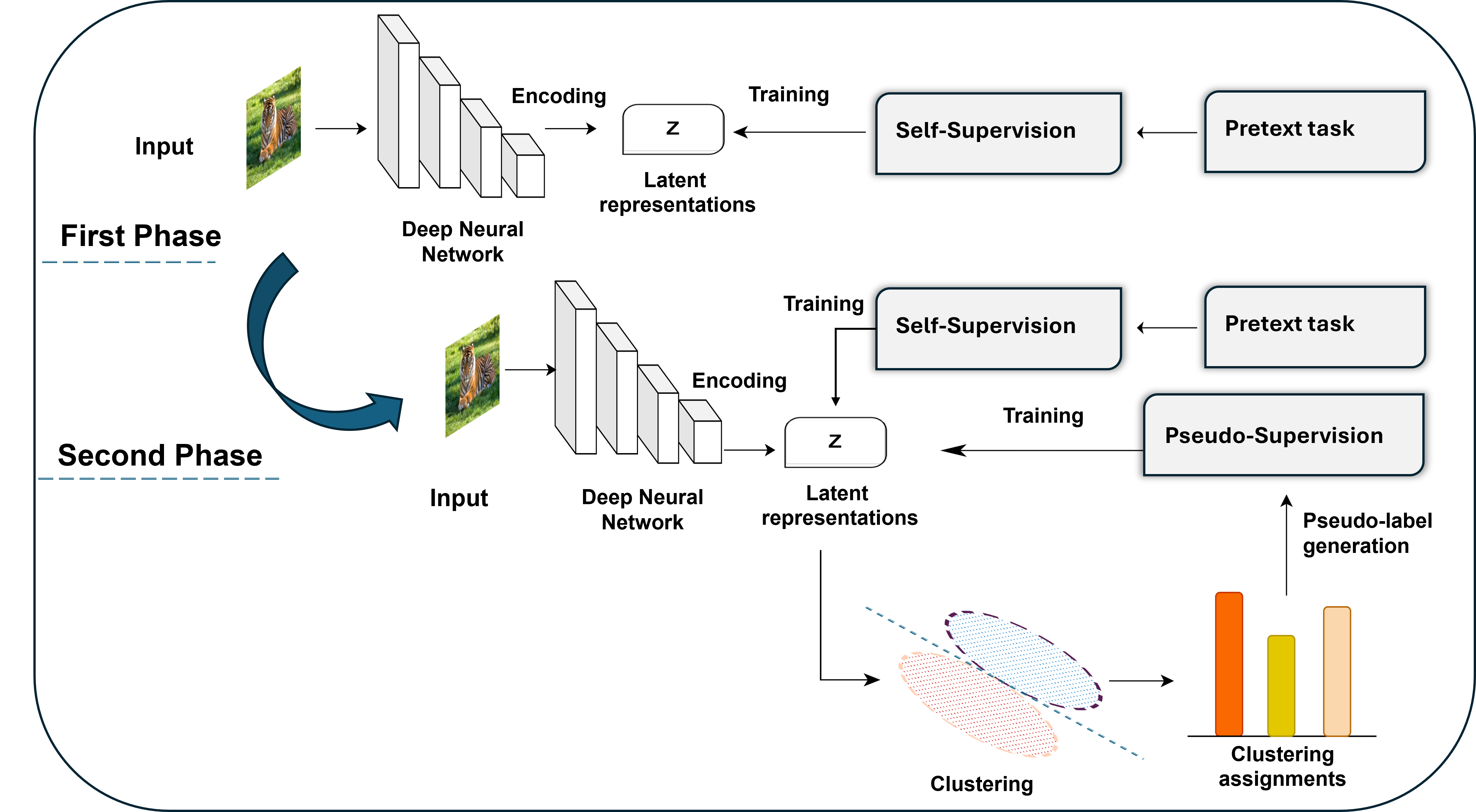}
     \caption{\textbf{The third deep clustering paradigm.} This paradigm involves two phases: pretraining and fine-tuning. Initially, a deep neural network is trained based on self-supervised learning. Then, the network is finetuned based on joint pseudo-supervision and self-supervision.}
  \label{fig:Picture3-deep}
\end{figure}

Various deep clustering approaches have embraced a two-stage training strategy. Similar to the second DC paradigm, the third paradigm leverages a two-stage training process. However, the main difference between them lies in the fine-tuning stage. While the second paradigm relies solely on pseudo-supervision, the third one performs joint pseudo-supervision and self-supervision. As presented in Figure \ref{fig:Picture3-deep}, the first step of the third paradigm is pretraining the models using self-supervision. Then, fine-tuning is performed by employing a combination of pseudo-supervision and self-supervision. Compared with the first and second paradigms, the third one has self-supervised training in the first and second phases, which can mitigate and protect against the effect of FR. Nonetheless, introducing self-supervision in the latter stage can lead to the occurrence of FD. Furthermore, the transition between the training phase and the clustering phase can introduce a drawback of FT.

The utilization of the reconstruction loss as a self-supervision task stands as a fundamental technique within the training frameworks of several DC methods. Improved Deep Embedded Clustering (IDEC) \cite{guo2017improved}, Embedded RegularIzed ClusTering (DEPICT) \cite{ghasedi2017deep}, Variational Deep Embedding (VaDE) \cite{jiang2017variational}, Deep Clustering Network (DCN) \cite{yang2017towards} and Pseudo-Supervised Deep Subspace Clustering (PSSC) \cite{lv2021pseudo} are methods with a two-phase training process, and employ the reconstruction loss as a self-supervision mechanism. For instance, IDEC jointly performs clustering and learns representative features with local structure preservation. Whereas VaDE combines a variational auto-encoder and Gaussian mixture (GM) to learn representations of data points that are discriminative for clustering. DEPICT uses a denoising auto-encoder and a softmax layer stacked on top of a multi-layer convolutional auto-encoder. Additionally, DCN performs joint reconstruction and embedding K-means clustering to learn the latent representations of the data. 
PSSC employs reconstruction loss as a self-supervision mechanism during the pre-training phase. Furthermore, it performs a second training phase using a reconstruction module, a self-expression module, and a pseudo-supervision module. In PSSC, the FR, FD, and FT problems remain notable concerns. Empirical results from these previous works provide strong evidence that combining self-supervision and pseudo-supervision during the fine-tuning stage brings clear improvement in clustering performance compared with solely performing pseudo-supervision. This improvement has been attributed to the effect of self-supervision in mitigating FR \cite{mrabah2020deep}. 
However, for all these methods (i.e., IDEC, DEPICT, VaDE, PSSC, and DCN), the FD and FT problems remain notable concerns. 

Some other approaches follow the third DC paradigm and leverage instance-level contrastive learning as a self-supervision task. For instance, CC \cite{li2021contrastive} combines instance-level contrastive learning as a self-supervision task and cluster-level contrastive learning as a pseudo-supervision task. This approach yields promising results thanks to the data augmentation strategies. Yet, it does not tackle the trade-off between FR and FD. In another work, Self Labelling (SeLa) \cite{asano2020selflabelling} maximizes a contrastive mutual information loss between the labels and input data, leading to an optimal transport problem that they efficiently solve with a fast variant of the Sinkhorn-Knopp algorithm. Meanwhile, a Semantic Pseudo-labeling framework for Image ClustEring (SPICE) \cite{niu2022spice} introduces a clustering network with a feature model for capturing instance-level similarity and a clustering head for capturing cluster-level discrepancy. In addition to that, SPICE employs two pseudo-labeling algorithms: prototype pseudo-labeling and reliable pseudo-labeling. The use of instance-level contrastive self-supervision in these methods gives an advantage of less FR. However, these approaches do not have any mechanism to address FD and FT.



To address the FD problem associated with the third DC paradigm, Adversarial Deep Embedded Clustering (ADEC) uses adversarial training to transfer the competition between pseudo-supervision and self-supervision beyond a single network. In particular, ADEC introduces a discriminator network and trains it to identify a better trade-off between FR and FD. To reduce FR, ADEC penalizes the generation of embedded features, which could not be decoded into realistic data points, using the Generative Adversarial Network loss \cite{mrabah2020adversarial}. To reduce FD, ADEC restrains the back-propagation of the reconstruction loss to the decoder layers. In another work, Mrabah et al. \cite{mrabah2020deep} have proposed a Dynamic auto-encoder model (DynAE) that gradually eliminates the reconstruction objective function in favor of a clustering-oriented embedding clustering loss to mitigate the risk of FD. Despite achieving a balance in the trade-off between FR and FD in ADEC and DynAE, the transition shift from self-supervision to pseudo-supervision raises concerns about the potential occurrence of FT in the latent manifolds.

\begin{figure}
    \includegraphics[width=\linewidth]{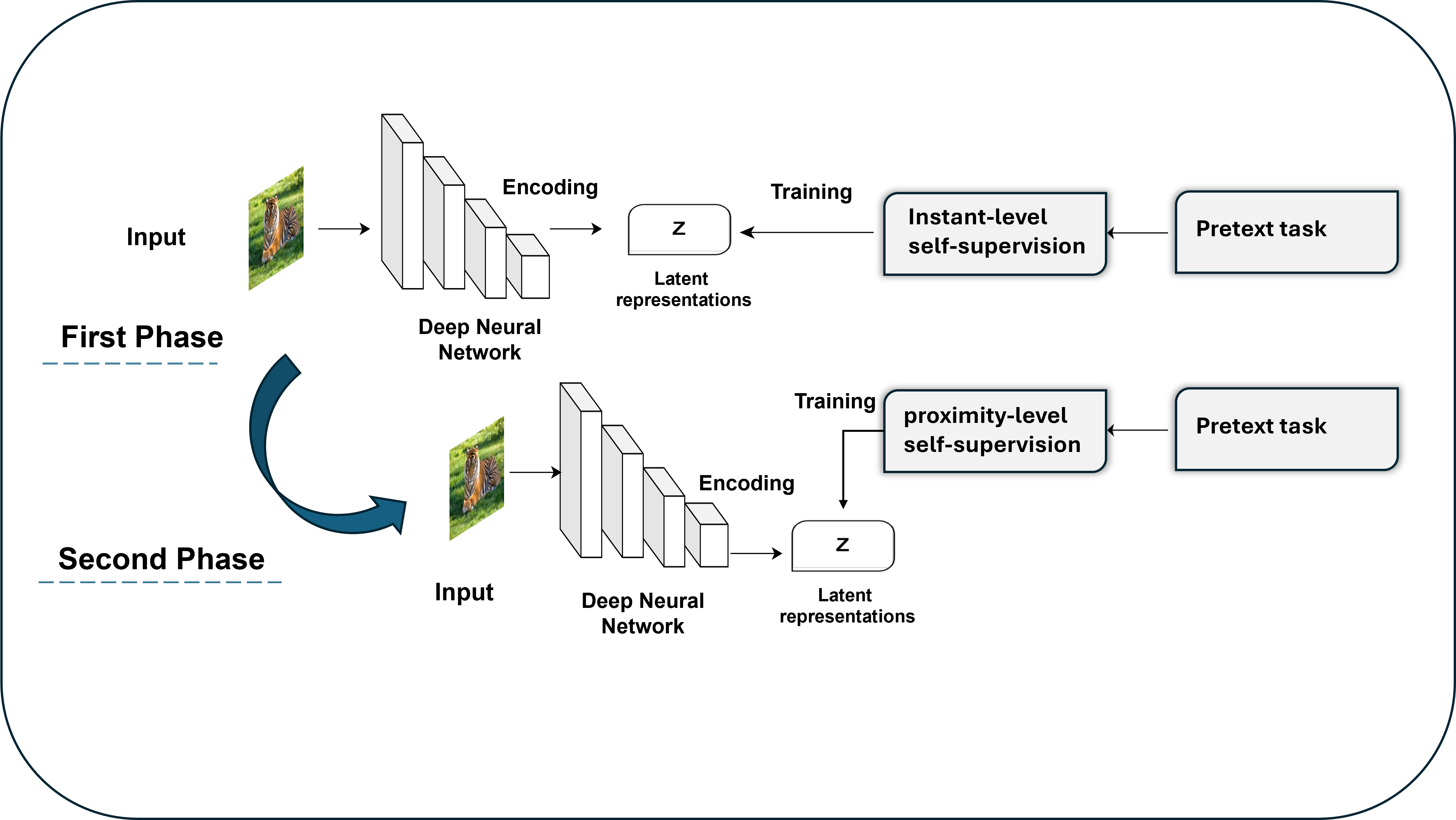}
     \caption{\textbf{The new deep clustering paradigm.} This paradigm involves two phases: pretraining and fine-tuning. Initially, a deep neural network is trained based on instance-level self-supervised learning. Then, the network is finetuned based on proximity-level self-supervision. No pseudo-supervision is required.}
  \label{fig:new-deep}
\end{figure}

\subsection{New Deep Clustering Paradigm}

Based on a critical analysis of the existing DC paradigms, it is clear that a new paradigm is needed with a different strategy that considers the three problems at the same time: FR, FD, and FT. In this context, we introduce a new DC paradigm that involves a dual-phase self-supervised training process. In other words, our paradigm completely abandons the pseudo-supervised training, positing that its drawbacks outweigh its advantages. Empirical results, presented in the introduction section, show that the improvement imputed to pseudo-supervision is small compared with the improvement attributed to self-supervised training. In this perspective, we propose to substitute pseudo-supervision with another level of self-supervised training, which is more aligned with the clustering task than the instance-level self-supervision tasks. As presented in Figure \ref{fig:new-deep}, the new paradigm does not rely on pseudo-supervision. The first phase consists of performing instance-level self-supervision. The second phase consists of performing proximity-level self-supervision. In the second self-supervision task, the model learns embedded features that are more focused on clustering compared to the first self-supervised task.

As opposed to previous paradigms, the new strategy offers multiple advantages in terms of FR, FT, and FD. The absence of the clustering-level pseudo-supervision task effectively prevents the risk of having FR. From the same angle, eliminating pseudo-supervision eliminates the inherent competition between embedding clustering and instance-level self-supervision. Thus, our paradigm does not suffer from FD. Furthermore, the transition from instance-level to proximity-level self-supervision occurs more smoothly from a geometric perspective compared to the shift from instance-level self-supervision to clustering-level pseudo-supervision. As a result, the occurrence of FT is reduced. The new paradigm is a more effective way to perform deep clustering. It addresses the FR, FD, and FT in a principled way.


\section{Proposed Method}

In this section, we introduce a novel model, which is conceived based on a Rethinking of the existing Deep Clustering paradigms (R-DC). In Subsection \ref{firstphase}, we elaborate on the first phase of our approach. In Subsection \ref{secondphase}, we describe the second phase, which is Proximity-Level Self-Supervision. Then, we explain the Nearest-Neighbor Centroid Construction and Proximity-Level Encoding in Subsections \ref{NNCC} and \ref{PLE}, respectively. Finally, the Algorithm and Optimization in subsection \ref{AO}. As stated before, using completely random pseudo-labels as training data can lead to inadequate model performance. To avoid this, neural networks are typically pretrained on pretext tasks that help them learn valuable information about the data. Based on that, we use a self-supervised loss function that consists of reconstruction regularized with adversarially constrained interpolation and data augmentation. Moreover, we introduce a filtering mechanism that facilitates a transition from instance-level self-supervision to proximity-level self-supervision, ensuring a smoother and more gradual progression. Our approach has shown competitive performance when compared to state-of-the-art methods. This improvement is attributed to our contributions.

Before we describe our methodology, we establish some useful notations. We are given a dataset $X$ of $N$ data points, where, $X$ = $\{ x_i \in \mathbb{R}^d \}_{i=1}^N$. We want to cluster these data points into $K$ clusters denoted as $\{ C_k \}_{k=1}^K$. In this context, we have two key functions: \(f_{\omega_e} \) and \(g_{\omega_d}\), where \(f_{\omega_e}\) represents the encoder, \(g_{\omega_d}\) represents the decoder, 
and $\omega_e$ and $\omega_d$ are the learnable parameters associated with the encoder and decoder, respectively. When we apply the encoder to a data point $x_{i}$, it transforms it into a latent representation $z_i = f_{\omega_e}(x_i) \in \mathbb{R}^p$, and the decoder takes this latent representation and reconstructs it into $\hat{x}_i = g_{\omega_d}(z_i) \in \mathbb{R}^d$. 
For proximity-level self-supervised learning, we denote by $\mathcal{NN}(z_{i}, k)$ the $k^{\text{th}}$ nearest neighbor of the latent code $z_i$. Furthermore, we fix the considered number of neighbors to $M$. 

The proposed paradigm utilizes a two-phase self-supervision strategy. Initially, it performs instance-level self-supervision, which captures high-level features and general-purpose representations of the data. This is followed by proximity-level self-supervision, which ensures a smoother and less abrupt transition compared to the typical switch from instance-level self-supervision to clustering-level pseudo-supervision. This smooth transition helps preserve the structure of the latent manifolds and prevents the twisting of the curved structures. By eliminating pseudo-supervision, the new paradigm avoids the risk of generating random features that are inherent in pseudo-labels. Pseudo-labels can be error-prone, leading to unreliable features. Instead, proximity-level self-supervision leverages the local structure of the data, ensuring that the learned features are more reliable and representative of the actual data distribution. In the absence of pseudo-supervision, there is no competition between clustering objectives and self-supervised tasks. This competition often causes the features to drift away from their clustering-oriented representations. The proposed paradigm's dual-stage self-supervision avoids this conflict, thereby maintaining the integrity of the learned features throughout the training process.



\subsection{First Phase: Instance-Level Self-Supervison} \label{firstphase}

Similar to previous auto-encoder-based clustering methods, R-DC first pretrains the auto-encoder to optimize a pretext objective function. Subsequently, the weights are fine-tuned based on a second objective function. Prior studies in deep clustering, such as \cite{mrabah2020deep} and \cite{mrabahescaping}, have confirmed the effectiveness of pretraining with a self-supervised objective function in capturing intrinsic features associated with the data distribution. 




\begin{figure}
    \includegraphics[width=\linewidth]{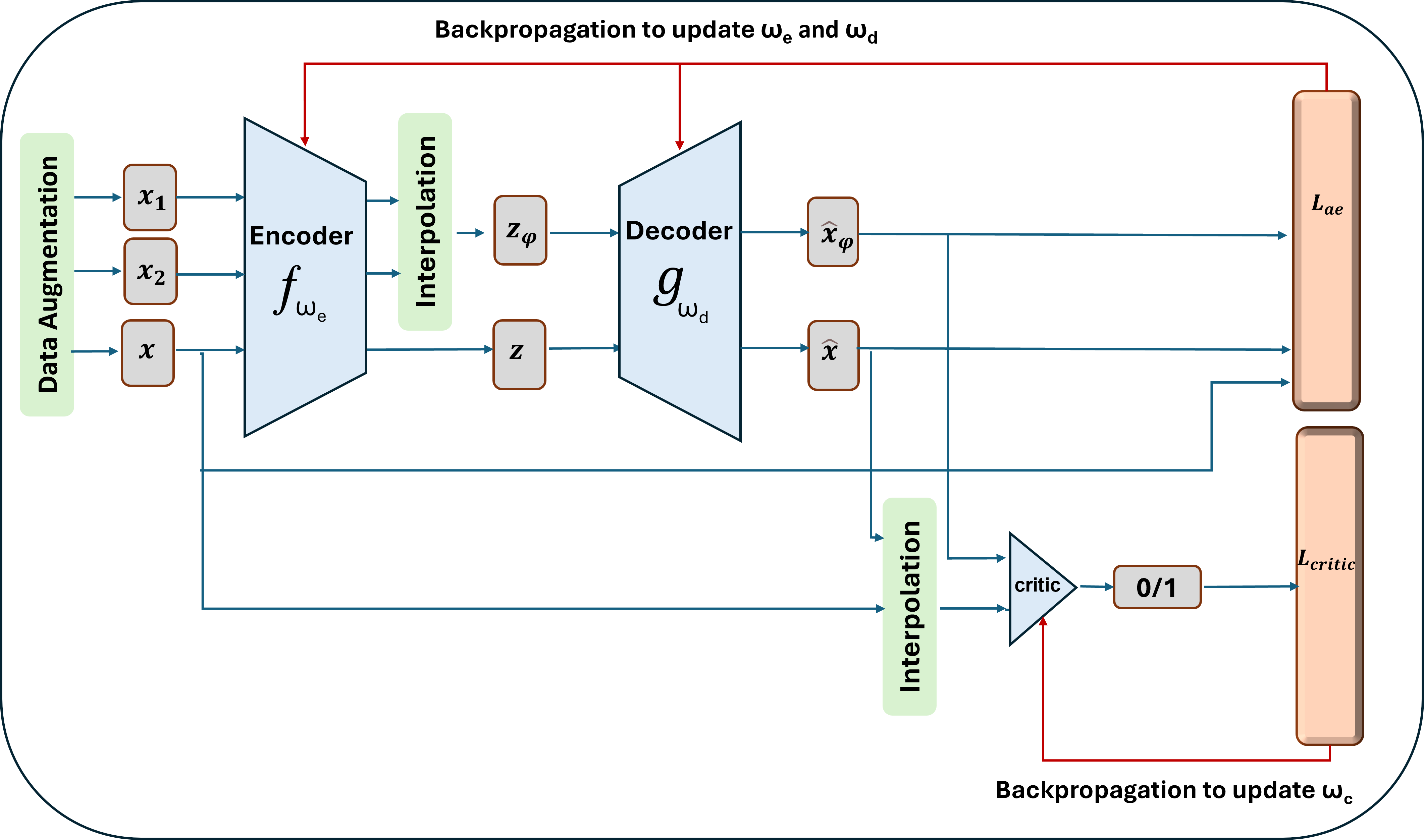}
     \caption{\textbf{The pretraining phase of R-DC.} The pretraining framework involves a reconstruction objective function regularized with adversarially constrained interpolation.}
  \label{fig:pretraining}
\end{figure}

For the pretraining phase, we opt for a reconstruction objective function regularized with adversarially contained interpolation (ACAI) \cite{Berthelot2019understanding}. The latent interpolation pushes the embedding codes close to each other so that we can find relevant centroids for the nearest neighbors of the second phase. An auto-encoder and a critic are two adversarial networks involved in a competitive interaction within the ACAI framework. The aim is to generate latent space interpolations of the input samples and decode them while preserving semantic characteristics. Each iteration involves randomly sampling coefficients $\varphi$ and $\eta$ from the interval $[0, 1]$. Let $z_1 = f_{\omega_e}(x_1)$ and $z_2 = f_{\omega_e}(x_2)$ represent the latent codes of the two samples $x_1$ and $x_2$, respectively. The interpolation of the latent codes
$z_1$ and $z_2$ is denoted by $z_{\varphi} = \varphi f_{\omega_e}(x_1) + (1 - \varphi)f_{\omega_e}(x_2)$. For randomly selected couples of data points $x_1$ and $x_2$, we compute $\hat{x}_{\varphi} = g_{\omega_d}(z_{\varphi})$, which represents the reconstructed data point with a latent representation interpolated from $z_1$ and $z_2$. The critic network, denoted by $c_t$, with learnable parameters $\omega_c$, is trained to regress the interpolation coefficient $\varphi$ from $\hat{x}_{\varphi}$, while the main network is trained to fool the critic into perceiving the interpolation-based points $x_{\varphi}$ as realistic. The loss functions for the auto-encoder $\mathcal{L}_{\text{ae}}$ and the critic $\mathcal{L}_{\text{critic}}$ are shown in Eq. \ref{eq:auto-encoder_loss} and Eq. \ref{eq:critic_loss}, respectively.  
The whole pretraining framework is illustrated in Figure \ref{fig:pretraining}.

\begin{equation}
\mathcal{L}_{\text{ae}}(\omega_e, \, \omega_d) =  \left \|  x - \hat{x} \right \| ^2_2 + \lambda_t \left \| c(\hat{x}_{\varphi}) \right \|^2_2,
\label{eq:auto-encoder_loss}
\end{equation}

\begin{equation}
\mathcal{L}_{\text{critic}}(\omega_c) = \left \| c(\hat{x}_{\varphi}) - \varphi \right \|^2_2 + \left \| c(\eta x + (1 - \eta)\hat{x}) \right \|^2_2.
\label{eq:critic_loss}
\end{equation}

The regularization term in the critic's loss function $\left \| c(\eta x + (1 - \eta)\hat{x}) \right \|^2_2$ pushes the output to be equal to $0$ for non-interpolation-based points. Thus, it ensures the generation of realistic representations of the decoded interpolants. As a key advantage of our pretraining strategy, the interpolation-constrained regularisation guarantees a dense latent space \cite{Berthelot2019understanding}. This aspect is crucial for the second phase of our approach because it allows for the effective decoding of the constructed points, corresponding to the centroids of the neighbors of each latent code $z_i$. As a result, the decoded latent codes of the constructed centroids are well-refined and representative of the input data distribution. Without the interpolation-constrained regularisation, the decoding process of the second phase yields blurry and unrepresentative decoded images.

\subsection{Second Phase: Proximity-Level Self-Supervision} \label{secondphase}


Following pretraining, the auto-encoder weights undergo a finetuning process through a second round of self-supervision training. We propose a proximity-level technique as a substitute for pseudo-supervision. We argue that proximity-level self-supervision has three advantages over pseudo-supervision. First, the transition between instance-level and neighborhood-level training is smoother and less abrupt than the transition between instance-level self-supervision and clustering-level pseudo-supervision. Second, eliminating the pseudo-supervision task prevents the risk of generating random features, which are caused by the error-prone nature of the pseudo-labels. Third, getting rid of pseudo-supervision relinquishes the need to perform joint clustering and instance-level training, and hence eliminates the risk of FD.



For the second phase, our objective function consists of two distinct components. Both are derived from a proximity-level abstraction of the input data. 
In particular, the first term, represented by $\mathcal{L}_{1}$, is a dynamic loss function that changes gradually during the training process from a vanilla reconstruction loss to a more sophisticated loss that achieves nearest-neighbor centroid construction of the latent codes. The second term, denoted by $\mathcal{L}_{2}$, is a latent space loss function that pushes the embedding points toward the centroids of some of their nearest neighbors. We leverage a dual filtering mechanism, illustrated in Figure \ref{fig:Picture_best}, to select the core points (i.e., located in dense regions) and the high-confidence neighbors of these core points. As a principal motivation, we hypothesize that the high-confidence nearest neighbors of a core latent code $z_{i}$ are highly likely to be in the same cluster as the point itself. We have rigorously tested and validated this hypothesis in the experiments section \ref{exp}, as detailed later. The full framework of the second phase is shown in Figure \ref{fig:autc}.


\subsection{Nearest-Neighbor Centroid Construction} \label{NNCC}

The concept of nearest-neighbor centroid construction entails using the auto-encoder to generate images that represent the centroids of the latent space neighbors for each sample $x_{i}$, as opposed to generating a reconstruction $\left \{ \hat{x}_i \right \}_{i=1}^{N}$ of the original images $\left \{ x_i \right \}_{i=1}^{N}$. To achieve this, we leverage the decoder $g_{\omega_d}$ to generate the images of the centroids. Over time, our objective function $\mathcal{L}_{1}$ methodically phases out the reconstruction term, and harnesses a filtering mechanism to shift towards a 
nearest-neighbor centroid construction.



To construct accurate nearest-neighbor centroids of the latent codes, our approach relies on a filtering mechanism that operates in two steps. Our filtering strategy offers an insightful perspective on capturing local structures and relationships between latent neighbors. We emphasize the idea that latent samples form two groups: core points and border points. The core points are located in densely populated regions of the latent space, where the concentration of similar latent codes is high. As opposed to that, the border points lie on the periphery of the dense regions. First, our filter initially selects the core latent points that lie in dense regions. After that, only the most reliable neighbors of each core point are selected. Then, we compute the centroid of the selected nearest neighbors for each core point. We decode these centroids and we train the auto-encoder to map the core samples to their decoded latent space centroids. As for the border points, they undergo a vanilla reconstruction process. As a result, the trained model progressively learns proximity-level abstraction of the input images and gradually reshapes the latent manifolds' structure to increase the number of core points.


\begin{figure}
    \includegraphics[width=\linewidth]{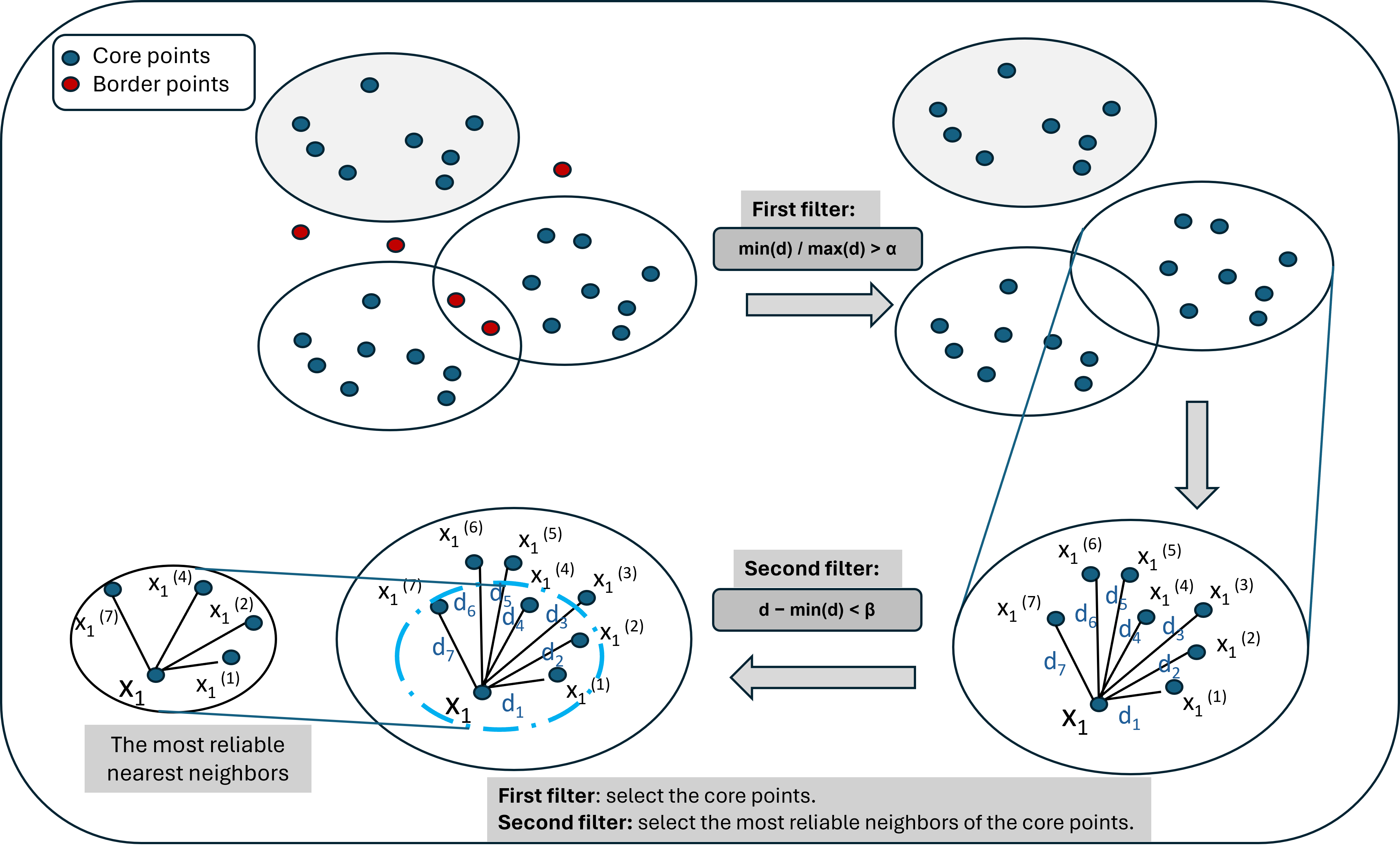}
     \caption{\textbf{ A dual filtering mechanism.} \textbf{First filter:} selecting the core points. \textbf{Second filter:} selecting the most reliable neighbors of the core points.}
  \label{fig:Picture_best}
\end{figure}

As illustrated in Figure \ref{fig:Picture_best}, the first filtering step aims to form two groups of samples, the core points and the border points. To identify the core points among the whole set of input samples $X$, we first compute the matrix $D = (d_{i,m})_{i \in \{ 1, \, \ldots, \, N \}, \ m \in \{1, \, \ldots, \, M\}}$, where $d_{i,m}$ captures the distance between each latent code $z_i$ and its $m^{\text{th}}$ nearest neighbor as illustrated by Eq. \ref{eq:d}. 

\begin{equation}
\begin{aligned}
    d_{i,m} &= \left \| z_i - \mathcal{NN}(z_i, \, m) \right \|_{2},  \quad i \in \{ 1, \, \ldots, \, N \}, \quad  \ m \in \{1, \, \ldots, \, M\}.
\end{aligned}
\label{eq:d}
\end{equation}

After computing the matrix $D$, the next step of the first filtering mechanism is to construct the set of core points $\Omega$. This is done by computing the ratio $r_i$ of the closest and farthest nearest neighbors for each latent code $z_i$, as explained in Eq. \ref{eq:r}, and then comparing this ratio with a threshold $\alpha$. The core points are located in dense regions characterized by a ratio $r_i$ greater than the threshold $\alpha$, as described by Eq. \ref{eq:omega}. A ratio $r_i$ close to $1$ means that all the $M$ nearest neighbors of $z_i$ are located within a small proximity. We denote by $S$ the set of latent codes associated with the core points in $\Omega$, as described by Eq. \ref{eq:S}. Border points are identified as points that do not satisfy the density criteria required to be considered as core points. The set of border points is mathematically defined as $\{ 1, \, \ldots, \, N \} - \Omega$. For the border points, they undergo a vanilla reconstruction process.


\begin{equation}
    r_i = \frac{\min_{m=1,\ldots, \, M}(d_{i,m})}{\max_{m=1,\ldots,M}(d_{i,m})},
\label{eq:r}
\end{equation}

\begin{equation}
\begin{aligned}
    \Omega &= \left\{ i \in \{ 1, \, \ldots, \, N \}  \; \middle| \; r_i \geq   \alpha \right\},
\end{aligned}
\label{eq:omega}
\end{equation}

\begin{equation}
\begin{aligned}
    S &= \left\{ z_i \in f(X)  \; \middle| \; i \in \Omega \right\}.
\end{aligned}
\label{eq:S}
\end{equation}


After forming the set of core points $\Omega$, the second filtering mechanism consists of selecting the most reliable nearest neighbors of each core point $z_{i}$ among all their neighbors, denoted by $\Omega_{nn}^{i}$. We start by computing the matrix $H = (h_{i,m})_{i \in \{ 1, \, \ldots, \, N \}, \ m \in \{1, \, \ldots, \, M\}}$, where $h_{i,m}$ captures the difference between $d_{i,m}$ (i.e., the distance between $z_i$ and its $m^{\text{th}}$ nearest neighbor) and $\bar{d_{i}}$ (i.e., the distance between the latent code $z_i$ and its closest nearest neighbor), as described by Eq. \ref{eq:d_bar} and Eq. \ref{eq:h}, respectively. Then, we compare the distance $h_{i,m}$ with a threshold $\beta$. The most reliable neighbors $\Omega_{nn}^{i}$ of a core point $z_{i}$ are located in a region characterized by a distance $h_{i,m}$ lower than the threshold $\beta$, as explained in Eq. \ref{eq:omega_nn}. The smaller the distance $h_{i,m}$, the more reliable the nearest neighbor is. We denote by $S_{nn}^{i}$ the set of the latent codes of the most reliable neighbors $\Omega_{nn}^{i}$ of a core point $z_i$, as described by Eq. \ref{eq:S_nn}.


\begin{equation}
   \bar{d_{i}} = \min_{m'=1,\ldots, \, M}(d_{i,m'}),
\label{eq:d_bar}
\end{equation}

\begin{equation}
   h_{i, m} = d_{i,m} - \bar{d_{i}},
\label{eq:h}
\end{equation}

\begin{equation}
\begin{aligned}
    \Omega_{nn}^{i} &= \left\{ m \in \{ 1, \, \ldots, \, M \}  \; \middle| \; h_{i,m} \leq \beta \right\},
\end{aligned}
\label{eq:omega_nn}
\end{equation}

\begin{equation}
\begin{aligned}
    S_{nn}^{i} &= \left\{ \mathcal{NN}(z_i, \, m)  \; \middle| \; m \in \Omega_{nn}^{i} \right\}.
\end{aligned}
\label{eq:S_nn}
\end{equation}

We define the function $\sigma_{\omega_{e}}$, such that $\sigma_{\omega_{e}}(x_{i})$ computes the latent space centroid of the most reliable neighbors of a core point $x_i$, as described by Eq. \ref{eq:sigma}. The first term $\mathcal{L}_1$ of our objective function is a dynamic loss that \textit{changes} gradually during the training process from vanilla reconstruction to a nearest-neighbor centroid construction, as illustrated in Eq. \ref{eq:L_{1}}. For centroid construction, only core points are chosen. The border points undergo a reconstruction process until the model learns a geometric configuration that pushes them to the core region of the latent manifolds. Thus, based on the data sample, the training can follow one of two schemes: either reconstruction or nearest-neighbor centroid construction.


    \begin{equation} \label{eq:sigma}
        \sigma_{\omega_{e}}(x_{i}) = \frac{1}{\left | S_{nn}^{i} \right |} \sum_{m \in S_{nn}^{i}} f_{\omega_{e}}(x_{i}),
    \end{equation}
    
    \begin{equation} \label{eq:L_{1}}
        \begin{aligned}
        \mathcal{L}_1(\omega_e,  \, \omega_d) = \sum_{i=1}^{N} \left\{
                    \begin{array}{ll}
                      \left \| x_{i} - \hat{x}_{i} \right \|_{2}^{2} \;\;\; \;\;\;\;\;\;\;\;\;\;\;\;\;\;\;    \text{if} \;  i \; \not \in \; \Omega,
                      \\
                      \left \| g_{\omega_{d}}(\sigma_{\omega_{e}}(x_{i})) - \hat{x}_{i} \right \|_{2}^{2} \quad  \text{otherwise}.
                    \end{array}
                  \right. 
        \end{aligned}
    \end{equation}
    
At the end of the dynamic training process, each sample becomes linked to its corresponding nearest-neighbor centroid. This leads to the generation of smoother output images that represent a whole neighborhood instead of the point itself.

\subsection{Proximity-Level Encoding} \label{PLE}
The goal of the proximity-level encoding process is to learn latent representations suitable to the clustering task. To achieve this, we leverage the encoder $f_{\omega_e}$ to project the data in a way that emphasizes proximity-level information. Our goal is to ensure a smooth transition from instance-level to proximity-level self-supervision that does not cause significant geometric distortions. In this context, we propose a second loss function that minimizes the distance between the embedding of each core point and the centroid of its most reliable nearest neighbors in the latent space. From a geometric perspective, we find that the second loss function smooths the local structures without twisting the manifolds.

Similar to the first loss function $\mathcal{L}_{1}$, we
leverage the same filtering mechanism to classify the samples into two groups (i.e., core points and border points) and extract the most reliable neighbors of each core point. Using the core points and their most reliable nearest neighbors, we propose a proximity-level latent space loss function $\mathcal{L}_{2}$. Our second loss function $\mathcal{L}_{2}$ minimizes the distance between the embedded core points and their corresponding nearest neighbors centroids in the latent space, as explained in Eq. \ref{eq:L2_F}.

\begin{equation}
\mathcal{L}_{2}(\omega_e) = 
\\
\sum_{x_{i} \in \Omega} \left \| f_{\omega_e}(x_i) -  \sigma_{\omega_e}(x_{i}) \right \|_{2}^{2}.
\label{eq:L2_F}
\end{equation}



\subsection{Algorithm and Optimization} \label{AO}

\begin{figure}
     \includegraphics[width=\linewidth]{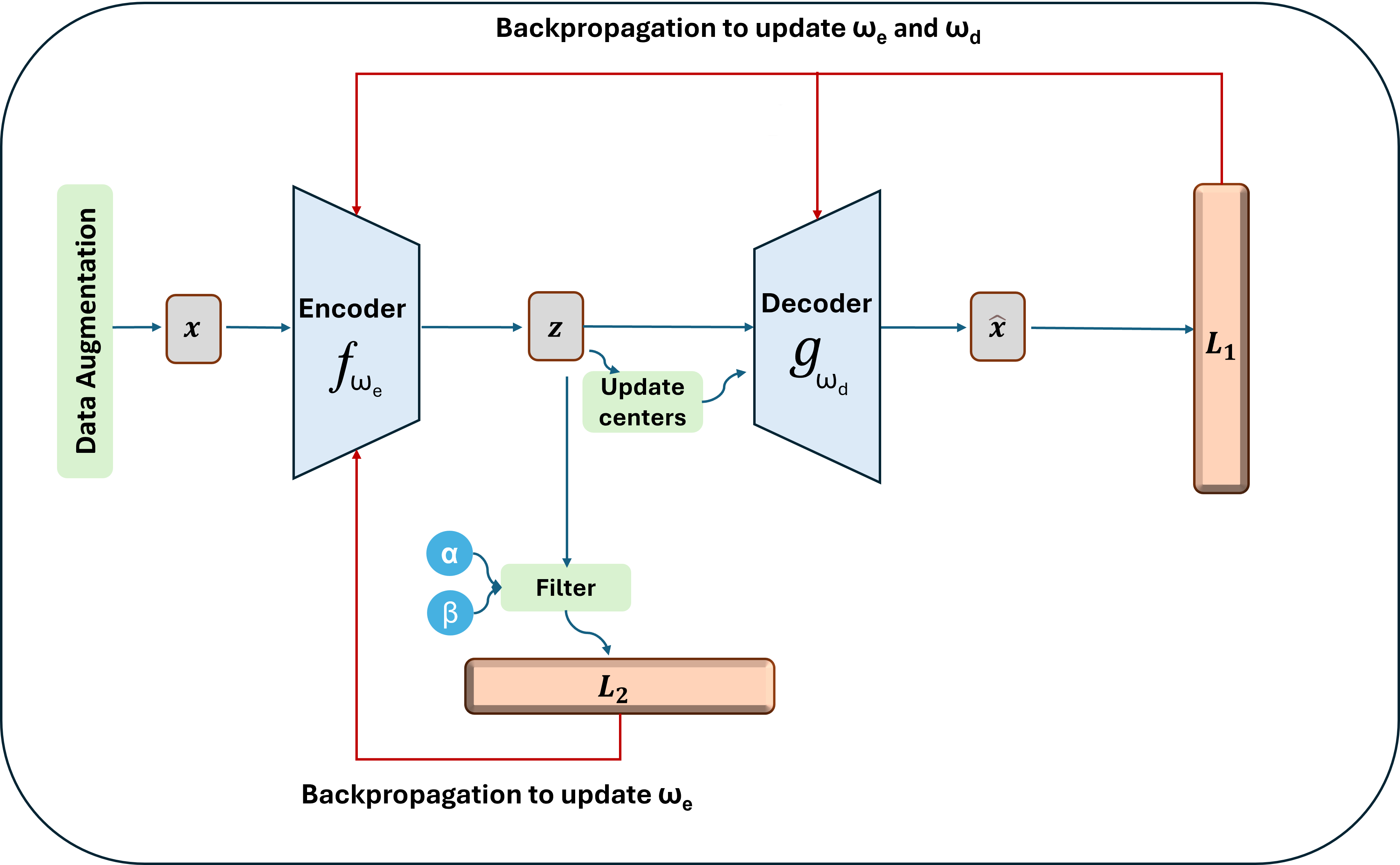}
     \caption{\textbf{The second phase of R-DC.} The auto-encoder weights undergo a finetuning process through a second round of self-supervision. The loss function $L_2$ represents a proximity-level self-supervision task. The loss function $L_1$  phases out the reconstruction term and harnesses dual filtering to shift towards a nearest-neighbor centroid construction.}
  \label{fig:autc}
\end{figure}

\begin{figure}
    \includegraphics[width=\linewidth]{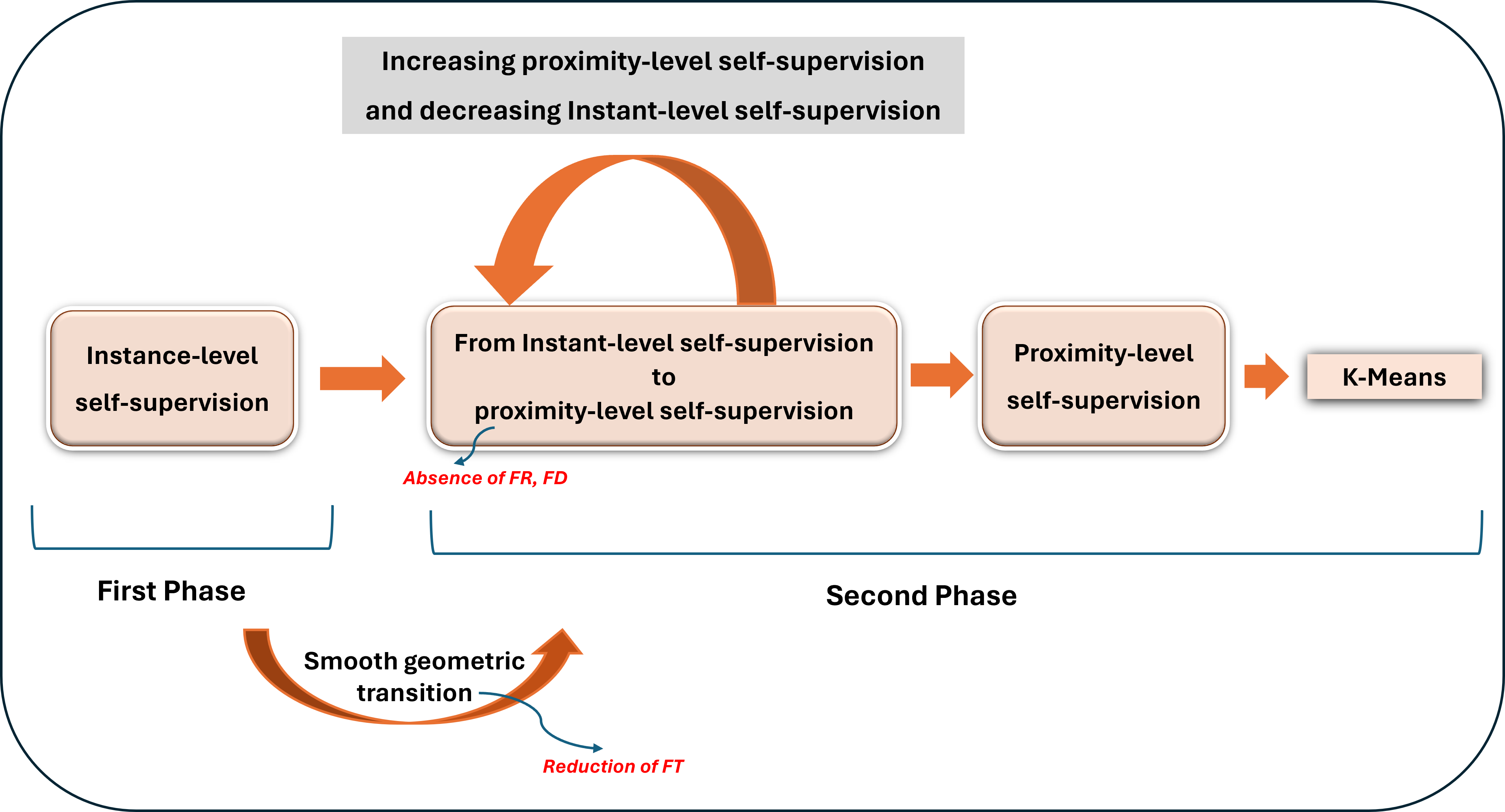}
     \caption{The advantages of R-DC from the perspective of FR, FD, and FT. }
  \label{fig:tr}
\end{figure}

The comprehensive objective function of the second phase $\mathcal{L}$ is a linear combination of the nearest-neighbor centroid construction loss $\mathcal{L}_{1}$ and the proximity-level encoding loss $\mathcal{L}_{2}$, as delineated in Eq. \ref{eq:total_loss}. Similar to the pretraining stage, the loss function $\mathcal{L}$ undergoes regularization via data augmentation. The training process of the second phase is illustrated in Figure \ref{fig:autc}.


\begin{equation}
\mathcal{L}(\omega_e, \, \omega_d) = \mathcal{L}_{1}(\omega_e,  \, \omega_d) + \mathcal{L}_{2}(\omega_e).
\label{eq:total_loss}
\end{equation}

Throughout the second phase training process, the number of core points will increase gradually. 
We denote by $\tau$ the ratio of the core points, as described by Eq. \ref{eq:tau}, and we train our model until this ratio remains stable. Local convergence is achieved when the loss functions $\mathcal{L}_{1}$ and $\mathcal{L}_{2}$ stabilize. To prevent falling into local convergence, we update the centroids of the most reliable nearest neighbors of the core points every training iteration. 

\begin{equation}
\tau = \frac{\text{$|S|$}}{|X|}.
\label{eq:tau}
\end{equation}

We continue minimizing the loss function $\mathcal{L}$ until we reach full stability. Full stability is achieved when updating the centroids does not bring any further increase in the number of core points. At the end of the dynamic training process, each core point becomes linked to its corresponding nearest-neighbor centroid. This leads to the generation of smoother output images that represent a whole neighborhood instead of reconstructing the input data.

In a nutshell, our approach R-DC involves two training phases by leveraging a dual-stage self-supervision strategy. In the pretraining phase, we focus on instance-level self-supervision. In essence, the model learns general-purpose features that capture high-level representations of the input data. Moving to the second phase, we fine-tune the model based on two proximity-level self-supervision loss functions. By eliminating pseudo-supervision, our model prevents the occurrence of FR and FD. Furthermore, the transition from instance-level to proximity-level self-supervision is smoother and less coarse than the transition from instance-level self-supervision to pseudo-supervision from a geometric perspective. The advantages of our approach from the perspective of FR, FD, and FT are illustrated in Figure \ref{fig:tr}.

We pretrain our model for $T_{1}$ iteration. At each iteration, we alternate between optimizing the auto-encoder parameters $\left \{ \omega_{e}, \, \omega_{d} \right \}$ using the loss function $\mathcal{L}_{\text{ae}}$ and the critic parameters $\left \{ \omega_{c} \right \}$ using the loss function $\mathcal{L}_{\text{critic}}$. Next, we fine-tune the model until we reach full stability, characterized by a constant ratio $\tau$. The second phase updates the auto-encoder parameters  $\left \{ \omega_{e}, \, \omega_{d} \right \}$ to minimize the loss function $\mathcal{L}$. For both training phases, we perform mini-batch back-propagation to adjust the auto-encoder and critic weights using the Adam optimizer. At the end of the finetuning process, we apply the K-means algorithm to the latent codes, provided in the $Z$ matrix, to identify the clustering assignments. The matrix $Z$ is the matrix of latent codes obtained after applying the encoder \( f_{\omega_e} \) to the input data \( X \). Specifically, each data point \( x_i \) is transformed into its latent representation \( z_i \) by the encoder, and the collection of these latent representations forms the $Z$ matrix. The $Z$ matrix is used as input for the clustering algorithm, such as K-means, to identify clusters in the latent space.  
Similar to previous deep clustering methods such as DEC, IDEC, DynAE, CC, PSSC, DSSC-UCO, and VaDE, our method requires the number of clusters $K$ to be a predefined parameter. This predefined $K$ is necessary to apply the K-means algorithm. It is important to highlight that all the methods considered in our comparison, except FINCH, use a predefined number of clusters. In scenarios where the dataset does not have a predefined number of clusters, we recommend using DBSCAN on the latent representations $Z$ as an alternative to K-means after the second training phase. DBSCAN can automatically determine the number of clusters based on the density of data points, which makes it suitable for datasets with unknown cluster counts.


The full proposed method is summarized in Algorithm \ref{algo}. The computational complexity of R-DC is  $\mathcal{O}(m \, L \, D^2 \, N_{B} \, + M \, N \, \text{log}(N))$, where \(m\) represents the full number of training iterations for pretraining and finetuning, \(L\) denotes the number of layers, \(N_{B}\) is the mini-batch size, \(N\) is the size of the full dataset and \(D\) represents the maximum number of neurons in all layers. It's noteworthy that our approach R-DC shares similar computational complexity compared with DEC, IDEC, and DynAE. The main difference from a computational perspective relies on applying the K-NN (k nearest neighbors) algorithm in our case. Consequently, with identical network architectures, batch sizes, optimizers, and pretraining phases, the execution times of the four approaches (DEC, IDEC, DynAE, and R-DC) are very similar.

\begin{algorithm}[htbp]
\caption{Training strategy of R-DC}
\begin{algorithmic}[1] 

\State \textbf{Input:} Input data: $X$, Number of Neighbours: $M$, Number of pretraining iterations: $T_{1}$, Number of Clusters: $K$, Thresholds: $\alpha$, $\beta$. 

\For{$i=1$ \textbf{to} $T_{1}$}
    \If{$i \mod 2 == 0$}
        \State Compute $\mathcal{L}_{\text{ae}}$ according to Eq. \ref{eq:auto-encoder_loss}
        \State Update the auto-encoder $\left \{ \omega_{e}, \, \omega_{d} \right \}$ to minimize $\mathcal{L}_{\text{ae}}$ using Adam optimizer
    \Else
        \State Compute $\mathcal{L}_{\text{critic}}$ to Eq. \ref{eq:critic_loss}
        \State Update the critic $\left \{ \omega_{c} \right \}$ to minimize $\mathcal{L}_{\text{critic}}$ using Adam optimizer
    \EndIf
\EndFor
\State $\tau_{\text{prev}} \leftarrow 0$ 
\State $\tau \leftarrow 1$ 
\While{$\tau$ $\neq$ $\tau_{\text{prev}}$}
    \State $Z \leftarrow f_{\omega_{e}}(X)$
    \State Compute the nearest neighbor distance matrix $D$ according to Eq. \ref{eq:d}
    \State Compute the set of core points $\Omega$ according to Eq. \ref{eq:omega}
    \State Compute the latent codes of the core points $S$ according to Eq. \ref{eq:S}
    \State Compute the set of most reliable neighbors $\Omega_{nn}^{i}$ according to Eq. \ref{eq:omega_nn} 
    \State Compute the latent codes of the most reliable neighbors $S_{nn}^{i}$ according to Eq. \ref{eq:S_nn}
    \State Compute $\mathcal{L}_{1}$ according to Eq. \ref{eq:L_{1}}
    \State Compute $\mathcal{L}_{2}$ according to Eq. \ref{eq:L2_F}
    \State Compute $\mathcal{L}$ according to Eq. \ref{eq:total_loss}
    \State Update the auto-encoder $\left \{ \omega_{e}, \, \omega_{d} \right \}$ to minimize $\mathcal{L}$ using Adam optimizer
    \State $\tau_{\text{prev}} \leftarrow \tau$  
    \State Compute $\tau$ according to Eq. \ref{eq:tau}  
\EndWhile
\State Apply K-means to the matrix $Z$ to identify the clusters 
\end{algorithmic}
\label{algo}
\end{algorithm}

\section{Experiments} \label{exp}


Our approach introduces a new model without pseudo-supervision to enhance the dynamic training process through dual-stage self-supervision. Our approach prevents the problems of FR and FD by eliminating pseudo-supervision. Furthermore, we alleviate the FT problem by smoothly flattening the latent manifolds during the transition from instance-level to proximity-level self-supervision. We carry out comprehensive experiments to show the merits of our method. The obtained results provide strong evidence of R-DC effectiveness. In particular, R-DC can address the FT problem compared with the state-of-the-art approaches. In Subsection 4.1, we detail the experimental configurations, and in Subsection 4.2, we present our results. 

\subsection{Experimental Settings}


All experiments are conducted on a Linux server, maintaining uniformity in hardware and software environments. The software setup includes Python version 3.6.8 and the TensorFlow deep learning library. On the hardware side, we utilize a processing unit equipped with 378 GB of RAM and a Tesla T4 GPU, which boasts 16 GB of memory. The code of R-DC is published on \url{Github}. \footnote{You can find the code at: \url{https://github.com/Amalsalem/Rethinking-DC}.}





\begin{table*}[htbp]
\scalebox{0.95}{ 
\begin{tabular}{|p{3.8cm}|c|c|p{4.1cm}|}
\hline
\textbf{Dataset} & \textbf{$\#$ Samples} & \textbf{$\#$ Classes} & \textbf{Description} \\ \hline
\textbf{PneumoniaMNIST} & 5,856  & 2  & Chest X-Ray \\ \hline
\textbf{BreastMNIST} & 780 & 3 & Breast Ultrasound \\ \hline
\textbf{BloodMNIST} & 17,092 & 8 & Blood Cell Microscope \\ \hline
\textbf{VesselMNIST3D} & 1,908 & 2 & Brain MRA \\ \hline
\textbf{CIFAR-10} & 60,000 & 10 &  Color Images \\ \hline
\textbf{FMNIST} & 70,000 & 10 & Fashion Images \\ \hline
\end{tabular}}
\centering
\caption{Descriptions of the used Datasets. }
\label{datasets}
\end{table*}

\subsubsection{Datasets and Baselines}
We have conducted comprehensive comparisons of R-DC against nine state-of-the-art deep clustering approaches across six benchmark datasets. 
The experimental analysis includes four MedMNIST datasets \cite{jiancheng_yang_2021_6496656}: VesselMist3D, PneumoniaMNIST2D, BreastMNIST, and BloodMNIST; along with two other benchmark datasets, namely Cifar10 \cite{cifar10} and FMNIST \cite{xiao2017fashionmnist}. Key statistics for these datasets are provided in Table \ref{datasets}. Before inputting the matrix $X$ into the neural network, we standardize it through row normalization.

\begin{itemize}
\item
\textbf{PneumoniaMNIST}: A binary-class classification of 5,856 pediatric chest X-Ray images. The source images are gray-scale, and their sizes are (384–2,916)×(127–2,713).
\item
\textbf{BreastMNIST}: A 3-class dataset of 780 samples of breast ultrasound images. The source images of 1×500×500.
\item
\textbf{BloodMNIST}: An 8-class dataset of 17,092 images. The input images have a resolution of 3×360×363 pixels.
\item
\textbf{VesselMNIST3D}: This dataset originates from the publicly available 3D intracranial aneurysm dataset, IntrA34, VesselMNIST3D, which consists of 103 3D models (meshes) representing entire brain vessels reconstructed from MRA images. The dataset includes 1,694 healthy vessel segments and 215 aneurysm segments extracted from these models.
\item
\textbf{CIFAR-10}: A 10-class dataset of 60,000 images and their sizes are 32×32 color images.
\item
\textbf{FMNIST}: A 10-class dataset that encompasses 60,000 images, each of 32×32 pixels, distributed across 10 distinct classes.

\end{itemize}

Our baselines include nine models, namely: FINCH \cite{sarfraz2019efficient}, VaDE \cite{jiang2017variational}, AE+K-means, DEC \cite{xie2016unsupervised}, IDEC \cite{guo2017improved}, PSSC \cite{lv2021pseudo}, CC \cite{li2021contrastive}, DDC-UCO \cite{trosten2024leveraging} and DynAE \cite{mrabah2020deep}. We have discussed all these methods in the related work section. Among the selected baselines, both FINCH and DEC are categorized within the second paradigm. On the other hand, DynAE, VaDE, and IDEC are aligned with the third paradigm. This diverse set of models provides a solid foundation for our performance evaluation and comparison. We have also compared R-DC against AE+K-Means, where we use the same auto-encoding architecture as DynAE, DEC, IDEC, and R-DC to perform vanilla reconstruction and apply K-Means at the end of the training process.

\subsubsection{Evaluation metrics}
For evaluation purposes, we utilize three standard clustering metrics, namely Accuracy (ACC), F1 Micro, and F1 Macro, and two geometric ones, namely ID and LID.
\begin{itemize}
\item {\textbf{Clustering performance:}

ACC is the most utilized evaluation metric in deep clustering. Recognizing the diversity of our datasets in terms of size and class distribution, we have found that both the Micro F1 score and Macro F1 score offer a more suitable means of evaluation. These F1 score variants take into consideration the details of varying dataset characteristics and class imbalances, enabling us to gauge the quality of the clustering outcomes more effectively. ACC, Micro F1 score, and Macro F1 score are reported in percentages. The higher these values, the more favorable the clustering outcomes.}

\item {\textbf{Intrinsic Dimension and Linear Intrinsic Dimension:}

Two metrics, ID and LID, help us understand how geometric transformations occur during training. ID represents the minimum number of parameters needed to accurately capture the fundamental features of the data. It reflects the underlying complexity of the latent manifolds. 
To estimate ID, we leverage the TwoNN technique \cite{facco2017estimating}. This computationally efficient method requires only the two nearest neighbors of each sample. This makes it well-suited for scenarios with highly curved and non-uniformly sampled manifolds, where traditional density-based approaches might struggle. On the other hand, LID represents the dimensionality of the lowest-rank subspace encompassing the entire data manifold. We estimate LID using PCA, similar to \cite{ansuini2019intrinsic}. PCA identifies the principal components (directions of greatest variance) that capture the data with minimal error, essentially defining the best-fitting linear subspace.

The difference between LID and ID serves as a measure of the curvature of the data manifold. In cases where the manifold exhibits significant curvature, the linear intrinsic dimension (\(LID\)) vastly exceeds the real intrinsic dimension (\(ID\)). Conversely, in scenarios where the manifold is relatively flat, the linear intrinsic dimension aligns closely with the real intrinsic dimension (\(LID \approx ID\)). However, having LID lower than ID can be beneficial because it suggests that a lower-dimensional linear representation can effectively capture the essential structure of the data. This can lead to simpler and more interpretable models, as well as potentially lower computational costs.}


\end{itemize}

\subsubsection{Implementation}

We use an auto-encoding architecture. Our encoder and decoder consist of fully connected processing layers. The auto-encoder comprises 8 layers with dimensions \(d - 500 - 500 - 2000 - 10 - 2000 - 500 - 500 - d\), where $d$ is the dimension of the input data. With the exception of the bottleneck layer and the final layer, all other layers utilize the Rectified Linear Unit (ReLu) as the activation function \cite{maas2013rectifier}. During the pretraining phase, the auto-encoder undergoes adversarial training end to end, competing with a critic network for a number of iterations. In this stage, all learnable parameters $\omega_c$, $\omega_d$, and $\omega_e$, are updated using the Adam optimizer \cite{Kingma2014AdamAM} with a learning rate of $0.001$. The number of iterations for the first phase $T_{1} = 10000$. For the second phase, the auto-encoder is trained until there is no more increase in the number of core data points. In this stage, the auto-encoder parameters $\omega_d$, and $\omega_e$, are updated using the Adam optimizer with a learning rate of $0.001$. 

We set the number of nearest neighbors in the second phase to \(M=5\). This choice is justified based on empirical testing. We evaluated different values of \(M\), such as 3 and 7, across all datasets. The results indicated that \(M=5\) provided the best balance between computational efficiency and clustering performance. With M = 5, each core point has a sufficient number of neighbors within the same cluster, while avoiding the inclusion of too many potentially irrelevant neighbors that could degrade performance. 

We have two data-dependent hyperparameters $\alpha$ and $\beta$. By setting $\alpha$, we control the number of core points, ensuring that only points in sufficiently dense regions are selected. Typically, we expect the value of $\alpha$ to be less than 1. Smaller values of $\alpha$ mean that fewer points must be close to each other to be considered a dense region, while larger values imply a stricter criterion for density, selecting only points in very dense areas. By setting $\beta$, we ensure that only the nearest neighbors with minimal distance variations are selected, which are more reliable. The value of $\beta$ is expected to be relatively small, as we want to capture neighbors that are very close to the core points. Smaller values of $\beta$ mean that only neighbors with very little variation in distance from the core point are selected, ensuring high reliability.

The two hyperparameters $\alpha$ and $\beta$ are selected, as described in Table \ref{Table:hyperparameters_dep}, based on grid search within the ranges [0.05, 0.1, 0.15, 0.2, 0.25] for $\alpha$ and [0.5, 0.6, 0.7, 0.8, 0.9] for $\beta$. The optimal values for these hyperparameters consistently fall within the intervals [0.5, 0.8] for $\alpha$ and [0.5, 1] for $\beta$. Therefore, we recommend setting $\alpha$ and $\beta$ within these intervals for other datasets.

Our method employs the same fully-connected neural network architecture as AE+K-means, DEC, IDEC, VaDE, and DynAE. To ensure a fair comparison with methods utilizing convolutional neural network architectures, we apply two data augmentation techniques to AE+K-means, DEC, IDEC, VaDE, DynAE, and R-DC. These techniques ensure invariance with respect to rotation and translation. Specifically, the height shift range is randomly selected within [0, 0.1], the width shift range within [0, 0.1], and the rotation angle within [0, 10] degrees.

 

\begin{table*}
  \begin{center}
  \begin{small}
  \resizebox{\textwidth}{!}{ 
  \begin{tabular}{|p{1.9cm}|c|c|c|c|c|c|c|c|}
    \hline
    \textbf{Parameter} & \textbf{PneumoniaMNIST} & \textbf{BreastMNIST} & \textbf{BloodMNIST} & \textbf{VesselMNIST3D} & \textbf{CIFAR10} & \textbf{FMNIST} \\ \hline
    \textbf{$\alpha$} & 0.8 & 0.6 & 0.5 & 0.3 & 0.6 & 0.6 \\ \hline
    \textbf{$\beta$} & 1  & 0.7  & 0.6  &  0.5 & 0.6 & 0.6 \\ \hline
  \end{tabular}
  }
  \end{small}
  \end{center}
  \caption{\textbf{Data-dependent hyperparameters:} The hyperparameters $\alpha$ and $\beta$ are critical for the filtering mechanism in our model. These parameters were carefully selected based on their role in identifying core points and the most reliable nearest neighbors in the latent space.}
 \label{Table:hyperparameters_dep}
\end{table*}

\subsection{Results}
In this section, we will present our results, beginning with the clustering performance, followed by the Feature Twist analysis, the ablation study, sensitivity analysis, and finally, some latent space visualizations.

\subsubsection{Clustering Performance}
Table \ref{table:com2_51}, Table \ref{table:Breast_Blood} and Table \ref{table:Vessel_Pneumonia} show a comparison of R-DC with nine state-of-the-art clustering methods across six datasets: CIFAR10, FMNIST, VesselMNIST3D, PneumoniaMNIST2D, BreastMNIST, and BloodMNIST. In our initial evaluation, we assess the clustering effectiveness of R-DC in comparison to the baseline approaches: FINCH, VADE, AE + k-means, DEC, IDEC, PSSC, CC, DSSC-UCO, and DynAE, utilizing metrics such as ACC, F1 Score Macro, and F1 Score Micro. As observed, R-DC achieves the highest values across all six datasets compared to the other methods. Similarly, for F1 Macro and F1 Micro, R-DC also achieves competitive and superior performance compared to the other methods.

When compared to DynAE, R-DC has shown significant improvements in all the datasets. In the PneumoniaMNIST dataset, R-DC achieves an accuracy of 86.7\%, significantly surpassing DynAE's accuracy of 62.1\% by 24.6\%. In the BloodMNIST dataset, R-DC achieves an accuracy of 59.1\%, showcasing a remarkable improvement of 12.8\% over DynAE's accuracy of 46.3\%. In the FMNIST dataset, R-DC achieves an accuracy of 66.6\%, outperforming DynAE's accuracy of 59.5\% by 7.1\%. Similarly, in the CIFAR10 dataset, R-DC exhibits an accuracy of 60.1\%, showcasing a notable 3.2\% increase over DynAE's accuracy of 56.9\%. In the BreastMNIST dataset, R-DC achieves an accuracy of 66.7\%, outperforming DynAE's accuracy of 63.8\% by 2.9\%. Lastly, in the VesselMNIST dataset, R-DC achieves an accuracy of 88.2\%, surpassing DynAE's accuracy of 86.2\% by 2\%. The finding approves that preventing FR and FD and alleviating FT significantly contribute to enhancing R-DC. The significant improvements of R-DC over DynAE and other methods across various datasets can be attributed to its effective dual-stage self-supervision strategy. Our approach leverages proximity-level self-supervision for the finetuning process. Unlike previous methods, R-DC does not rely on pseudo-supervision. Thus, it prevents the occurrence of FR and FD. Furthermore, our strategy progressively refines the latent representations, ensuring a smoother transition between the two training phases and reducing the risk of geometric distortions. The gradual transition from instance-level to proximity-level self-supervision ensures a smooth transformation of the latent manifolds, contrasting with other deep clustering methods that perform an abrupt switch from self-supervision to pseudo-supervision. These advantages enable R-DC to learn robust clustering-oriented latent representations, leading to better clustering performance.



\begin{table*}
  \centering
  \scalebox{0.87}{ 
    \begin{tabular}{|p{3.3cm}|c|c|c|c|c|c|}
      \hline
      {\textbf{Method}} & \multicolumn{3}{c|}{\textbf{CIFAR10}}  & \multicolumn{3}{c|}{\textbf{FMNIST}} \\
      \cline{2-7}
      & \textbf{ACC} & \textbf{F1 Macro} & \textbf{F1 Micro} &  \textbf{ACC} & \textbf{F1 Macro} & \textbf{F1 Micro} \\
      \hline
      \textbf{FINCH} & 48.2 & 45.2 & 48.2 &  52.5 & 49.1 & 52.5 \\
     
      \textbf{VADE} & 20.3 & 18.8 & 20.3  & \underline{59.6} & \underline{59.4} & \underline{59.6} \\
      
      \textbf{AE + K-means} & 52.7 & 51.4 & 52.7 & 53.2 & 50.7 & 53.2 \\
     
      \textbf{DEC} & 51.4 & 49.9 & 51.4 & 51.9 & 50.0 & 51.9 \\
      
      \textbf{IDEC} & 50.3 & 48.4 & 50.3 &  53.9  & 50.2 & 53.9 \\
      \textbf{PSSC} & 23.3  & 22.0  & 23.3 & 56.6 & 54.9 & 56.6    \\
      \textbf{CC} &37.8 &  37.2 & 37.8 & 42.5 & 42.6 & 42.5 \\
      
      \textbf{DDC-UCO} & 33.8  & 19.8  & 33.8 & 49.9  & 44,2 & 31.4  \\

      \textbf{DynAE} & \underline{56.9} & \underline{55.6} & \underline{56.9} & 59.5 & 58.8 & 59.4 \\
      
      \textbf{R-DC} & \textbf{60.1} & \textbf{59.3} & \textbf{60.1} &  \textbf{66.6} & \textbf{65.7} & \textbf{66.6} \\
      \hline
    \end{tabular}}
    \caption{Clustering performances in terms of ACC, F1 Macro and F1 Micro for CIFAR10 and FMNIST. The best method is in bold and the second best is underlined.}
    \label{table:com2_51}
\end{table*}

\begin{table*}
  \centering
  \begin{small}
  \scalebox{0.87}{ 
    \begin{tabular}{|p{3.3cm}|c|c|c|c|c|c|}
      \hline
      {\textbf{Method}} & \multicolumn{3}{c|}{\textbf{VesselMNIST3D}} & \multicolumn{3}{c|}{\textbf{PneumoniaMNIST}} \\
      \cline{2-7}
      & \textbf{ACC} & \textbf{F1 Macro} & \textbf{F1 Micro} & \textbf{ACC} & \textbf{F1 Macro} & \textbf{F1 Micro} \\
      \hline
      \textbf{FINCH} & 83.5 & 45.5 & 83.5 & 51.8 & 37.0 & 51.8 \\
      \textbf{VADE} & 84.6 & \underline{46.6} & 84.6 & 71.8 & 68.9 & 71.8 \\
      \textbf{AE + K-means} & 85.3 & 46.0 & 85.3 & 61.4 & 58.6 & 61.4 \\
      \textbf{DEC} & 81.1 & 47.9 & 81.1 & 59.4 & 57.1 & 59.4 \\
      \textbf{IDEC} & 81.8 & 46.8 & 81.8 & 66.7 & 66.1 & 66.7 \\
      \textbf{PSSC} & 72.4 & 44.5 & 72.4 & \underline{81.1} & \underline{77.7}  &  \underline{81.1} \\
      \textbf{CC} & 88.7 & 47.0  & 88.7 & 53.4 & 52.8  & 53.4 \\
      \textbf{DDC-UCO} &  57.4 & 40.2  & 57.4  & 74.4 & 69.6 & 74.4  \\
      \textbf{DynAE} & \underline{86.2} & 46.2 & \underline{86.2} & 62.1 & 58.8 & 62.1 \\
      
      \textbf{R-DC} & \textbf{88.2} & \textbf{46.9} & \textbf{88.2} & \textbf{86.7} & \textbf{83.4} & \textbf{86.7} \\
      \hline
    \end{tabular}}
  \caption{Clustering performances in terms of ACC, F1 Macro, and F1 Micro for VesselMNIST3D and PneumoniaMNIST. The best method is in bold and the second best is underlined.}
  \label{table:Vessel_Pneumonia}
  \end{small}
\end{table*}

\begin{table*}
  \centering
  \begin{small}
  \scalebox{0.87}{ 
    \begin{tabular}{|p{3.3cm}|c|c|c|c|c|c|}
      \hline
      {\textbf{Method}} & \multicolumn{3}{c|}{\textbf{BreastMNIST}} & \multicolumn{3}{c|}{\textbf{BloodMNIST}} \\
      \cline{2-7}
      & \textbf{ACC} & \textbf{F1 Macro} & \textbf{F1 Micro} & \textbf{ACC} & \textbf{F1 Macro} & \textbf{F1 Micro} \\
      \hline
      \textbf{FINCH} & 60.6 & 46.2 & 60.6 & 40.2 & 38.5 & 40.2 \\
      \textbf{VADE} & 53.9 & 52.9 & 53.9 & 40.6 & 35.5 & 40.6 \\
      \textbf{AE + K-means} & 59.5 & 56.4 & 59.5 & 44.6 & 42.1 & 44.6 \\
      \textbf{DEC} & 55.5 & 53.6 & 55.5 & 42.2 & 40.3 & 42.2 \\
      \textbf{IDEC} & 59.8 & 56.8 & 59.8 & 44.9 & 42.3 & 44.9 \\
      \textbf{PSSC} &  68.0 & 63.6 & 68.0  & 43.2 & 41.7  & 43.2   \\
      \textbf{CC} & 57.1 & 41.6  & 57.1 & 43.4  & 42.5 & 43.4  \\
      \textbf{DDC-UCO} &  60.2 & 58.6  & 60.2 & 45.6 & 35.9  &  26.4 \\
      
      \textbf{DynAE} & \underline{63.8} & \underline{60.3} & \underline{63.8} & \underline{46.3} & \underline{43.4} & \underline{46.3} \\
      \textbf{R-DC} & \textbf{66.7} & \textbf{62.8} & \textbf{66.7} & \textbf{59.1} & \textbf{55.6} & \textbf{59.1} \\
      \hline
    \end{tabular}}
  \caption{Clustering performances in terms of ACC, F1 Macro, and F1 Micro for BreastMNIST and BloodMNIST. The best method is in bold and the second best is underlined.}
  \label{table:Breast_Blood}
  \end{small}
\end{table*}

\begin{table*}[!htbp]
   \centering
  \begin{small}
\begin{tabular}{|p{3cm}|c|c|c|c|c|c|}
    \hline
    {\textbf{Method}} & {\textbf{FMNIST}} & {\textbf{VesselMNIST3D}} & {\textbf{PneumoniaMNIST}}  \\ \hline
    \textbf{FINCH} & 26  & 2  & 2 \\ \hline
    \textbf{VaDE}  & 365 & 152 & 181 \\ \hline
    \textbf{AE+K-means}  &  108 & 28 &  118  \\ \hline
    \textbf{DEC} &  107 & 57 & 50 \\ \hline
    \textbf{IDEC} & 120 & 30 & 55\\ \hline
    \textbf{PSSC}  & 1530   & 489 &  1587 \\ \hline
    \textbf{CC} & 611  & 47 & 231 \\ \hline
    \textbf{DDC-UCO} & 3990 & 45 &  385 \\ \hline
   \textbf{DynAE} & 459 & 167 & 144 \\ \hline
    \textbf{R-DC} & 425 & 237 &  199\\ \hline
  \end{tabular}
  \end{small}
    \caption{Execution time (in seconds) of the compared approaches on FMNIST, VesselMNIST3D, and PneumoniaMNIST.}
    \label{table:exec_time}
\end{table*}


\begin{table*}
\scalebox{0.95}{ 
\begin{tabular}{|p{3.3cm}|c|c|p{4.1cm}|}
\hline
\textbf{Method} & \textbf{Computational Complexity} & \textbf{ Hyperparameters} \\ \hline
\textbf{FINCH} &  $\mathcal{O}(N \log(N))$ & 0 \\ \hline
\textbf{VaDE} & $\mathcal{O}(m \, L \, D^2 \, N_{B})$ &  4 \\ \hline
\textbf{AE + K-means} &  $\mathcal{O}(m \, L \, D^2 \, N_{B}) $ & 1 \\ \hline
\textbf{DEC} &  $\mathcal{O}(m \, L \, D^2 \, N_{B}) $ & 3 \\ \hline
\textbf{IDEC} &  $\mathcal{O}(m \, L \, D^2 \, N_{B} )$ & 4 \\ \hline
\textbf{PSSC} &  $\mathcal{O}(m \, D \, N^3)$ & 6 \\ \hline
\textbf{CC} & $\mathcal{O}(m \, L \, D^2 \, N_{B} + m \, (D + K) \, N_{B}^2) $ & 4  \\ \hline
\textbf{DDC-UCO} & $\mathcal{O}(m \, L \, D^2 \, N_{B} + m \, L \, D \, K^2 \, N^2 )$ &  4 \\ \hline
\textbf{DynAE} &  $\mathcal{O}(m \, L \, D^2 \, N_{B})$ & 3 \\ \hline
\textbf{R-DC} &  $\mathcal{O}(m \, L \, D^2 \, N_{B} \, + M \, N \, \text{log}(N))$ & 4 \\ \hline
\end{tabular}}
\centering
\caption{The computational complexity and the number of hyperparameters of our approach and the compared methods. For simplicity, we assume that the number of neurons for all layers is constant. Furthermore, we assume that $D$  $\gg$ $L$, $K$ and $N$ $\gg$ $D$, $L$, $K$.}
\label{comp_complexity}
\end{table*}

In Table \ref{table:exec_time}, we compare the runtime of R-DC with the state-of-the-art models across three datasets. In this experiment, it is important to note that all these models have been trained for the same number of iterations $1000$ for a fair comparison. In Table \ref{comp_complexity}, we provide the computational complexity and the number of hyperparameters of each approach to analyze the difference between them in runtime.  First, FINCH has a lower execution time than all the other methods. It is worthy of note that FINCH is not a deep clustering method. Thus, it does not benefit from the expressive power of deep neural networks, which makes it less effective than the other methods. Second, we observe that the execution time of R-DC is slightly higher than the execution time of AE+K-means, VaDE, DEC, IDEC, and DynAE. As R-DC uses the k-NN algorithm to compute the nearest neighbor distance matrix, it naturally requires more execution time. In particular, our approach has an additional computational complexity term $\mathcal{O}( M \, N \, \text{log}(N))$ compared with these methods. Third, we observe that the execution time of R-DC is generally lower than the execution time of PSSC, DDC-UCO, and CC. These results are supported by the computational complexity analysis, which shows quadratic and cubic complexity terms with respect to $N$ or $N_B$ for PSSC, DDC-UCO, and CC. Finally, the number of hyperparameters in our approach does not exceed the number of hyperparameters of several state-of-the-art deep clustering methods such as DDC-UCO, CC, PSSC, IDEC, and VaDE, as provided in Table \ref{comp_complexity}.

\begin{figure}
  \centering
  \begin{subfigure}[b]{0.45\textwidth}
     \includegraphics[width=\linewidth]{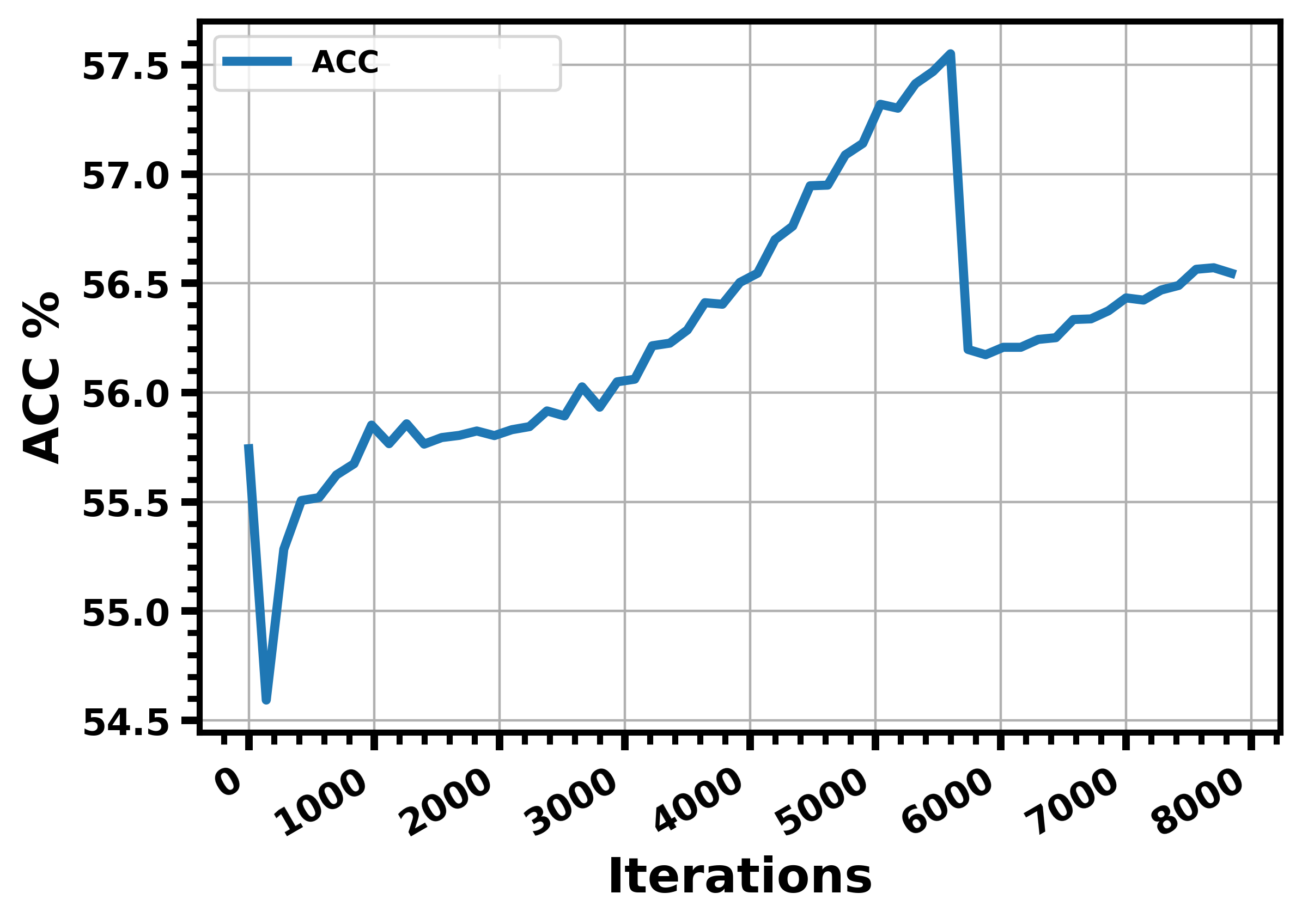}
     \caption{ACC: DynAE}
  \end{subfigure}\hfill
  \begin{subfigure}[b]{0.45\textwidth}
     \includegraphics[width=\linewidth]{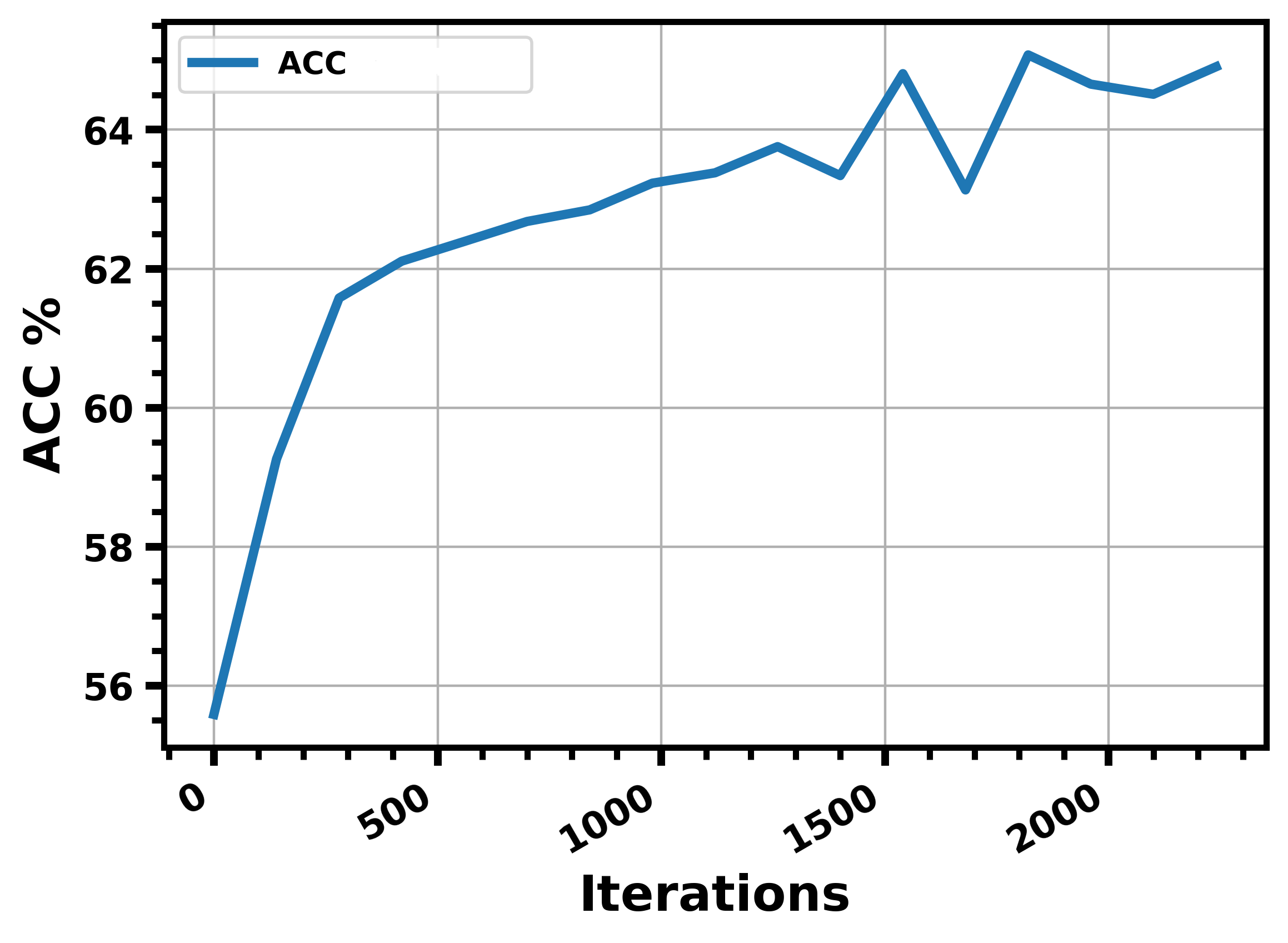}
     \caption{ACC: R-DC}
  \end{subfigure}
  
  \caption{Learning dynamics of DynAE and R-DC on FMNIST.}
  \label{fig:lr_truefalse1}
\end{figure}

Figure \ref{fig:lr_truefalse1} illustrates the learning dynamics observed during the second phase of DynAE and R-DC on FMNIST. It is important to highlight that both models not only use the same architecture and experimental settings but also undergo identical pretraining phases. As we can see, both models start the second phase from the same accuracy level of 55.8\%. The ACC for R-DC increases nearly in a consistent way during the training process as opposed to the evolution of ACC for DynAE. The results of R-DC and DynAE show that eliminating pseudo-supervision enhances clustering performance. 

\begin{figure}
  \centering
  \begin{subfigure}[b]{0.45\textwidth}
     \includegraphics[width=\linewidth]{ACC_fmnist_1.png}
     \caption{ACC (FMNIST)}
  \end{subfigure}\hfill
  \begin{subfigure}[b]{0.45\textwidth}
     \includegraphics[width=\linewidth]{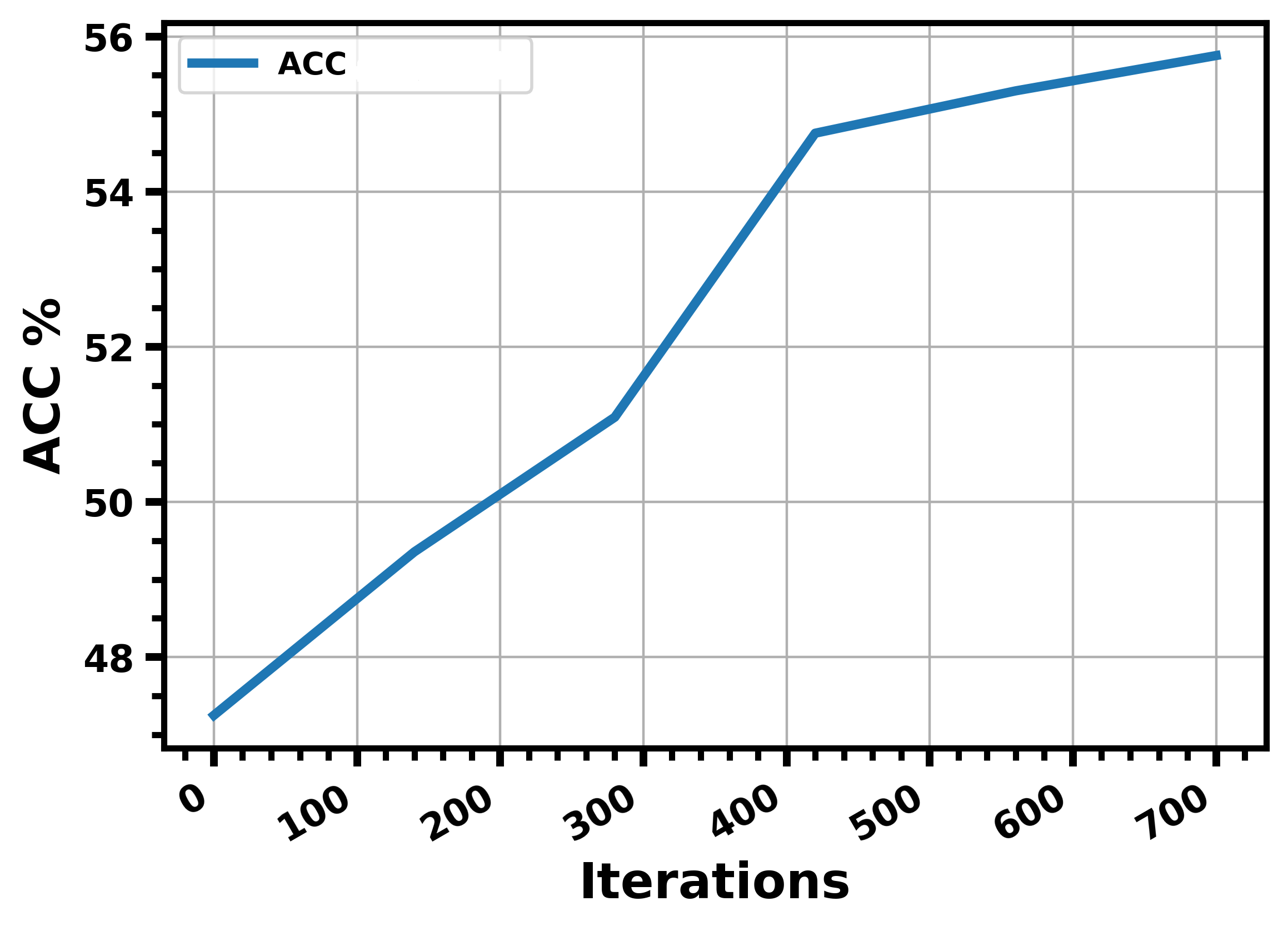}
     \caption{ACC (BloodMNIST)}
  \end{subfigure}
  
  \caption{Learning dynamics of R-DC on FMNIST and BloodMNIST (ACC).}
  \label{fig:ACC_FMNIST_BLOODMNIST}
\end{figure}

\begin{figure}
  \centering

  \begin{subfigure}[b]{0.45\textwidth}
     \includegraphics[width=\linewidth]{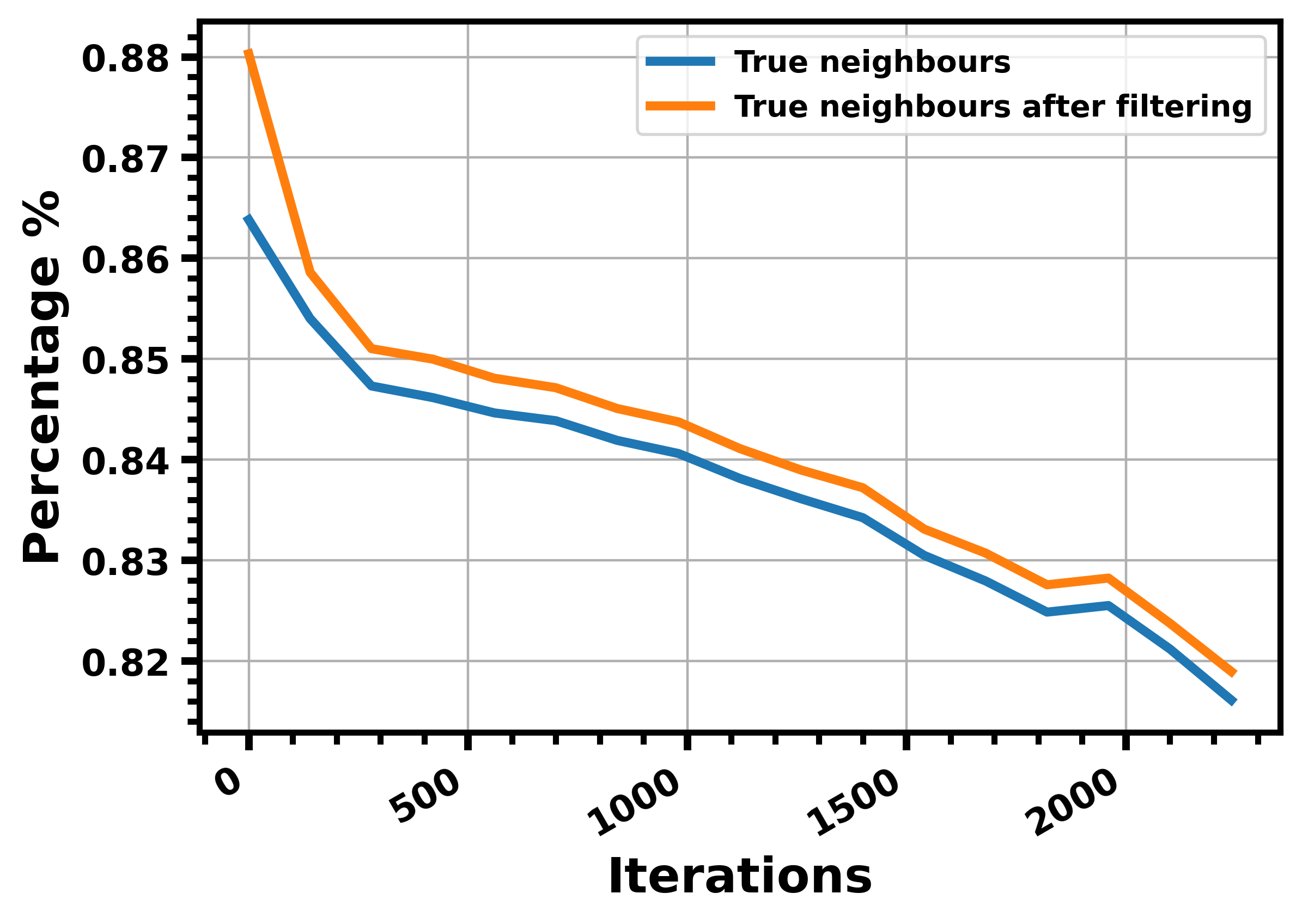}
     \caption{\centering\% of true neighbors among NN (FMNIST)}
  \end{subfigure}\hfill
  \begin{subfigure}[b]{0.45\textwidth}
    \includegraphics[width=\linewidth]{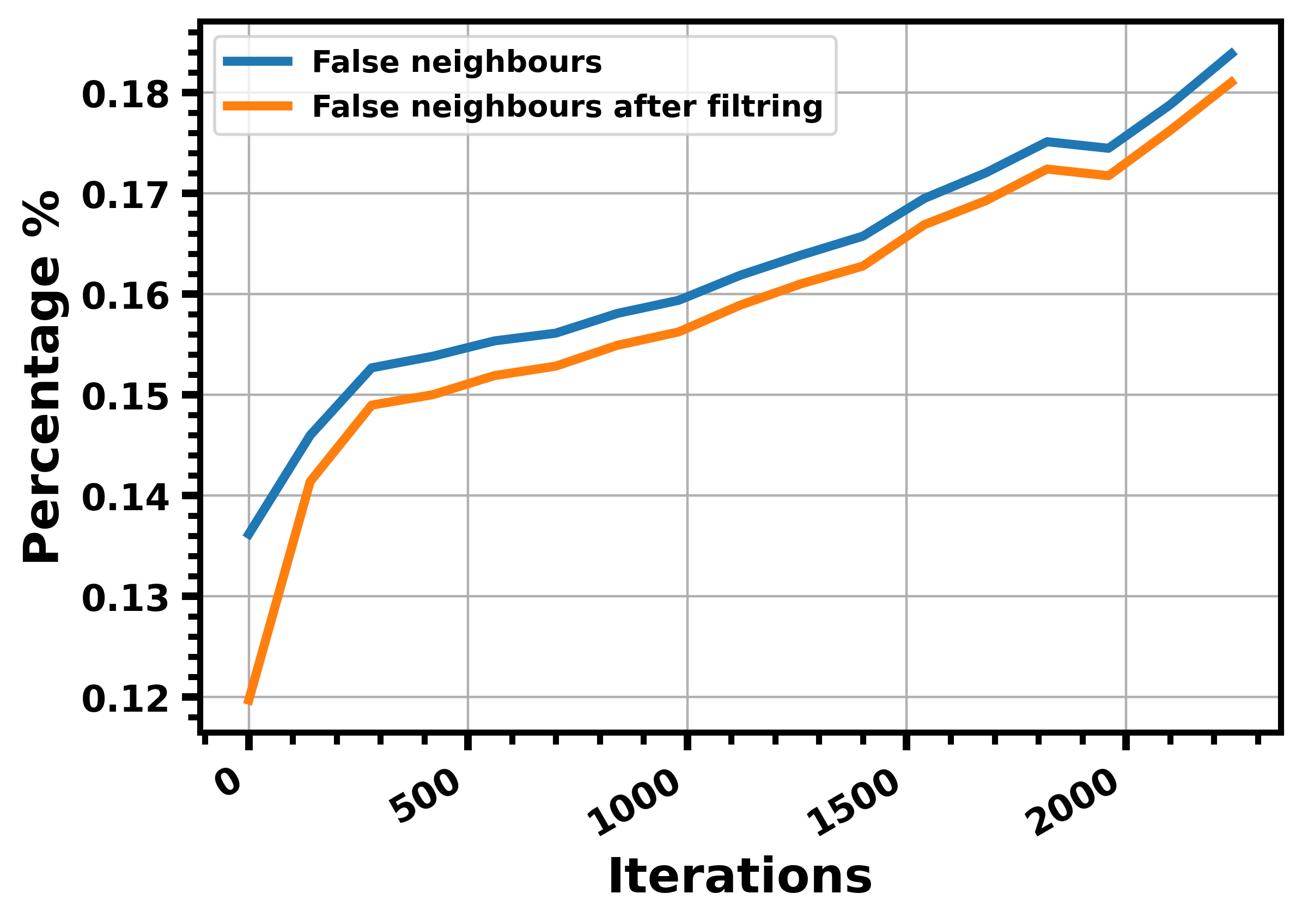}
    \caption{\centering\% of false neighbors among NN (FMNIST)}
  \end{subfigure}
  
  \medskip

  \begin{subfigure}[b]{0.45\textwidth}
     \includegraphics[width=\linewidth]{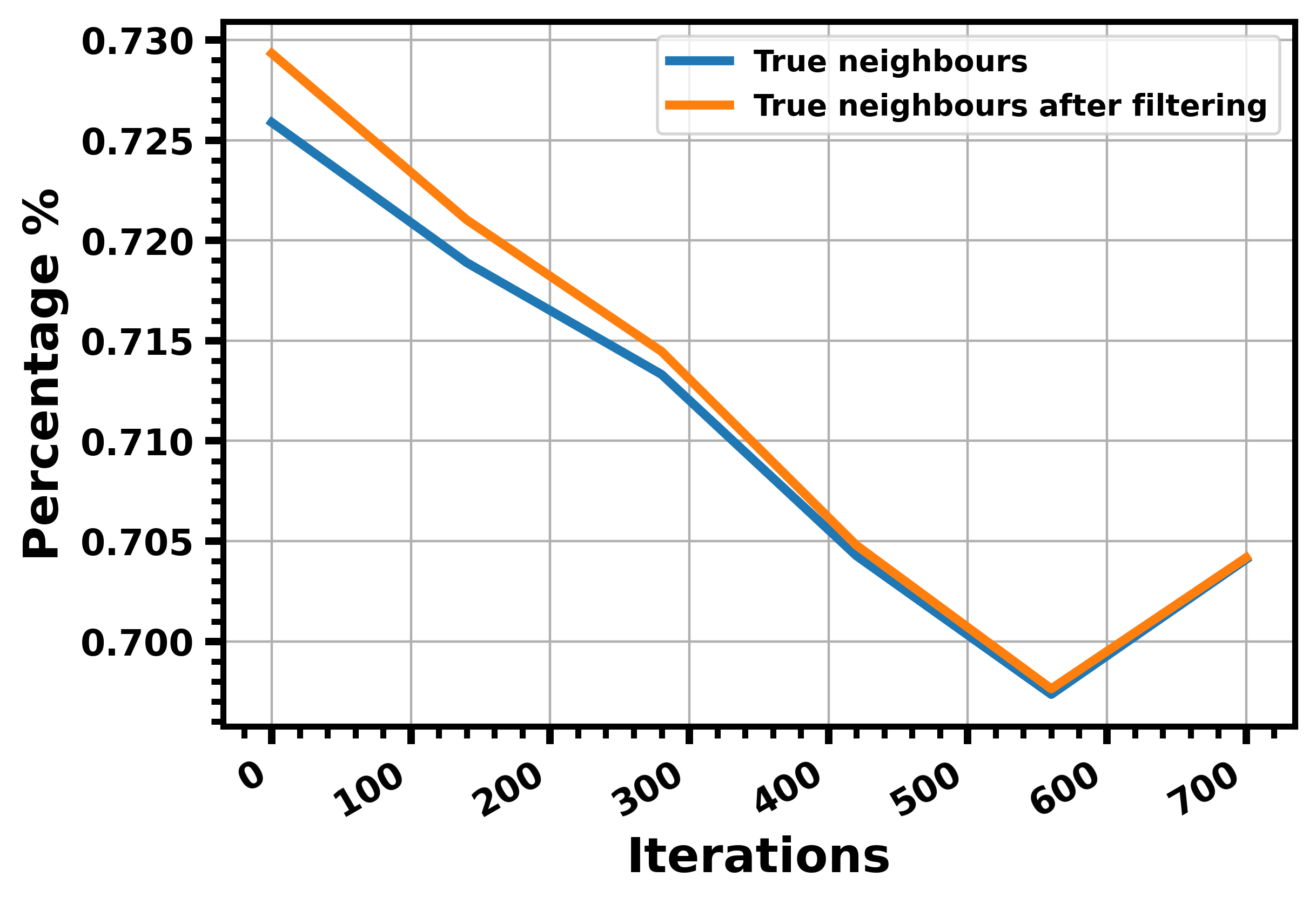}
     \caption{\centering\% of true neighbors among NN (BloodMNIST)}
  \end{subfigure}\hfill
  \begin{subfigure}[b]{0.45\textwidth}
    \includegraphics[width=\linewidth]{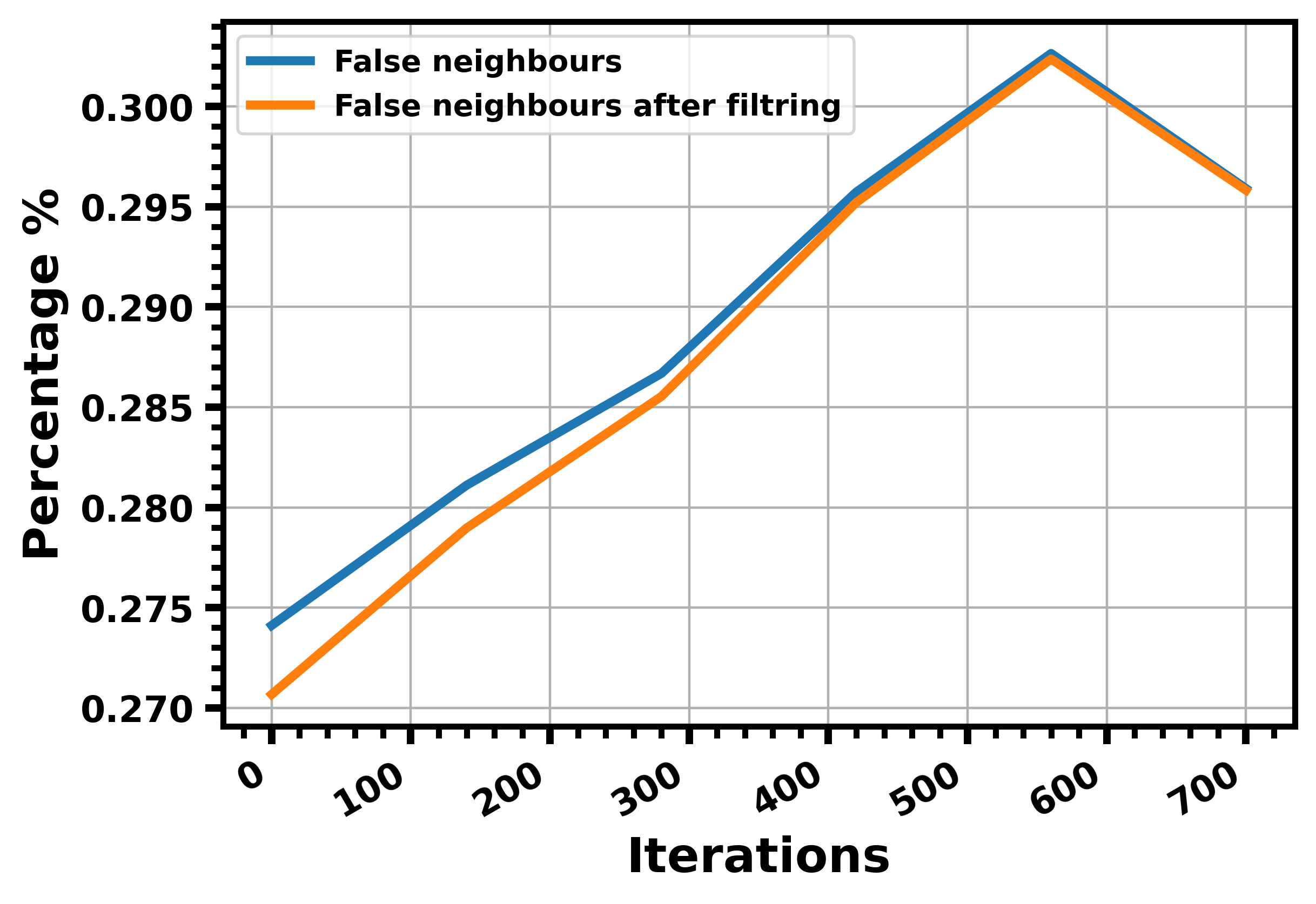}
    \caption{\centering\% of false neighbors among NN (BloodMNIST)}
\end{subfigure}
 
  \caption{Learning dynamics of R-DC on FMNIST and BloodMNIST. NN denotes Nearest Neighbours.}
  \label{fig:lr_truefalse}
\end{figure}

In Figure \ref{fig:ACC_FMNIST_BLOODMNIST} and Figure \ref{fig:lr_truefalse}, we illustrate the learning dynamics of R-DC on FMNIST and BloodMNIST. Specifically, Figure \ref{fig:ACC_FMNIST_BLOODMNIST}.a and Figure \ref{fig:ACC_FMNIST_BLOODMNIST}.b present the evolution of ACC. Moreover, Figures \ref{fig:lr_truefalse}.a and \ref{fig:lr_truefalse}.c provide insights into the behavior of the true nearest neighbors in R-DC. The true neighbors of a latent code $z_i$ of a core point represent the neighbors that belong to the same cluster as this core point. The percentage of initial true neighbors and true neighbors after filtering is plotted against the number of iterations. It is observed that the number of true neighbors after filtering surpasses the count of initial true neighbors. This observation confirms our hypothesis that the most reliable nearest neighbors of a core point in the latent space are more likely to belong to the same cluster than the other neighbors. Additionally, there is a gradual decline in the percentage of true neighbors after filtering, albeit starting from a higher value. Our algorithm gradually expands the set of core points, \(\Omega\). As more points infiltrate the set of core points, \(\Omega\), the reliability of these added core points is lower compared to the reliability of the initial core points. In other words, the number of true neighbors among the \(M\) nearest neighbors for the added core points is lower than the number of true neighbors among the \(M\) nearest neighbors for the initial core points. Although the absolute number of true neighbors increases by adding new core points, the percentage of true neighbors may decrease. In our case, it is reasonable to expect a slight decrease in the overall percentage of true neighbors during the training process due to the lower quality of the added core points compared to the initial ones. This is opposite to the trend observed for false neighbors, as shown in Figures \ref{fig:lr_truefalse}.b and \ref{fig:lr_truefalse}.d.

\subsubsection{Geometric Study}
\begin{figure}
  \centering
\begin{subfigure}[b]{0.4\textwidth}
    \includegraphics[width=\linewidth]{fmnist_32_0.82_2000.png}
    \caption{DynAE}
  \end{subfigure} \hfil
   \begin{subfigure}[b]{0.4\textwidth}
    \includegraphics[width=\linewidth]{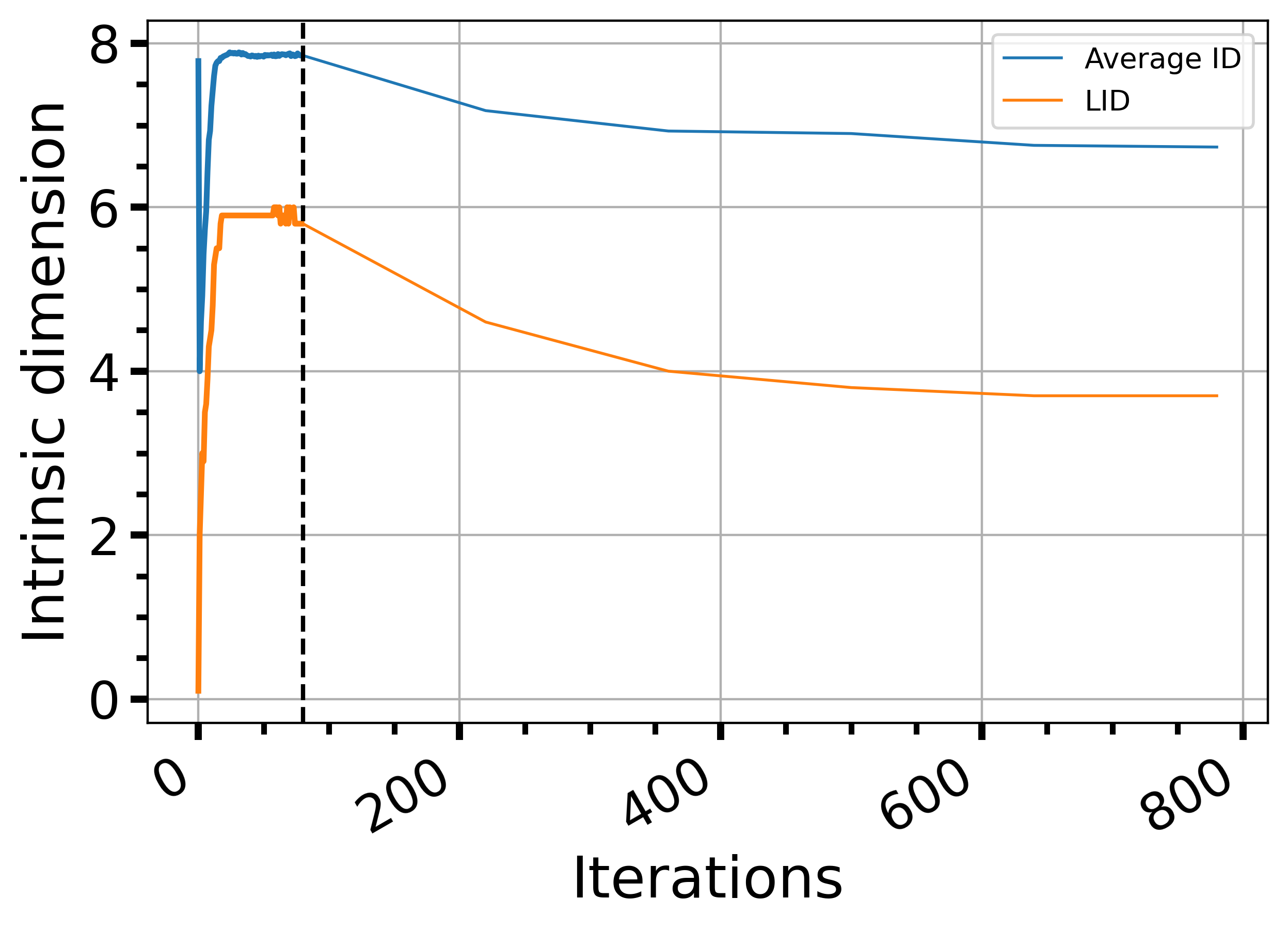}
    \caption{R-DC}
  \end{subfigure} \hfil
 \caption{\textbf{First evidence of alleviating Feature Twist:} preserving the difference between ID and LID.}
  \label{fig:FT_1}
\end{figure}

In Figure \ref{fig:FT_1}, we examine the training process of R-DC and DynAE on FMNIST from a geometric perspective based on ID and LID. As we can see, the training process of DynAE shows an abrupt transition in ID and LID from the first to the second phase. We observe that LID and ID converge to the same values, which means that the pseudo-supervision task flattens the latent manifolds. As opposed to that, our model exhibits a smooth transition in ID and LID throughout the two phases. Furthermore, the two curves of ID and LID do not meet, which means that the manifold structure is preserved. Our results provide strong evidence that R-DC, which does not involve a transition from self-supervision to pseudo-supervision, alleviates the FT problem.

\begin{figure*}[!htbp]
\begin{subfigure}[b]{0.16\textwidth}
    \includegraphics[width=\linewidth]{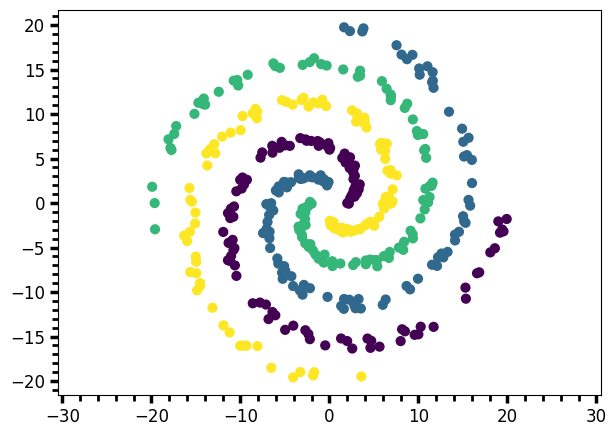}
    \caption{Input data}
\end{subfigure}
\begin{subfigure}[b]{0.16\textwidth}
    \includegraphics[width=\linewidth]{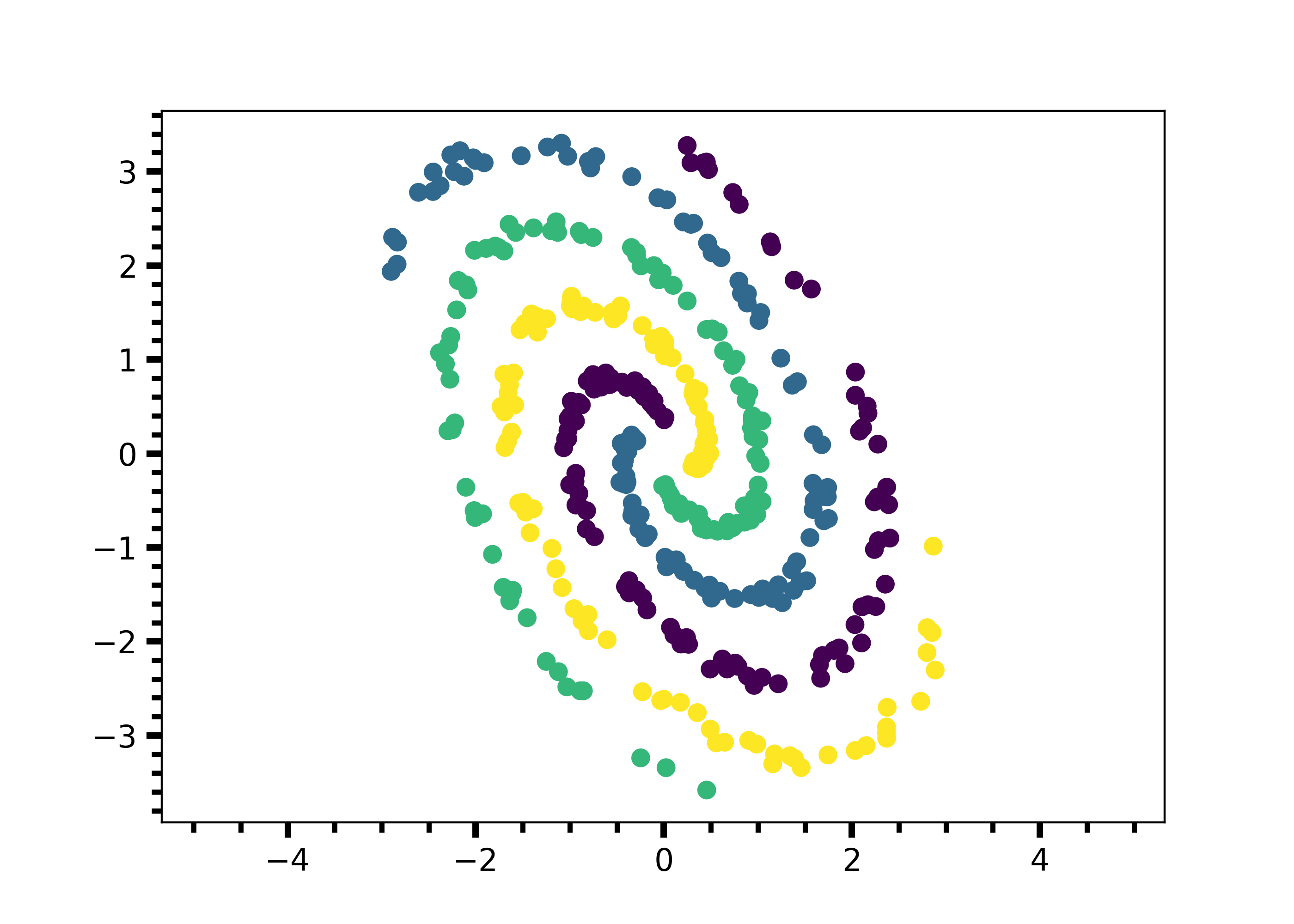}
    \caption{Epoch $200$}
\end{subfigure}
 \begin{subfigure}[b]{0.16\textwidth}
     \includegraphics[width=\linewidth]{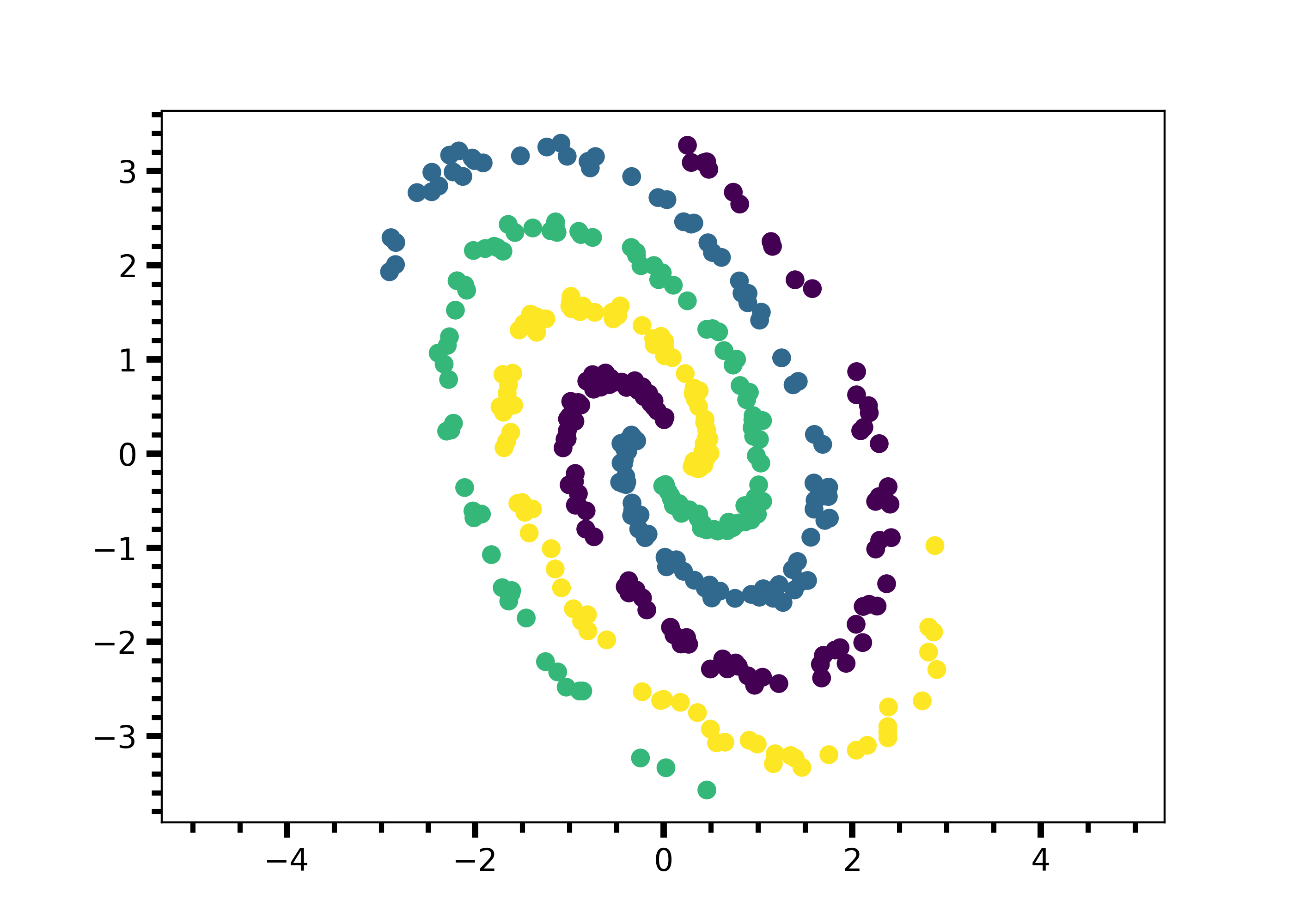}
     \caption{Epoch $225$}
 \end{subfigure}
  \begin{subfigure}[b]{0.16\textwidth}
     \includegraphics[width=\linewidth]{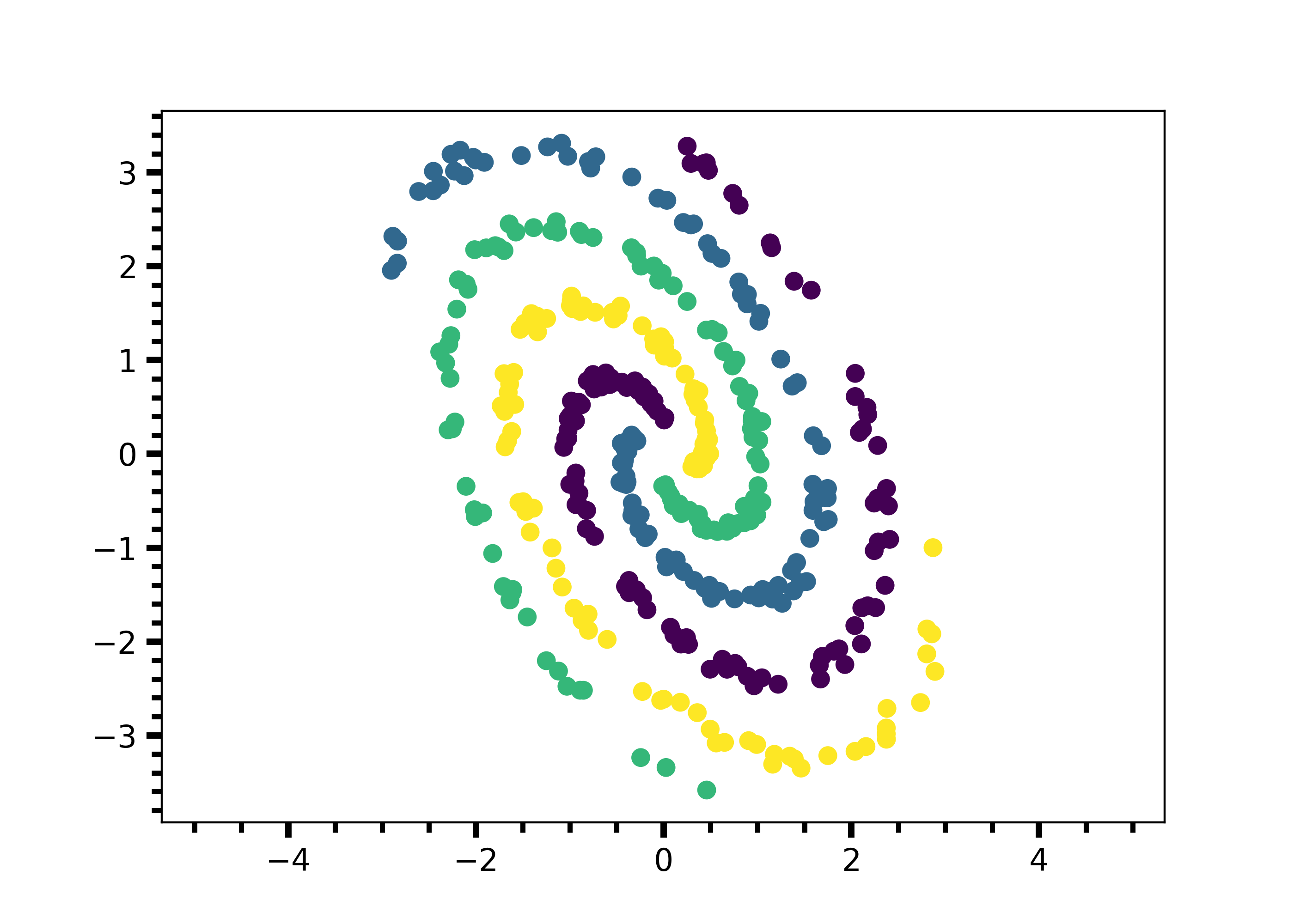}
     \caption{Epoch $250$}
 \end{subfigure}
  \begin{subfigure}[b]{0.16\textwidth}
     \includegraphics[width=\linewidth]{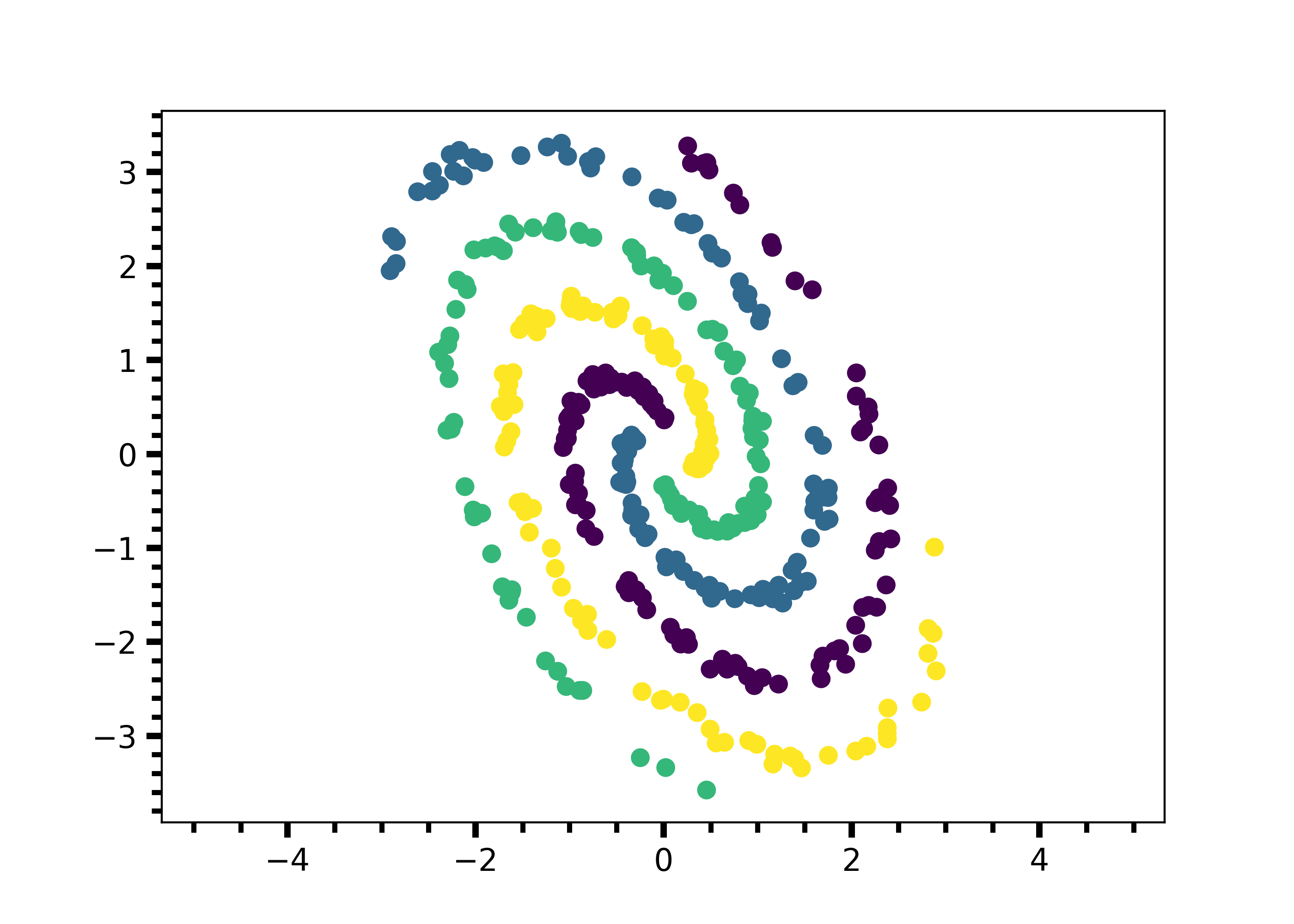}
     \caption{Epoch $275$}
\end{subfigure}
\begin{subfigure}[b]{0.16\textwidth}
    \includegraphics[width=\linewidth]{emb_275.png}
    \caption{Epoch $300$}
\end{subfigure}
\caption{\textbf{Second evidence of alleviating Feature Twist:} preserving the geometric latent structures in a synthetic dataset with curved manifolds.}
\label{fig:FT_2_RDC}
\end{figure*}

To further validate the impact of our approach from a geometric perspective, we conduct experiments on a 2D synthetic dataset with four curved clusters. We adopt the same architecture described in the experiments outlined in Figure \ref{fig:FT_2}. In this context, we leverage a linear two-layer fully-connected auto-encoder to map the data onto a 2D latent space. Our model undergoes a pretraining phase with vanilla reconstruction for 200 epochs, followed by a fine-tuning phase based on the loss function of R-DC (Eq. \ref{eq:total_loss}). The visualizations of the embedded space provided in Figure \ref{fig:FT_2_RDC} reveal that the second phase of R-DC effectively preserves the structure of the latent manifolds without inducing the geometric distortions typically associated with twisting and collapsing the clustering structures, as previously observed in Figure \ref{fig:FT_2}.

\subsubsection{Ablation Study}

For the ablation study, we introduce a third phase that performs pseudo-supervision using the DEC clustering loss. We report the clustering performance on different combinations of the training phases. In a nutshell, the first phase ($P_1$) is the instance-level self-supervision phase; the second phase ($P_2$) is the proximity-level self-supervision phase; the third phase ($P_3$) is the pseudo-supervision phase based on the DEC loss. The third phase is considered to see the potential impact of pseudo-supervision if applied.

In Table \ref{Table:ablation_study}, we present the outcomes of the ablation experiments carried out across four datasets to elucidate the significance of our contributions. First, we observe that performing $P_1$ $\&$ $P_2$ gives better results than $P_1$ alone. In particular, introducing the second phase $P_2$ brings a considerable improvement in clustering performance compared to the second phase of the existing DC paradigms as illustrated earlier in Figure \ref{fig:resulta}. The highest improvement is shown in BloodMnist, which is 11.9\% in terms of accuracy, followed by 11\% in FMNIST, and 3.6\% and 2.5\% in PneumoniaMNIST and BreastMNIST, respectively. This confirms that our strategy in R-DC not only achieves promising results but also brings a significant improvement by introducing proximity-level self-supervision, as evidenced by the notable increase in results during the second phase. Second, we observe that introducing a pseudo-supervision phase $P_3$ after the first phase $P_1$, as illustrated by the results of the model annotated $P_1$ \& $P_3$, does not bring \textit{consistent} improvement in clustering results compared with $P_1$ alone. These results can be explained by the effect of the false pseudo-labels and the geometric distortions caused by the abrupt transition from self-supervision to pseudo-supervision (i.e., the occurrence of FR and FT). Furthermore, we observe that combining $P_1$ \& $P_2$ \& $P_3$ brings consistent improvement compared with the model that performs only $P_1$ \& $P_3$. This improvement can be explained by the impact of adding the $P_2$ phase, which ensures a smoother geometric transition and alleviates the FT problem. Despite the improvement achieved by combining $P_1$ \& $P_2$ \& $P_3$, the pseudo-supervision task remains problematic. Therefore, by eliminating $P_3$, we can see that the results of our model, which combines $P_1$ \& $P_2$, are better than the results of the model that performs $P_1$ \& $P_2$ \& $P_3$. By discarding the third phase, we can also relieve the execution time overhead. As a key finding, our results provide strong evidence that pseudo-supervision is not required at all. 

\begin{table}[!ht]
  \begin{center}
  \begin{small}
  \scalebox{0.90}{\begin{tabular}{|c|c|c|c|c|c|}
    \hline
    \textbf{Dataset} & \textbf{Metrics} & \textbf{$P_1$} & \textbf{$P_1$} \& \textbf{$P_2$} & \textbf{$P_{1}$} \& \textbf{$P_{3}$} & \textbf{$P_{1}$} \& \textbf{$P_{2}$} \&  \textbf{$P_{3}$} \\
    \hline
            &  \textbf{ACC} & 55.6 & \textbf{66.6} & 58.8 & 63.8 \\ 
    \textbf{FMNIST}  & \textbf{F1-Macro} & 53.2 & \textbf{65.7} & 58.1 & 62.3 \\ 
            & \textbf{F1-Micro} & 55.6 & \textbf{66.6} & 58.8 & 63.8 \\ \hline 
            & \textbf{ACC} & 82.9 & \textbf{86.5} & 63.3 & 83.2\\ 
    \textbf{PneumoniaMNIST} & \textbf{F1-Macro} & 79.9 & \textbf{83.7} & 62.3 & 79.1   \\ 
              & \textbf{F1-Micro} & 82.9 & \textbf{86.5} & 63.3 & 83.2 \\  \hline 
              & \textbf{ACC} & 47.2 & \textbf{59.1} & 48.2 & 48.3 \\
    \textbf{BloodMNIST} & \textbf{F1-Macro} & 47.7 & \textbf{55.6} & 49.5 & 49.5 \\ 
                & \textbf{F1-Micro} & 47.2 & \textbf{59.1} & 48.2 & 48.3 \\ \hline 
                  & \textbf{ACC} & 64.2  & \textbf{66.7} & 53.1 & 58.2 \\ 
    \textbf{BreastMNIST} & \textbf{F1-Macro} & 60.9 & \textbf{62.8} & 44.0 & 57.8  \\ 
                & \textbf{F1-Micro} & 64.2 & \textbf{66.7}  &  53.1 & 58.2 \\ \hline 
  \end{tabular}}
  \end{small}
  \end{center}
  \caption{Ablation study of R-DC. $P_{i}$ denotes the $i^{th}$ phase.}
  \label{Table:ablation_study}
\end{table}

\begin{figure}
  \centering
  
  \begin{subfigure}[b]{0.3\textwidth}
    \includegraphics[width=\linewidth]{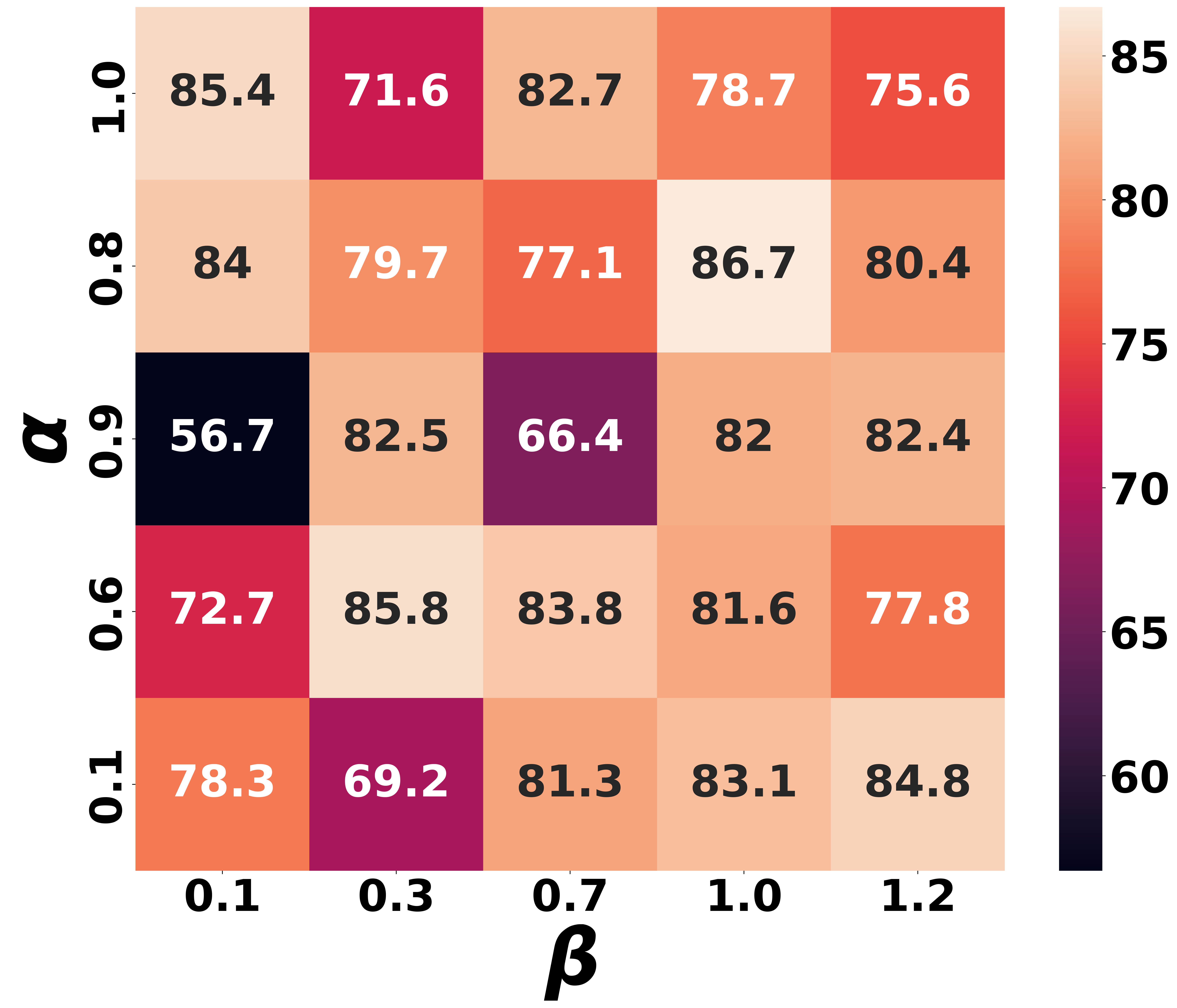}
    \caption{ACC}
  \end{subfigure}\hfill
  \begin{subfigure}[b]{0.3\textwidth}
    \includegraphics[width=\linewidth]{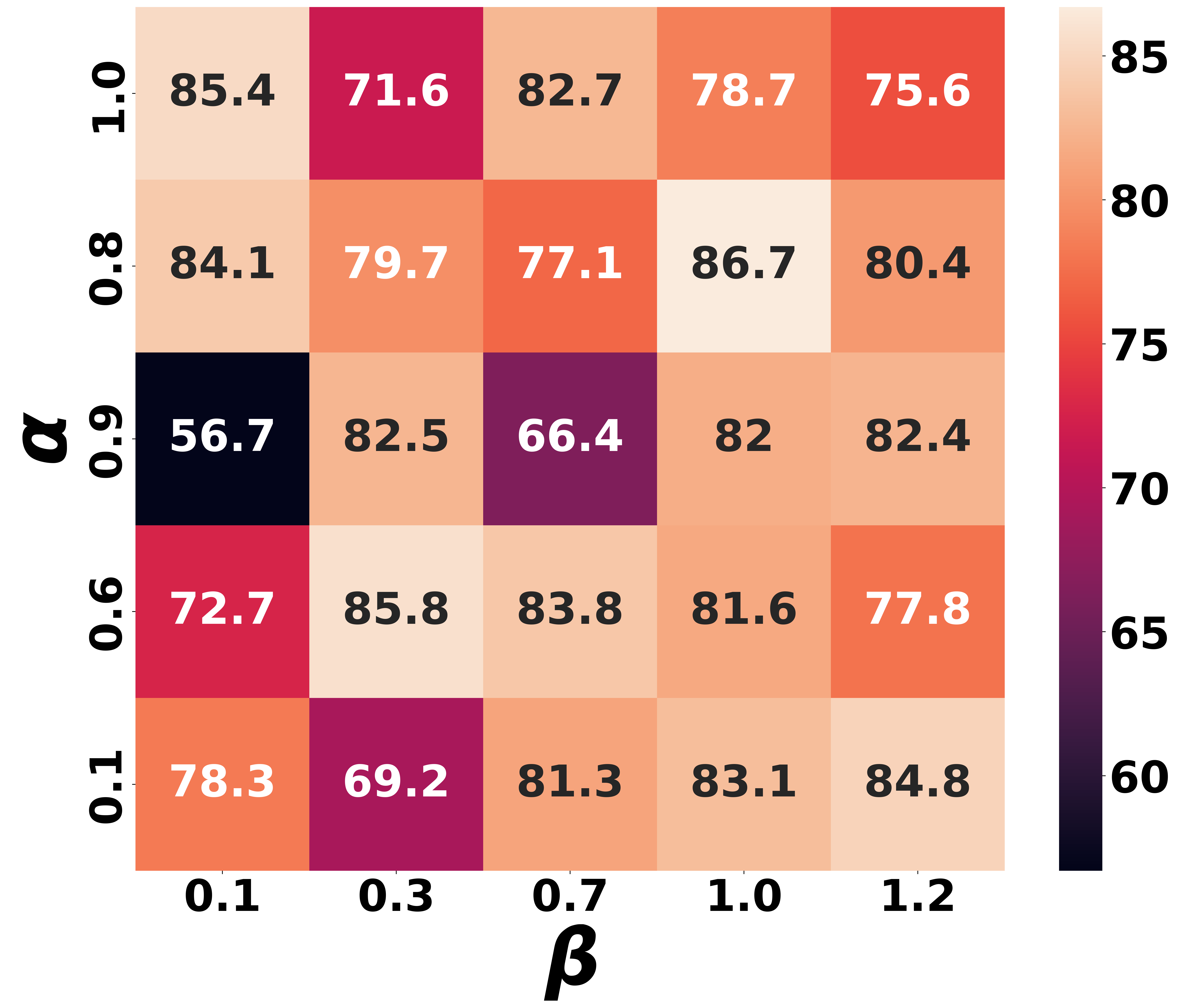}
    \caption{F1 Micro Score}
  \end{subfigure}\hfill
  \begin{subfigure}[b]{0.3\textwidth}
    \includegraphics[width=\linewidth]{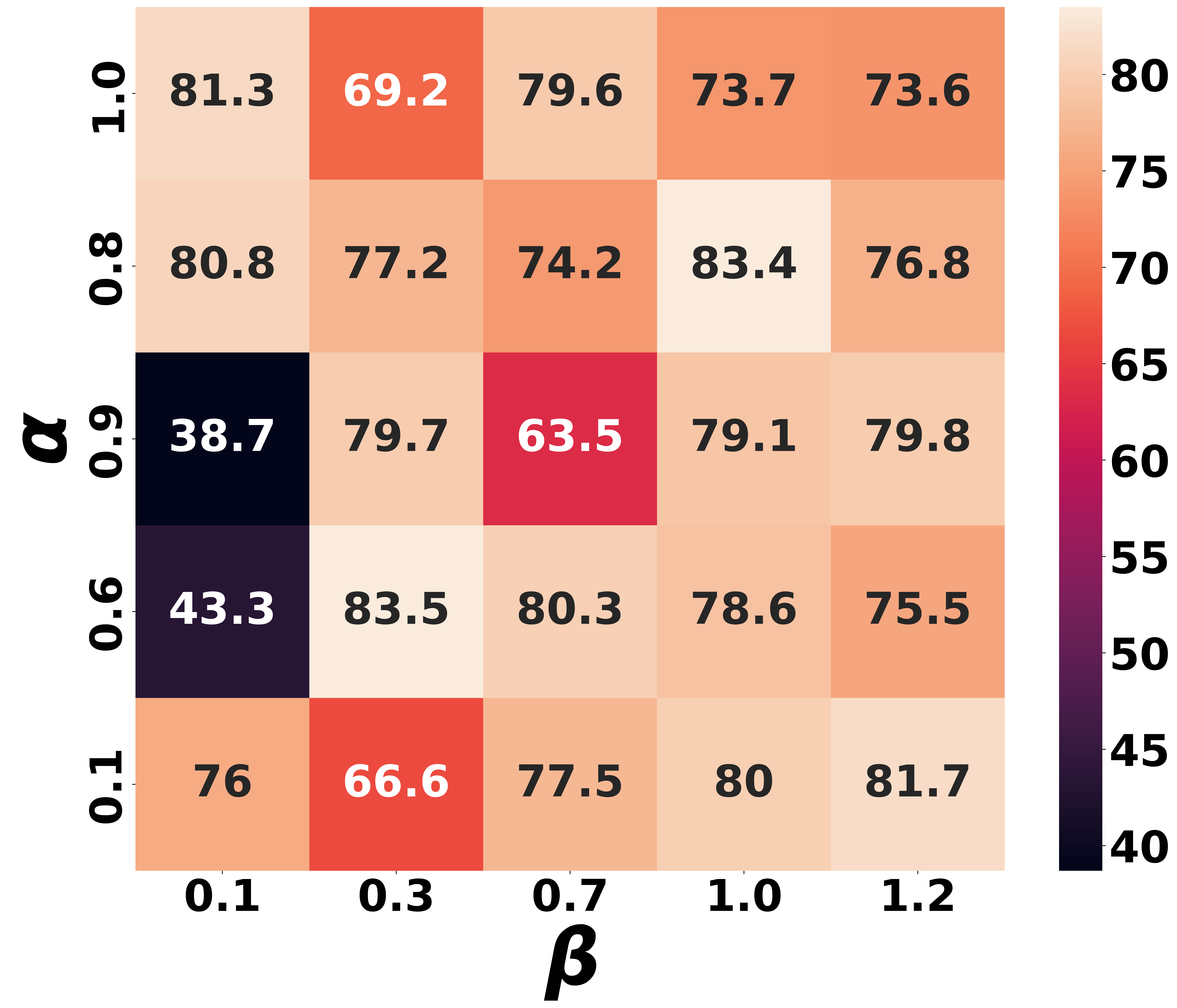}
    \caption{F1 Macro Score}
  \end{subfigure}
  
  \caption{Sensitivity of R-DC to $\alpha$ and $\beta$ in terms of ACC and F1 Score. Results on PneumoniaMNIST.}
  \label{fig:lr_dyn_R-GMM-VGAE}
\end{figure}

\begin{figure}
  \centering
  
  \begin{subfigure}[b]{0.3\textwidth}
    \includegraphics[width=\linewidth]{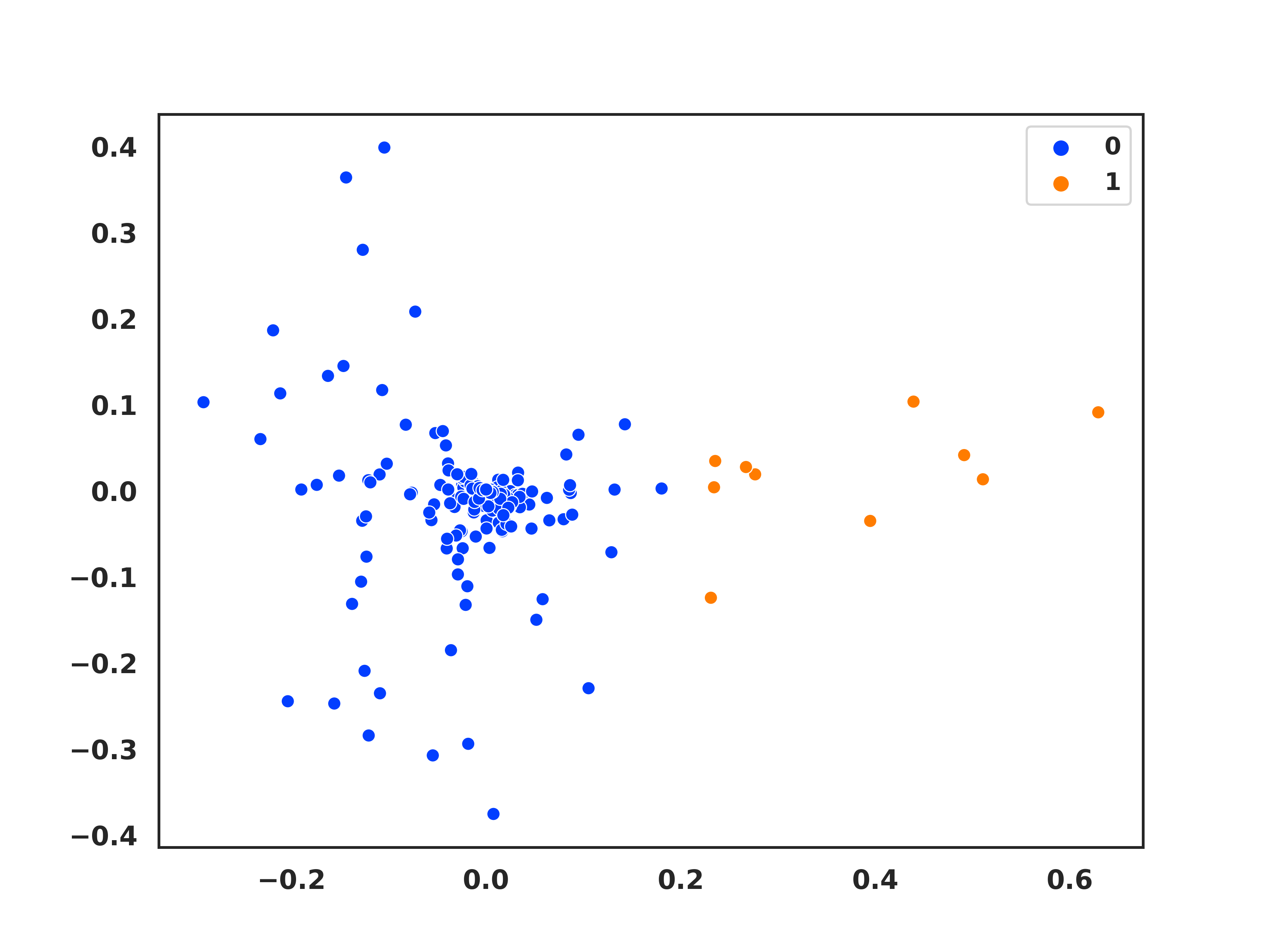}
    \caption{VesselMNIST}
  \end{subfigure}\hfill
  \begin{subfigure}[b]{0.3\textwidth}
    \includegraphics[width=\linewidth]{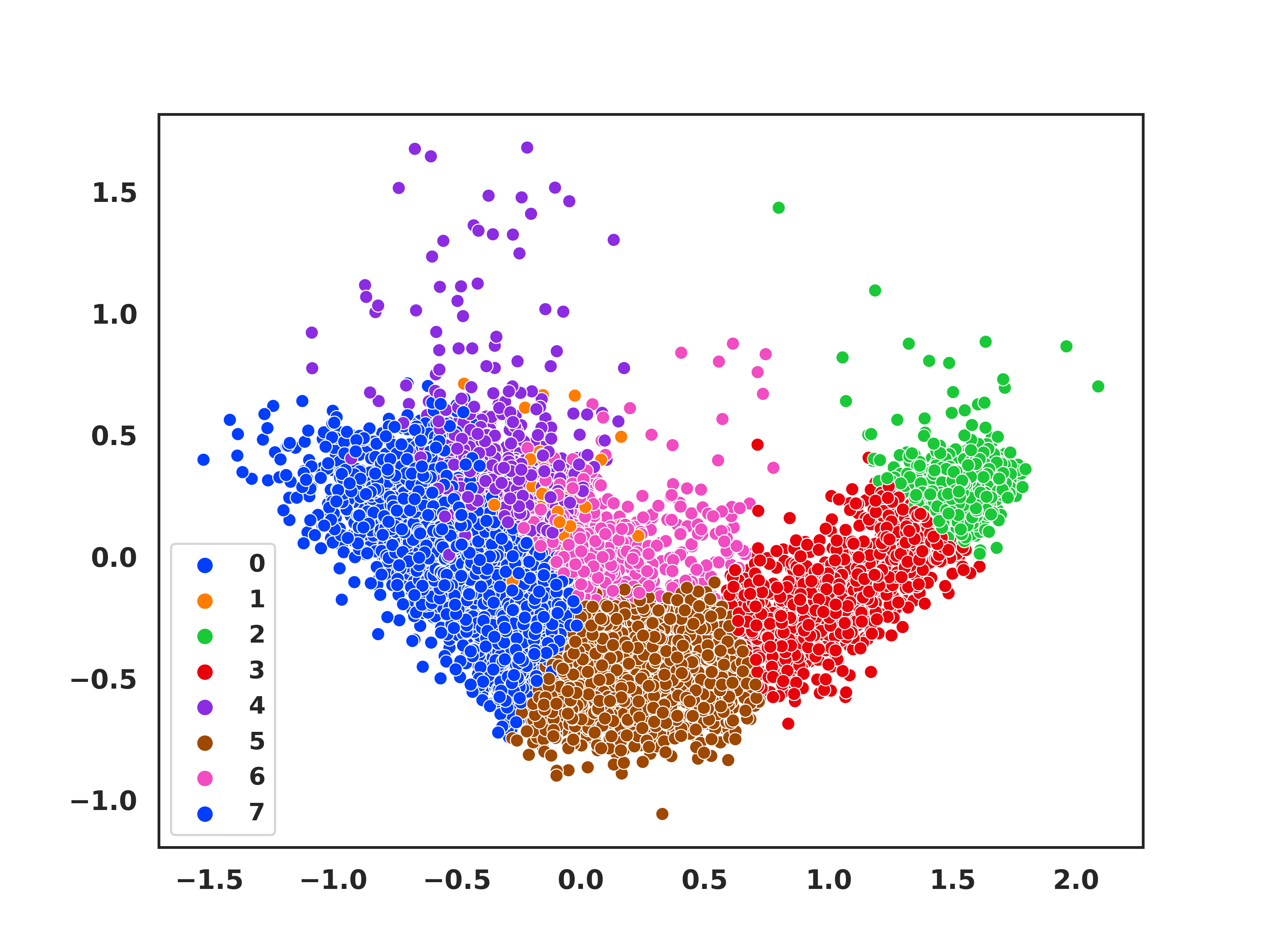}
    \caption{BloodMNIST}
  \end{subfigure}\hfill
  \begin{subfigure}[b]{0.3\textwidth}
   \includegraphics[width=\linewidth]{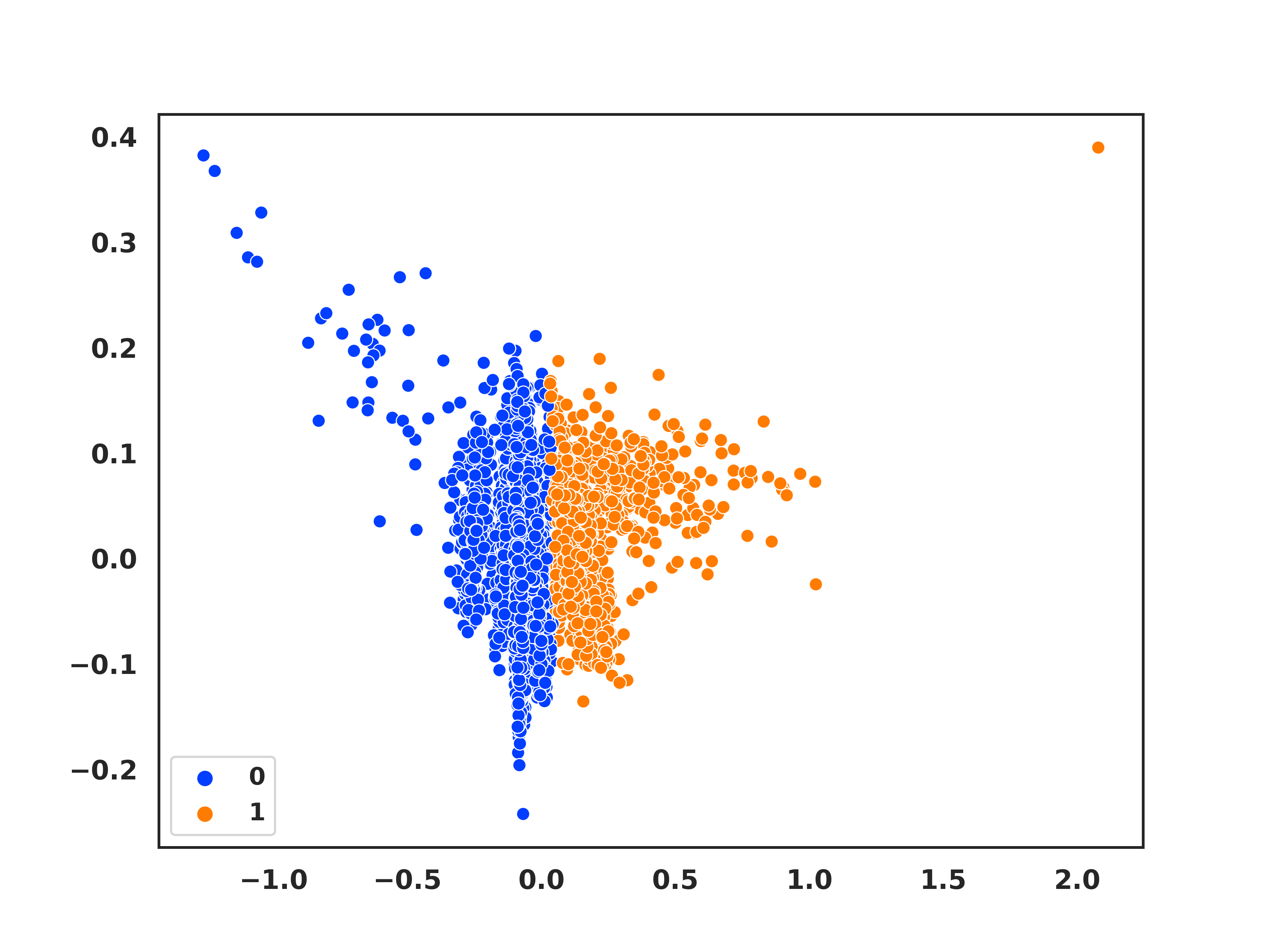}
    \caption{PneumoniaMNIST2D}
  \end{subfigure}
  
  \medskip
  
  \begin{subfigure}[b]{0.3\textwidth}
    \includegraphics[width=\linewidth]{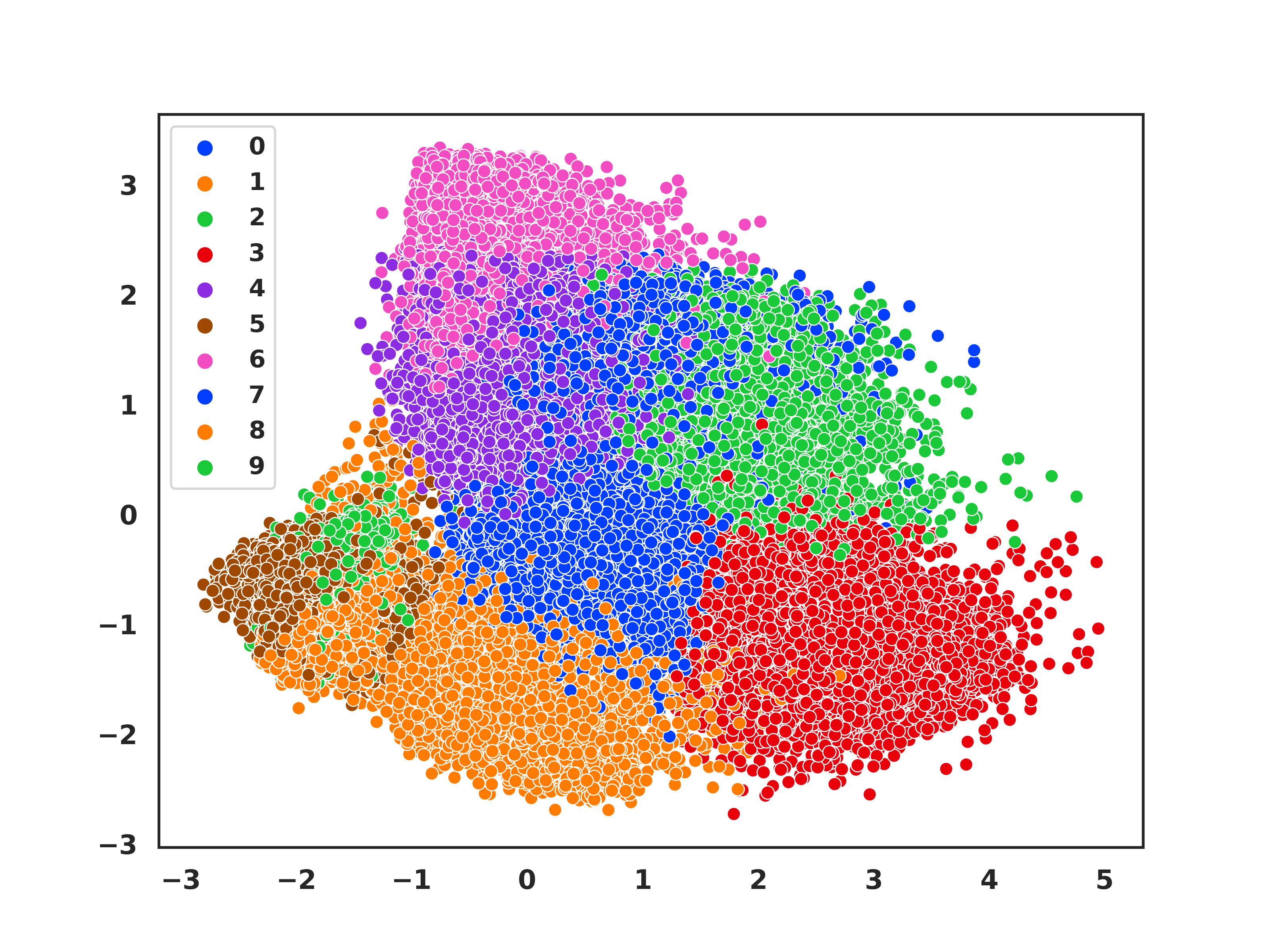}
    \caption{FMNIST}
  \end{subfigure}\hfill
  \begin{subfigure}[b]{0.3\textwidth}
    \includegraphics[width=\linewidth]{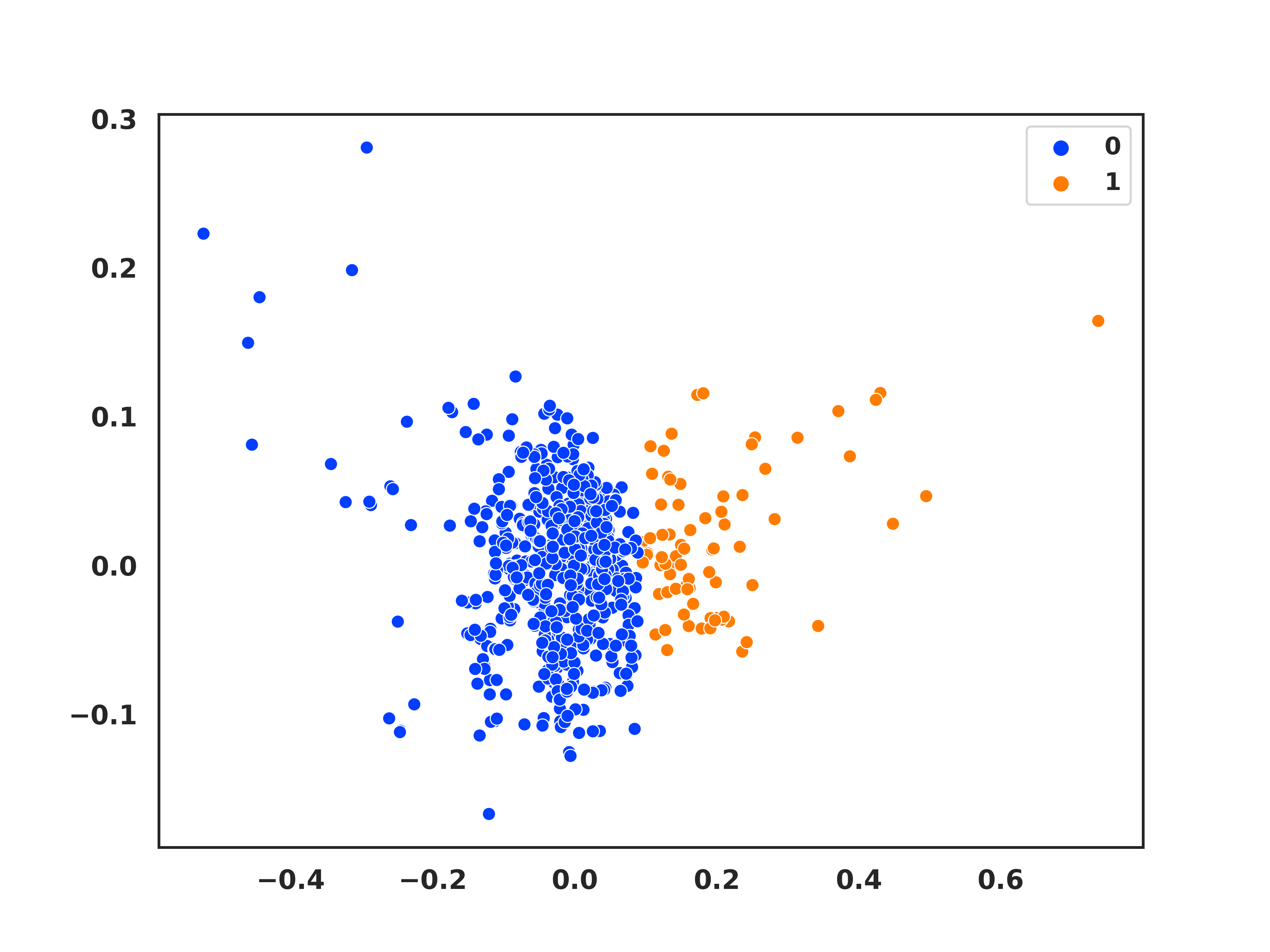}
    \caption{BreastMNIST}
  \end{subfigure}\hfill
  \begin{subfigure}[b]{0.3\textwidth}
    \includegraphics[width=\linewidth]{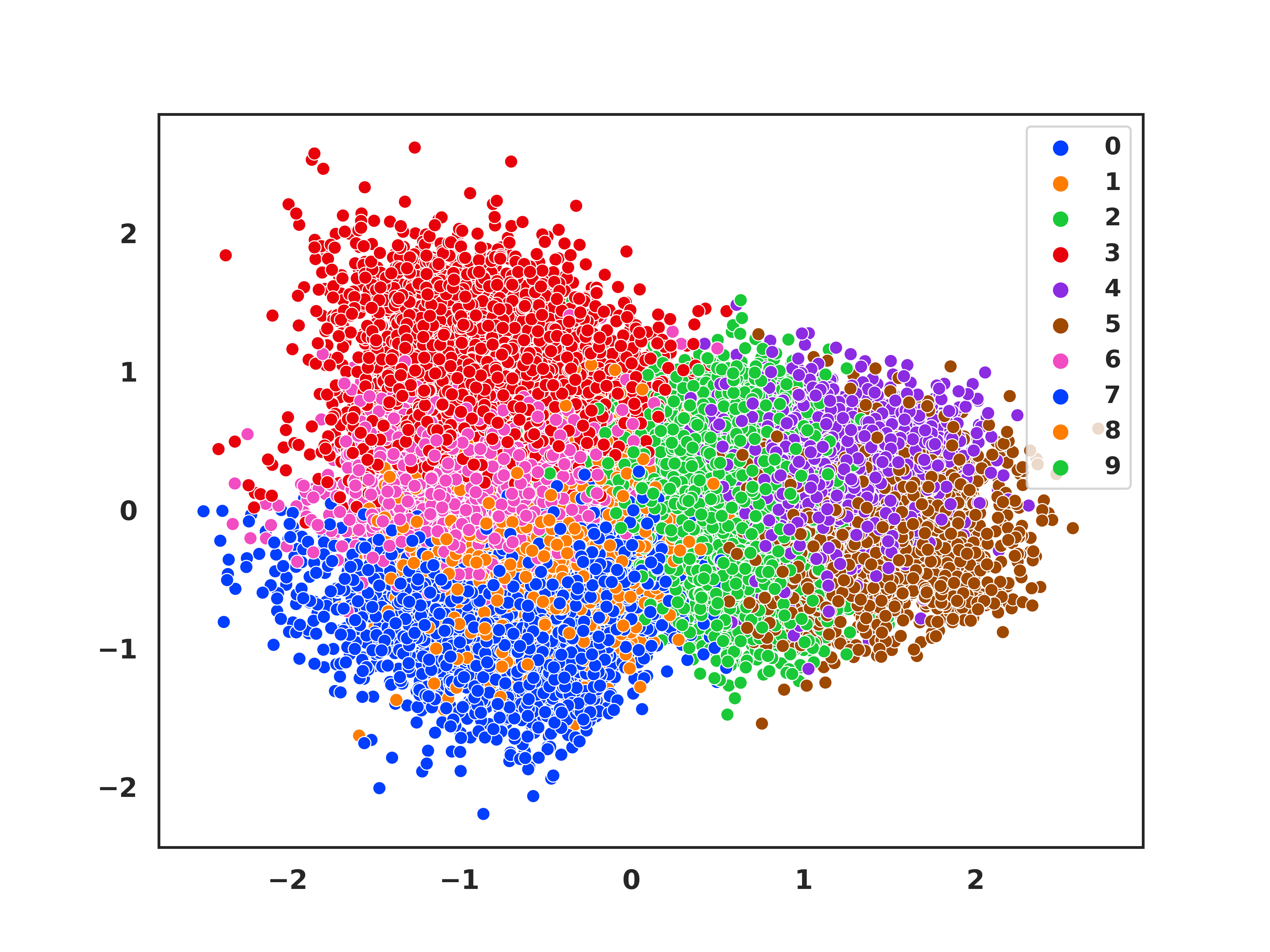}
    \caption{CIFAR10}
  \end{subfigure}
  
  \caption{Visualizations of the latent representations of R-DC after performing PCA.}
  \label{fig:lr_RDC}
\end{figure}

\begin{figure}
  \centering
  
  \begin{subfigure}[b]{0.3\textwidth}
    \includegraphics[width=\linewidth]{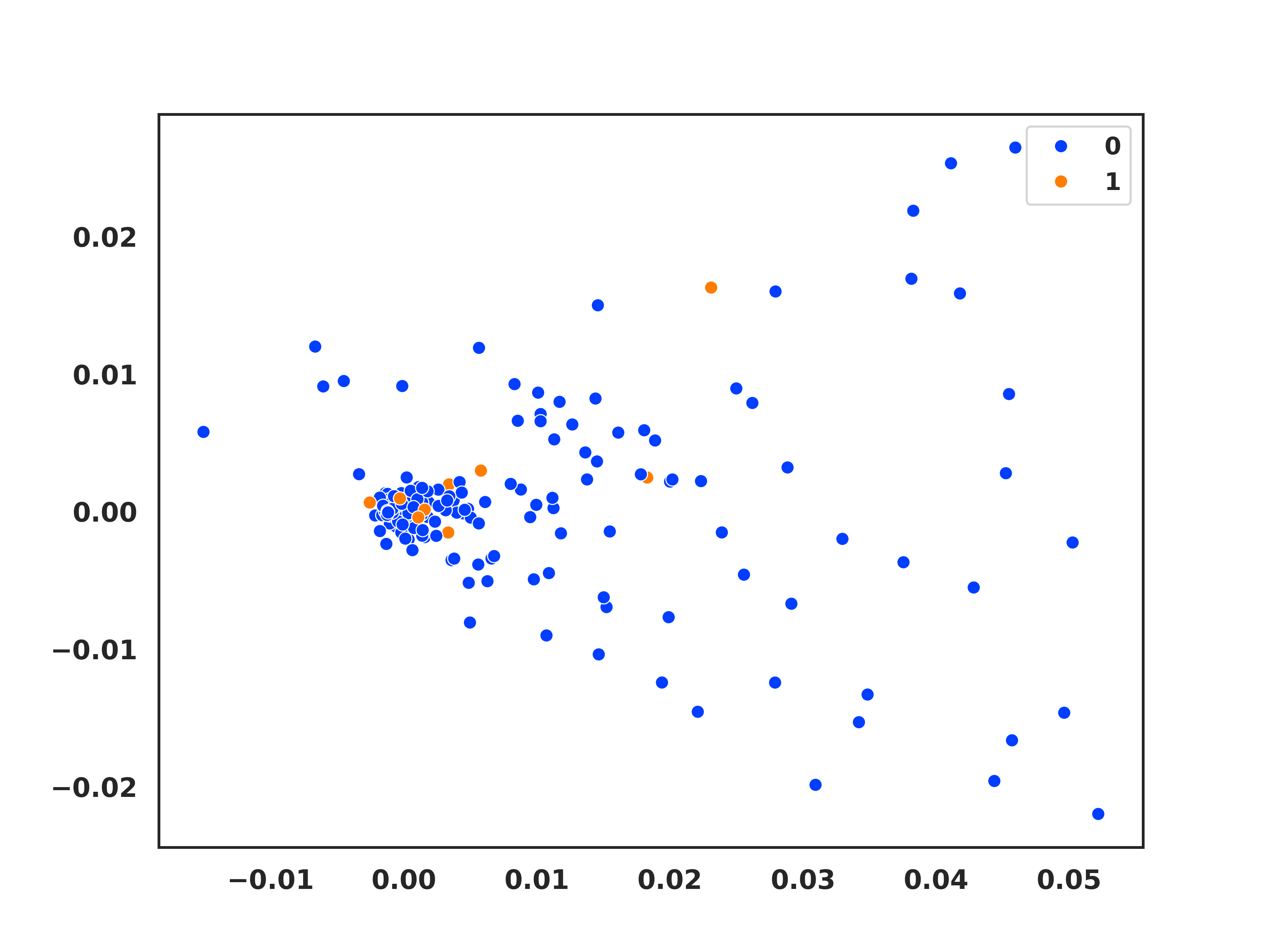}
    \caption{VesselMNIST}
  \end{subfigure}\hfill
  \begin{subfigure}[b]{0.3\textwidth}
    \includegraphics[width=\linewidth]{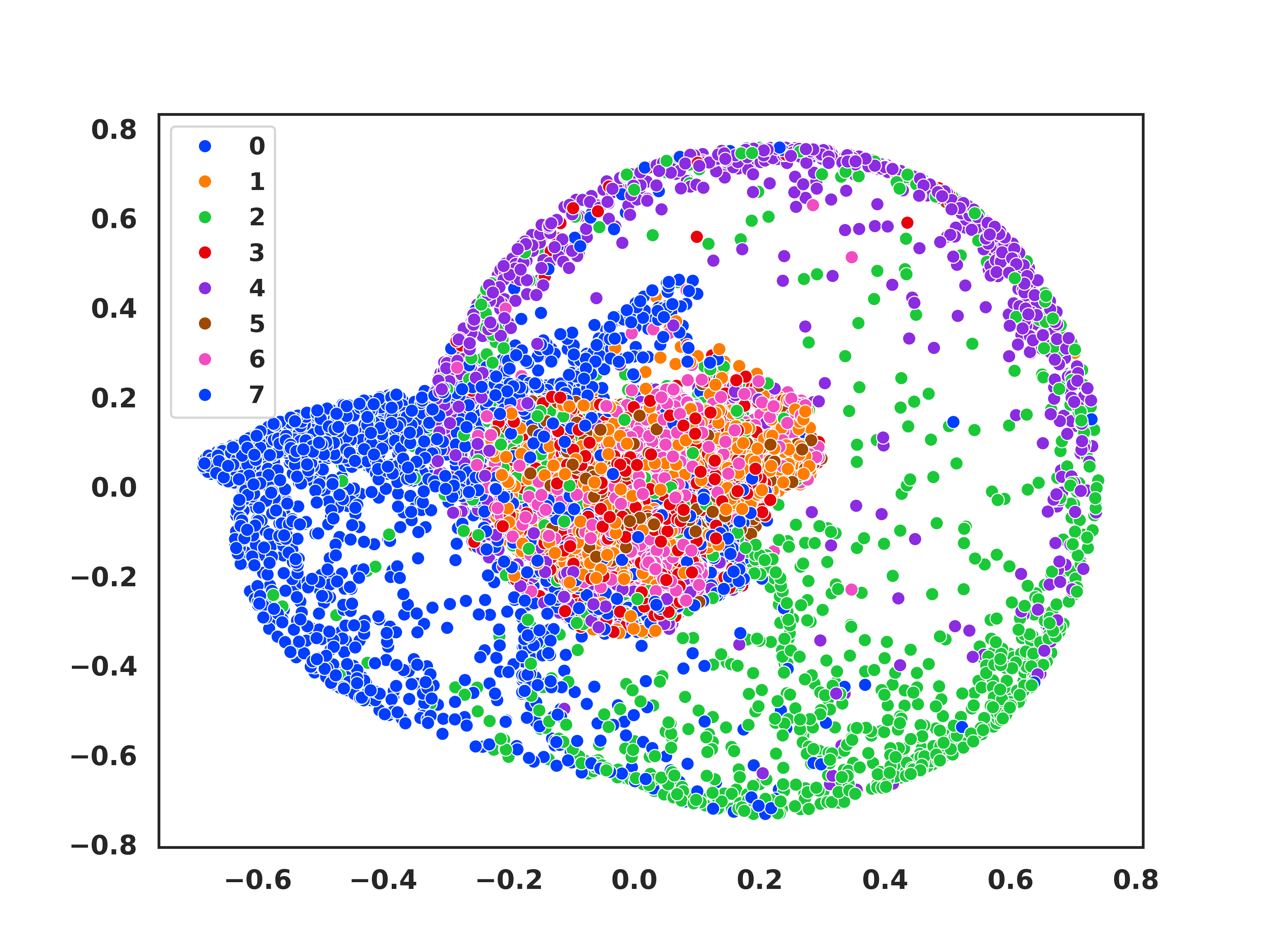}
    \caption{BloodMNIST}
  \end{subfigure}\hfill
  \begin{subfigure}[b]{0.3\textwidth}
    \includegraphics[width=\linewidth]{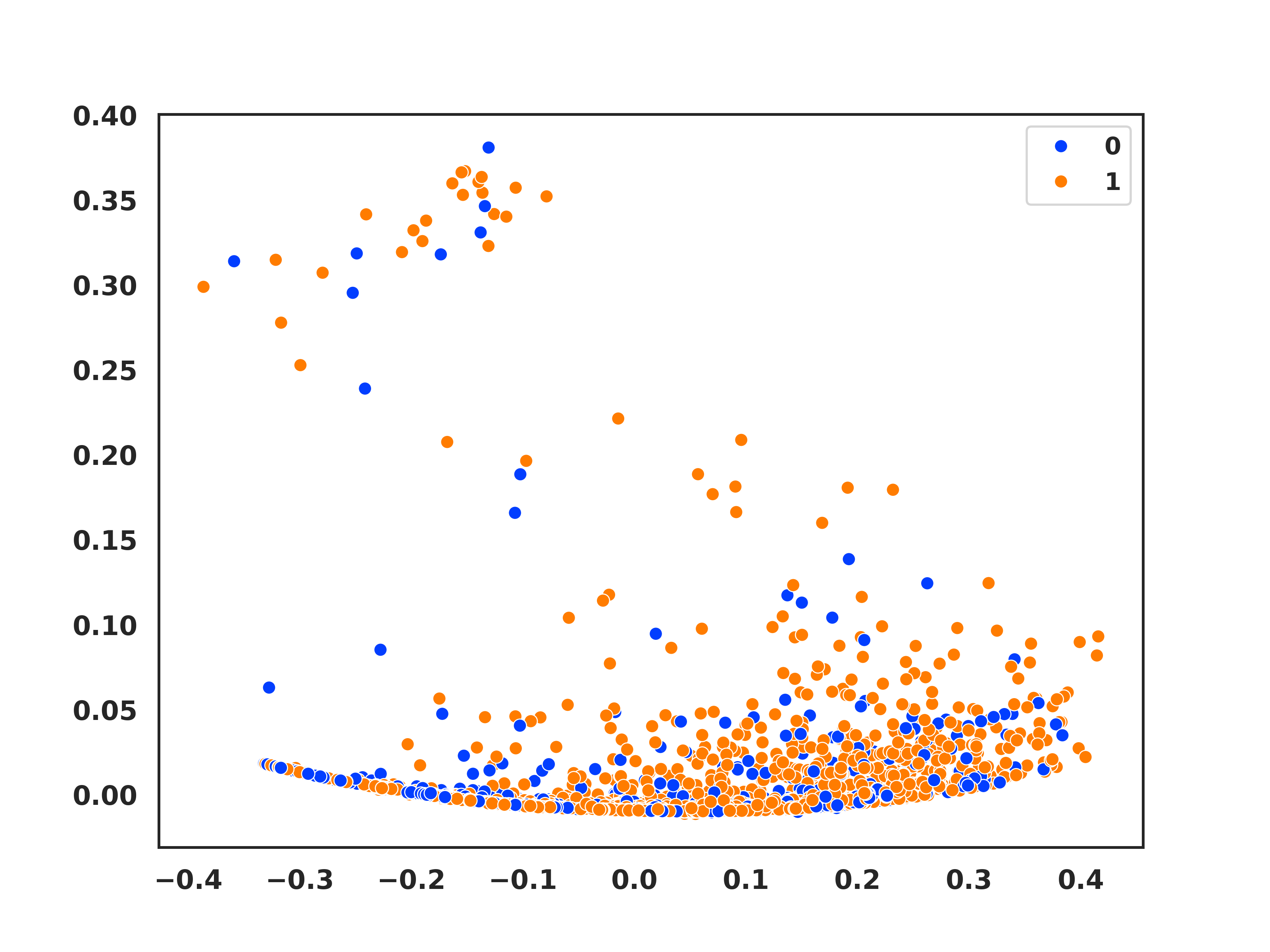}
    \caption{PneumoniaMNIST2D}
  \end{subfigure}
  
  \medskip
  
  \begin{subfigure}[b]{0.3\textwidth}
    \includegraphics[width=\linewidth]{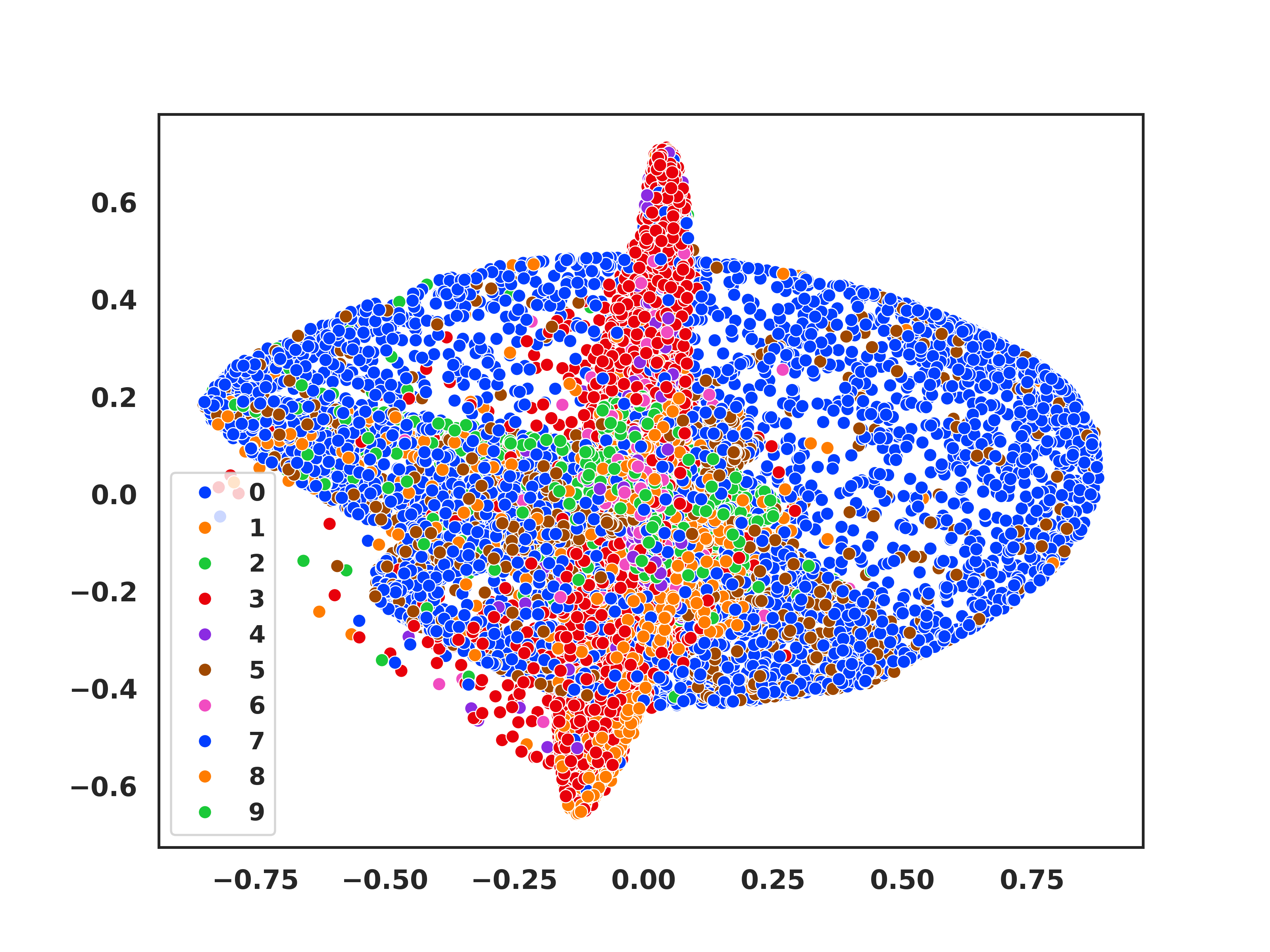}
    \caption{FMNIST}
  \end{subfigure}\hfill
  \begin{subfigure}[b]{0.3\textwidth}
    \includegraphics[width=\linewidth]{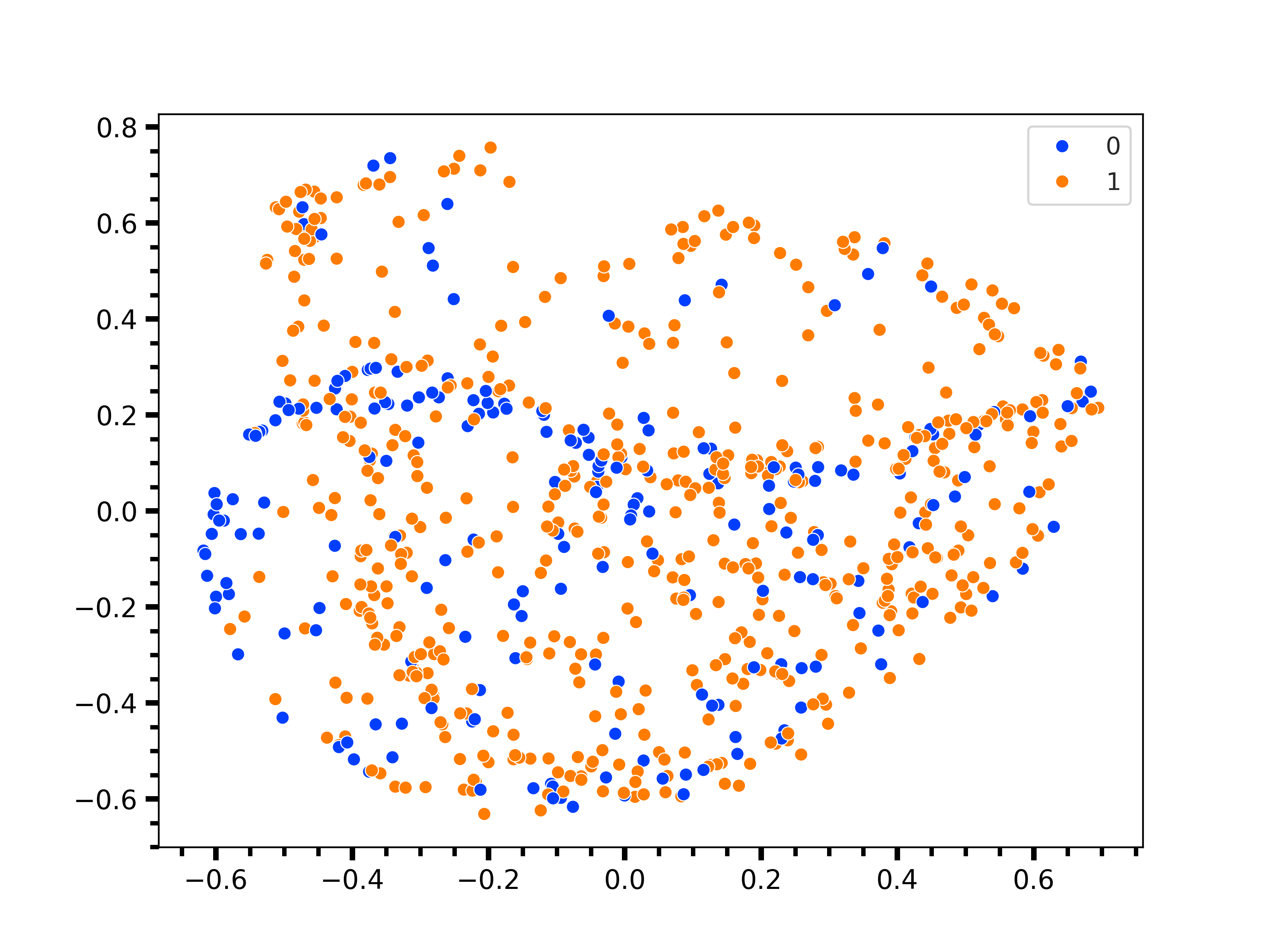}
    \caption{BreastMNIST}
  \end{subfigure}\hfill
  \begin{subfigure}[b]{0.3\textwidth}
    \includegraphics[width=\linewidth]{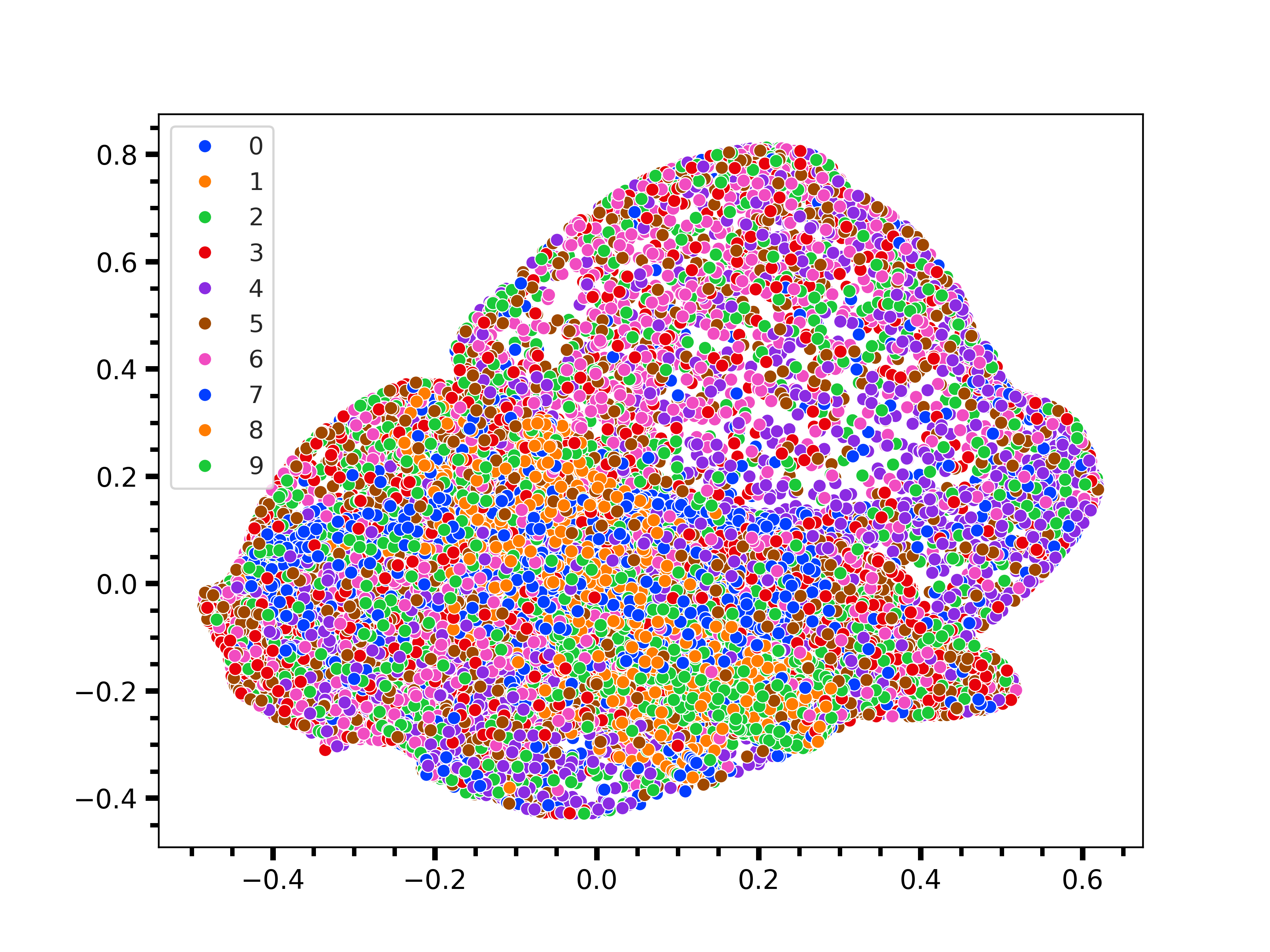}
    \caption{CIFAR10}
  \end{subfigure}
  
  \caption{Visualizations of the latent representations of CC after performing PCA.}
  \label{fig:lr_CC}
\end{figure}

\subsubsection{Sensitivity}
We investigate the sensitivity of R-DC to the data-dependant hyperparameters $\alpha$ and $\beta$ on PneumoniaMNIST. We explore values both higher and lower than the optimal ones that yielded the highest accuracy. Therefore, we select $\alpha$ and $\beta$ from the ranges [0.1, 0.6, 0.8, 0.9, 1] and [0.1, 0.3, 0.7, 1, 1.2], respectively. The other hyperparameters are design choices and are maintained fixed independently of the input dataset. As we can see, our model shows robust performance across a wide spectrum of values, excelling in both ACC and F1 Score metrics, as illustrated in Figure \ref{fig:lr_dyn_R-GMM-VGAE}.

\begin{figure}
  \centering
  
  \begin{subfigure}[b]{0.3\textwidth}
    \includegraphics[width=\linewidth]{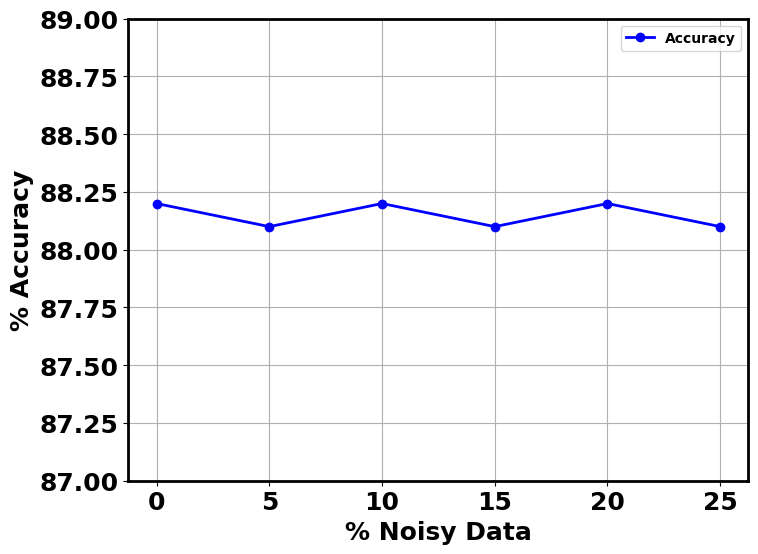}
    \caption{Accuracy-VesselMNIST3D}
  \end{subfigure}\hfill
  \begin{subfigure}[b]{0.3\textwidth}
    \includegraphics[width=\linewidth]{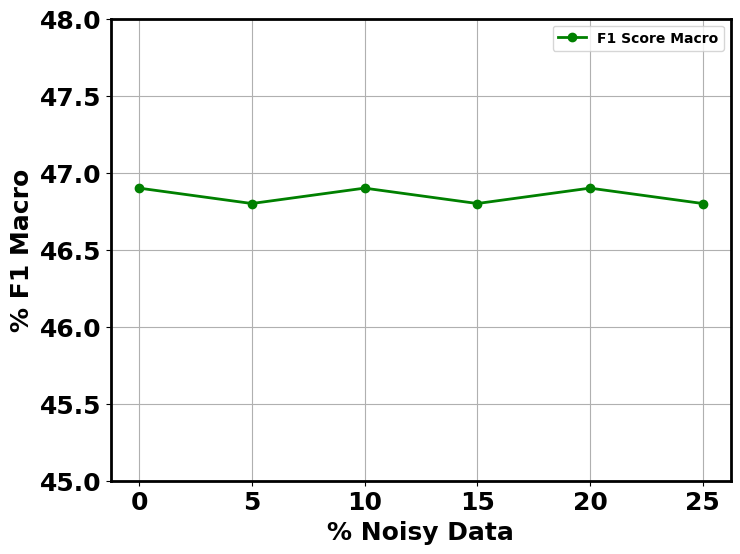}
    \caption{F1 Macro-VesselMNIST3D}
  \end{subfigure}\hfill
  \begin{subfigure}[b]{0.3\textwidth}
    \includegraphics[width=\linewidth]{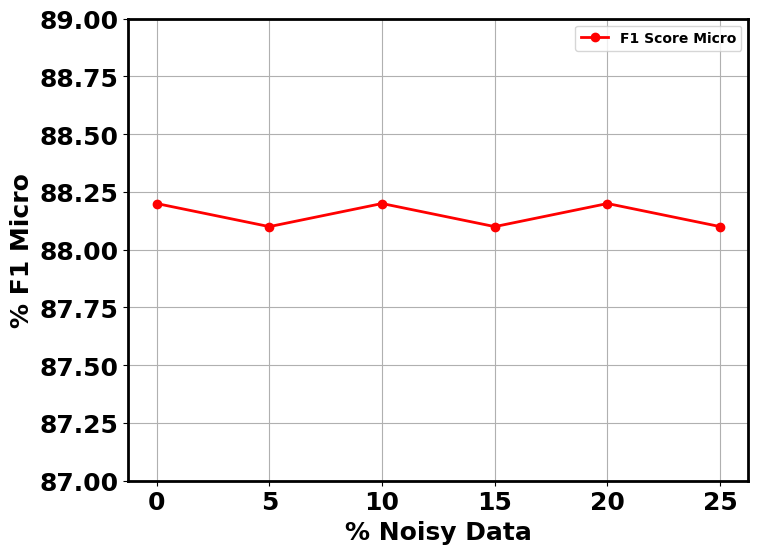}
    \caption{F1 Micro-VesselMNIST3D}
  \end{subfigure}
  
  \medskip
  
  \begin{subfigure}[b]{0.3\textwidth}
    \includegraphics[width=\linewidth]{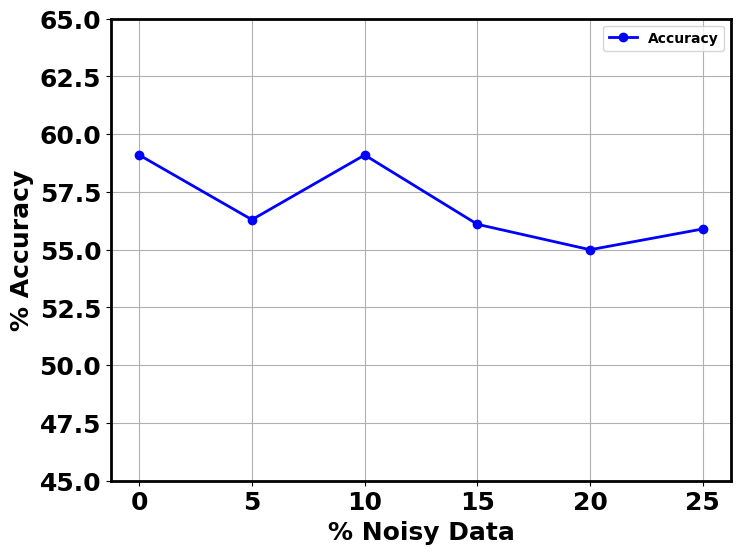}
    \caption{Accuracy-BloodMNIST}
  \end{subfigure}\hfill
  \begin{subfigure}[b]{0.3\textwidth}
    \includegraphics[width=\linewidth]{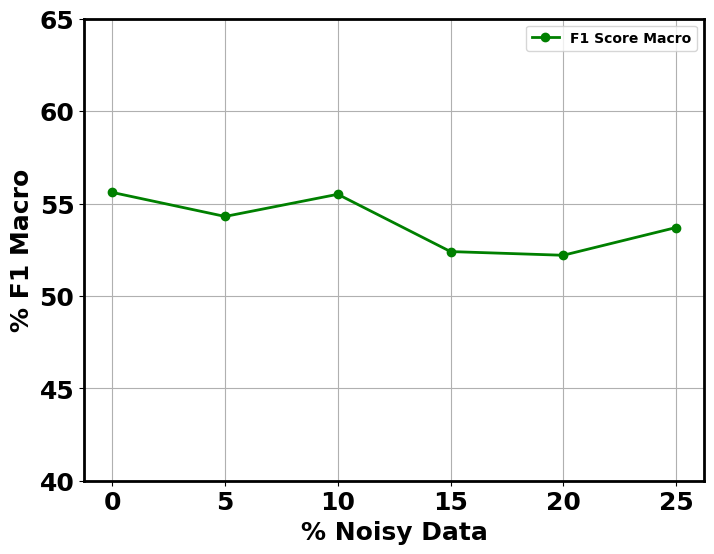}
    \caption{F1 Macro-BloodMNIST}
  \end{subfigure}\hfill
  \begin{subfigure}[b]{0.3\textwidth}
    \includegraphics[width=\linewidth]{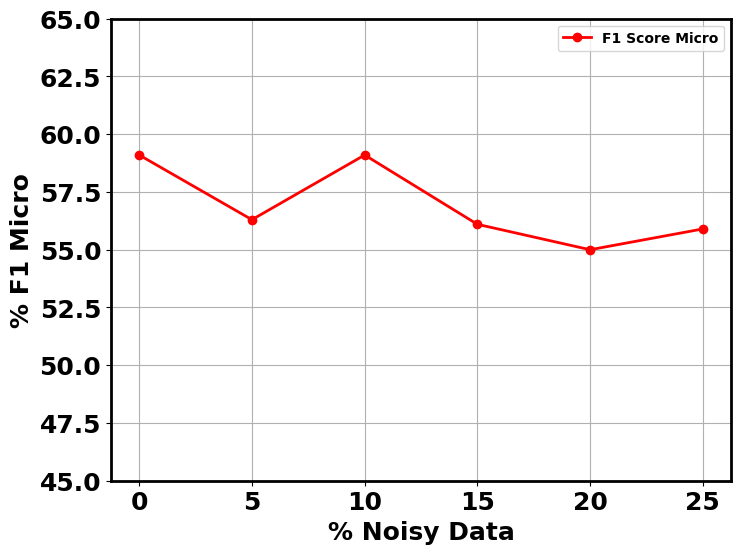}
    \caption{F1 Micro-BloodMNIST}
  \end{subfigure}
  
  \caption{Performance of R-DC on BloodMNIST and VesselMNIST3D, in terms of ACC, F1 Macro and F1 Micro, after adding Gaussian noise to the input data.}
  \label{fig:noisey_data}
\end{figure}

\subsubsection{Robustness}

We validate the robustness of our R-DC model by adding random noise. Initially, we perform feature-wise Z-score normalization on the input data matrix \(X\), resulting in the normalized matrix \(X^{\text{norm}}\). Subsequently, we randomly add Gaussian noise with a mean value of zero and standard deviation varying in \([0\%, 5\%, 10\%, 15\%, 20\%, 25\%]\) of the data's standard deviation, which is equal to 1 after normalization. The matrix obtained after adding Gaussian noise is denoted as \(X_{\text{noisy}}\). 

The process of normalization and adding Gaussian noise is described in Eq. \ref{Zscore} and Eq. \ref{noise}, respectively:
\begin{equation}
    x^{\text{norm}}_{ij} = \frac{x_{ij} - \mu_j}{\sigma_j},
\label{Zscore}
\end{equation}

\begin{equation}
    X_{\text{noisy}} = X^{\text{norm}} + \mathcal{N}(0, (\sigma_p)^2),
\label{noise}
\end{equation}
\noindent where \(x^{\text{norm}}_{ij}\) denotes the element of the \(i^{\text{th}}\) row and \(j^{\text{th}}\) column of matrix \(X^{\text{norm}}\), \(\mu_j\) is the mean of the \(j^{\text{th}}\) feature of \(X\), \(\sigma_j\) is the standard deviation of the \(j^{\text{th}}\) feature of \(X\), \(\mathcal{N}(0, (\sigma_p)^2)\) represents Gaussian noise with mean 0 and standard deviation \(\sigma_p\), and \(\sigma_p \in \left[0, 0.05, 0.1, 0.15, 0.2, 0.25\right]\).


The results, as depicted in Figure \ref{fig:noisey_data}, illustrate the performance of R-DC on BloodMNIST and VesselMNIST3D, in terms of ACC, F1 Macro, and F1 Micro, after adding Gaussian noise to the input data. R-DC yields stable performance across varying noise levels. These results validate the robustness of our model and its capacity to effectively handle real-world datasets that contain noise.




\subsubsection{Visualization}
We perform dimensionality reduction using PCA to get the 2D representations of the latent codes and then visualize these 2D representations. PCA gives better intuitions about the FT problem. The less curved the manifolds, the easier to identify the clusters in the latent space. In other words, we expect a model alleviating FT to yield clustering-oriented latent representations. This means that the similarities can be assessed effectively based on the Euclidean distance and the latent clusters can be detected based on approaches like K-means. Unlike T-SNE, PCA is a linear projector. When dealing with clustering-oriented representations, PCA is a better choice to assess the quality of the clusters. This dimensionality reduction technique preserves the global structure of the data by maintaining the directions of maximum variance. Thus, the overall distribution and Euclidean distances between data points are better preserved in the lower-dimensional space.

In Figure \ref{fig:lr_RDC} and Figure \ref{fig:lr_CC}, we provide the 2D visualizations of the latent representations of R-DC and CC, respectively, after performing PCA. From our results, it is clear that PCA preserves clustering-oriented structures for R-DC. For this latter, the similarities can be assessed effectively based on the Euclidean distance and latent clusters can be detected based on approaches like K-means. These results confirm that our approach can mitigate the FT problem effectively. Unlike R-DC, for CC the clustering structures are curved and highly overlapped after performing PCA. 


\section{Conclusion}

In this work, we rethought existing deep clustering paradigms to introduce a new strategy named R-DC. We addressed three limitations in existing DC paradigms, which are Feature Randomness, Feature Drift, and Feature Twist. Our new paradigm eliminates pseudo-supervision and solely relies on two levels of self-supervision. The proposed model R-DC extracts the core points and their most reliable neighbors to perform proximity-level self-supervision, which gradually replaces the instance-level self-supervision task. Remarkably, R-DC achieves state-of-the-art results compared to the most relevant DC approaches with notable enhancements in the finetuning phase across all datasets. The highest improvement is recorded in the BloodMNIST dataset, which shows a 24.6\% increase compared to the DynAE model. 

As a limitation, our approach requires identifying the nearest neighbors, which can introduce additional time overhead compared to some previous methods. In future work, we aim to explore other self-supervision techniques that can be used as proximity-level self-supervision strategies within the new paradigm. Additionally, we plan to extend our approach to graph-structured data by integrating a filtering mechanism that accommodates the graph's topological structure.

\newpage
\bibliographystyle{elsarticle-num-names} 
\bibliography{ref_FT}

\end{document}